\documentclass[twocolumn]{svjour3}
\smartqed

\usepackage{graphicx}

\usepackage{amsmath}
\usepackage{amssymb}
\usepackage{ascii}
\usepackage[T1]{fontenc}
\usepackage[misc]{ifsym}

\usepackage{algorithm}% http://ctan.org/pkg/algorithm
\usepackage{algpseudocode}% http://ctan.org/pkg/algorithmicx

\usepackage{epsfig}
\usepackage{support-caption}
\usepackage{subcaption}

\usepackage{booktabs}
\usepackage{pifont}
\usepackage[table,x11names]{xcolor}

\usepackage{multirow}

\usepackage{float}
\usepackage{times}
\usepackage{amsfonts}
\usepackage{color}
\usepackage{tabu}
\usepackage[authoryear,sort&compress,]{natbib}
\usepackage[colorlinks=true,urlcolor=blue,citecolor=blue,linkcolor=blue,bookmarks=true]{hyperref}
\usepackage[title]{appendix}
\usepackage[english]{babel}
\usepackage{bm}
\usepackage{colortbl}
\usepackage{enumitem}
\usepackage{array}
\usepackage{threeparttable}
\usepackage{dsfont}
\usepackage{nccmath}
\usepackage{listings,lstautogobble}
\usepackage{dsfont}

\graphicspath{{./fig/intro/}{./fig/approach/}{./fig/exp/}{./}}

\definecolor{codegreen}{rgb}{0,0.5,0}
\definecolor{codeblue}{rgb}{0.25,0.5,0.5}
\definecolor{codegray}{rgb}{0.6,0.6,0.6}

\makeatletter\renewcommand\paragraph{\@startsection{paragraph}{4}{\z@}
	{.5em \@plus1ex \@minus.2ex}{-.5em}{\normalfont\normalsize\bfseries}}\makeatother
\newlength\savewidth\newcommand\shline{\noalign{\global\savewidth\arrayrulewidth
	\global\arrayrulewidth 1.5pt}\hline\noalign{\global\arrayrulewidth\savewidth}}

\newcommand{\blue}[1]{\color{blue}{#1}}

\journalname{IJCV}
\begin{document}\sloppy
	
	\title{Revisiting DETR Pre-training for Object Detection}
	
	\titlerunning{Revisiting DETR Pre-training for Object Detection}        %
	\author{
		Yan Ma$^{\textbf{1}}$\and
		Weicong Liang$^\textbf{2}$\and 
		Bohan Chen$^{\textbf{3}}$\and 
            Yiduo Hao$^\textbf{6}$\and \\
            Bojian Hou$^{4}$\;
		Xiangyu Yue$^{\textbf{5}}$\and
		Chao Zhang$^\textbf{2}$\and
            Yuhui Yuan$^\textbf{6}$
	}
	
	\authorrunning{Ma et al.} %
	
	\institute{
		$^{\textbf{1}}$University of Toronto\;\;
		$^{\textbf{2}}$Peking University\;\;
            $^{\textbf{3}}$Xi'an Jiaotong-Liverpool University\;\;
            $^{\textbf{4}}$University of Pennsylvania\;\;
		$^{\textbf{5}}$CUHK\;\;
            $^{\textbf{6}}$Microsoft Research Asia \quad
		\Letter \; yuhui.yuan@microsoft.com
	}
	
	
	\maketitle
	 
	\vspace{-2mm}
	\begin{abstract}
          Motivated by the remarkable achievements of DETR-based approaches on COCO object detection and segmentation benchmarks, recent endeavors have been directed towards elevating their performance through self-supervised pre-training of Transformers while preserving a frozen backbone. Noteworthy advancements in accuracy have been documented in certain studies.
          Our investigation delved deeply into a representative approach, DETReg, and its performance assessment in the context of emerging models like $\mathcal{H}$-Deformable-DETR. Regrettably, DETReg proves inadequate in enhancing the performance of robust DETR-based models under full data conditions. To dissect the underlying causes, we conduct extensive experiments on COCO and PASCAL VOC probing elements such as the selection of pre-training datasets and strategies for pre-training target generation. By contrast, we employ an optimized approach named Simple Self-training which leads to marked enhancements through the combination of an improved box predictor and the Objects$365$ benchmark.
          The culmination of these endeavors results in a remarkable AP score of $59.3\%$ on the COCO \texttt{val} set, outperforming $\mathcal{H}$-Deformable-DETR + Swin-L without pre-training by $1.4\%$. Moreover, a series of synthetic pre-training datasets, generated by merging contemporary image-to-text(LLaVA) and text-to-image (SDXL) models, significantly amplifies object detection capabilities.\
	 \end{abstract}
	
%%%%%%%%%%%%%%%%%%%%%%%%%%%%%%%%%%%%%%%%%%%%%%%%%%%%%%%%%%%
\section{Introduction}
\label{sec:introduction}
%%%%%%%%%%%%%%%%%%%%%%%%%%%%%%%%%%%%%%%%%%%%%%%%%%%%%%%%%%%
Recently, the DETR-based approaches~\citep{carion2020end,zhang2022dino,jia2023detrs,li2023mask,zhu2020deformable} have achieved significant progress and pushed the frontier on both object detection and segmentation tasks.
For example, DINO-DETR~\citep{zhang2022dino}, $\mathcal{H}$-Deformable-DETR~\citep{jia2023detrs}, and Group-DETRv2~\citep{chen2022group} have set new state-of-the-art object detection performance on COCO benchmark.
Mask-DINO~\citep{li2023mask} further extends DINO-DETR and establishes the best results across COCO instance segmentation and panoptic segmentation tasks.
To some degree, this is the first time that end-to-end transformer approaches can achieve an even better performance than the conventional heavily tuned strong detectors~\citep{liu2022swin,li2021improved} based on convolution, e.g., Cascade Mask-RCNN and HTC++.
    
\begin{figure}[t]
\centering
\begin{minipage}[t]{0.5\textwidth}
\begin{subfigure}[b]{0.425\textwidth}
{\includegraphics[width=\textwidth]{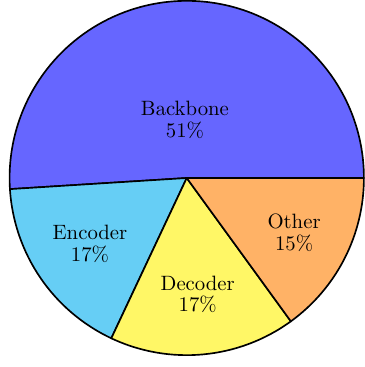}}
\caption{\#parameters}
\label{fig:intro:parameters}
\end{subfigure}
\hspace{5mm}
\begin{subfigure}[b]{0.425\textwidth}
{\includegraphics[width=\textwidth]{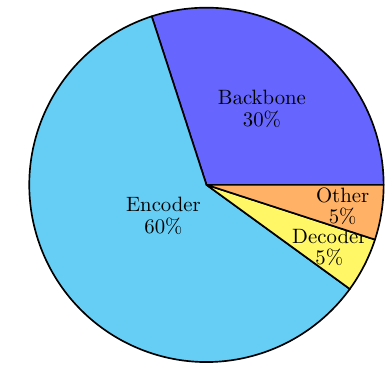}}
\caption{\#GFLOPs}
\label{fig:intro:gflops}
\end{subfigure}
\end{minipage}
\vspace{1mm}
\begin{minipage}[t]{0.5\textwidth}
\hspace{-5mm}
\begin{subfigure}[b]{1\textwidth}
{\includegraphics[width=\textwidth]{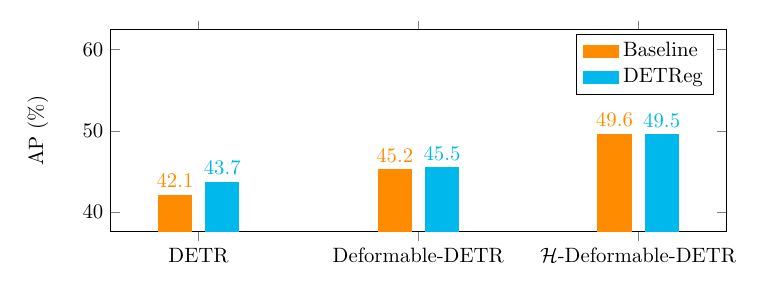}}
\caption{COCO object detection results of DETReg.}
\label{fig:intro:detreg}
\end{subfigure}
\end{minipage}
\caption{\small{\textbf{The distribution of the number of parameters and GFLOPs within Deformable-DETR network with a ResNet$50$ backbone, and the pre-training performance of DETReg.} As shown in (a) and (b), we can see that around $34\%$ parameters and $65\%$ GFLOPs are distributed in the randomly initialized Transformer encoder and decoder.
According to (c), DETReg only improves the vanilla DETR and Deformable-DETR by +$1.6\%$ and +$0.3\%$ while showing no gains over the stronger $\mathcal{H}$-Deformable-DETR.
}}
\label{fig:intro}
\end{figure}

Despite the great success of these DETR-based approaches, they still choose a randomly initialized Transformer and thus fail to unleash the potential of a fully pre-trained detection architecture like~\citep{wei2021aligning}, which already verifies the benefits of aligning the pre-training architecture with the downstream architecture.
Figure~\ref{fig:intro:parameters} and~\ref{fig:intro:gflops} illustrate the distribution of the number of parameters and GFLOPs within a standard Deformable-DETR network based on ResNet$50$ backbone.
We can see that the Transformer encoder and decoder occupy $34\%$ of the parameters and $65\%$ of the GFLOPs, which means there exists much room for improvement along the path of performing pre-training on the Transformer part within DETR.

Several recent works have improved DETR-based object detection models by performing self-supervised pre-training on the Transformer encoder and decoder while freezing the backbone.
For example, UP-DETR~\citep{dai2021up} pre-trains Transformer to detect random patches in an image, DETReg~\citep{bar2022detreg} pre-trains Transformer to match object locations and features with priors generated from Selective Search algorithm, and most recently, Siamese DETR locates the target boxes with the query features extracted from a different view's corresponding box. However, these works utilize either the vanilla DETR network (AP=$42.1\%$ in terms of object detection performance on COCO) or the Deformable-DETR variant (AP=$45.2\%$). Their results fall significantly short when pre-training on the latest much stronger DETR model like $\mathcal{H}$-Deformable-DETR~\citep{jia2023detrs} (AP=$49.6\%$).
In Figure~\ref{fig:intro:detreg}, we present the object detection results of different DETR models on COCO under two conditions: without pre-training of the Transformer component (referred to as the \textit{baseline}) and with pre-training using the DETReg method. In both cases, the backbones of these models are ResNet$50$ initialized with SwAV~\citep{caron2020unsupervised}. Notably, in the case of the $\mathcal{H}$-Deformable-DETR, the utilization of the DETReg pre-training actually leads to a performance decrease rather than an improvement.

In this work, we first take a closer look at how much self-supervised pre-training methods, exemplified by DETReg, can improve over the increasingly potent DETR models on COCO object detection benchmark.
Our investigation unveils a significant limitation in the efficacy of DETReg when applied to fortified DETR networks bolstered by improvements like the SwAV pre-trained backbone, deformable techniques in Deformable-DETR, and the hybrid matching scheme in $\mathcal{H}$-Deformable-DETR. 
We pinpoint the crux of the issue as originating from unreliable box proposals generated by unsupervised methods like Selective Search, which contribute to noisy localization targets, and the weak semantic information provided through feature reconstruction which is not an efficient classification target either. These drawbacks make the self-supervised pre-training methods ineffective when applied to an already strong DETR model. 

To fix this, we propose to use a COCO object detector to get more accurate pseudo-boxes with informative pseudo-class labels. Extensive ablation experiments underscore the impact of three pivotal factors: the choice of pre-training datasets (ImageNet vs. Objects$365$), localization pre-training targets (Selective Search proposals vs. pseudo-box predictions), and classification pre-training targets (object-embedding vs. pseudo-class predictions).

Our findings reveal that a Simple Self-training scheme, employing pseudo-box and pseudo-class predictions as pre-training targets, outperforms the DETReg approach in various settings. Notably, this simple design yields discernible pre-training enhancements even for the state-of-the-art DETR network without accessing the pre-training benchmark's ground-truth label. For example, with a ResNet$50$ backbone and the Objects$365$ pre-training dataset, Simple Self-training elevates DETReg's COCO object detection results on $\mathcal{H}$-Deformable-DETR by $3.6\%$. Furthermore, a remarkable performance is observed with the Swin-L backbone, yielding competitive results $59.3\%$.

Additionally, we delve into an exploration of contemporary image-to-text and text-to-image generation models, aiming to create a sequence of synthetic datasets for object detection pre-training. Empirically, our observations yield encouraging outcomes, as pre-training with these synthetic datasets demonstrates commendable performance even when compared against the widely adopted Objects365 benchmark, which entails substantial annotation costs. In general, our efforts are poised to provide a more authentic assessment of the progress in the formidable task of DETR pre-training. 
	
%%%%%%%%%%%%%%%%%%%%%%%%%%%%%%%%%%%%%%%%%%%%%%%%%%%%%%%%%%%
\section{Related Work}
%%%%%%%%%%%%%%%%%%%%%%%%%%%%%%%%%%%%%%%%%%%%%%%%%%%%%%%%%%%

\paragraph{DETR for object detection.} Since the emergence of DETR~\citep{carion2020end} as the first fully end-to-end object detector, many works have extended DETR with novel techniques to achieve state-of-the-art results on various vision tasks. 
To accelerate the convergence of the original DETR, Deformable-DETR~\citep{zhu2020deformable} proposes a novel multi-scale deformable self/cross-attention to focus on a sparse set of important sampling points around a reference point.
Furthermore, based on DAB-DETR~\citep{liu2022dab} with a different query formulation, DINO-DETR~\citep{zhang2022dino} introduces a query denoising scheme and sets new records on object detection tasks.
Besides, to address the training efficiency bottleneck caused by one-to-one matching in DETR,
$\mathcal{H}$-Deformable-DETR~\citep{jia2023detrs} and Group-DETR~\citep{chen2022group} propose to train with more queries in the transformer decoder with an additional one-to-many matching scheme, which helps to achieve even faster convergence and better performance.

    \begin{figure*}[t]
      \centering
       \includegraphics[width=0.9\linewidth]{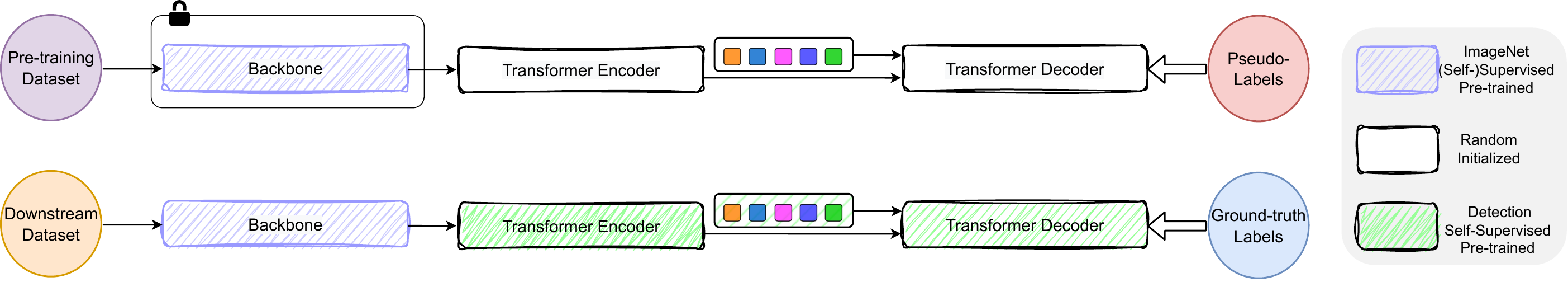}
       \caption{\small{\textbf{The overall framework of self-supervised pre-training scheme}.
    There are two steps to pre-train the DETR network. In the first step, we freeze the backbone and pre-train a randomly initialized Transformer encoder and decoder with the well-designed pre-training target on a large-scale pre-training benchmark. In the second step, we initialize the encoder and decoder with pre-trained weights and fine-tune all the parameters of the DETR network on the downstream dataset supervised by ground-truth labels.
    }}
       \label{fig:detr_pretrain}
    \end{figure*}

\vspace{1mm} 
\paragraph{Self-supervised pre-training.} Self-supervised learning (SSL) has achieved remarkable results in image classification methods such as MoCo~\citep{he2020momentum}, SimCLR~\citep{chen2020simple}, and BYOL~\citep{grill2020bootstrap}. However, SSL on object detection has shown limited transferability. To overcome this challenge, many works have proposed pretext tasks that leverage region or pixel localization cues to enhance the pre-training signals. For example, InsLoc~\citep{yang2021instance} uses contrastive learning on foreground patches to learn instance localization. UniVIP~\citep{li2022univip} exploits scene similarity, scene-instance correlation, and instance discrimination to capture semantic affinity. CP$^2$~\citep{wang2022cp} employs pixel-wise contrastive learning to facilitate both image-level and pixel-level representation learning. Unlike most of these methods that aim to improve conventional object detectors such as Faster R-CNN or Cascade R-CNN, we focus on designing an effective pre-training scheme for the state-of-the-art DETR-based detector.
	
\vspace{1mm}
\paragraph{DETR pre-training.} DETR typically relies on a supervised pre-trained backbone on ImageNet and random initialization of the transformer encoder and decoder. Some recent works have explored pre-training the transformer component of DETR for enhanced object detection performance. For example, UP-DETR~\citep{dai2021up} introduces an unsupervised pretext task to detect and reconstruct random patches of the input. DETReg~\citep{bar2022detreg} refines the pretext task by using unsupervised region proposals from Selective Search~\citep{uijlings2013selective} instead of random patches and also reconstructs the object embeddings of these regions from its SwAV~\citep{caron2020unsupervised} backbone to learn invariant representations. Siamese DETR~\citep{huang2023siamese} employs a siamese self-supervised learning approach to pre-train DETR in a symmetric pipeline where each branch takes one view as input and aims to locate and discriminate the corresponding regions from another view. However, these pre-training methods only yield minor improvements to a strong DETR variant like Deformable-DETR.

\vspace{1mm}
\paragraph{Self-training.} Self-training is a powerful technique for improving various computer vision tasks, such as image classification~\citep{li2023mask,sahito2022better}, object detection~\citep{yang2021interactive,vandeghen2022adaptive}, and segmentation~\citep{zhu2021improving}. A common self-training method is NoisyStudent~\citep{xie2020self}, which trains a teacher model on labeled data and uses it to generate pseudo-labels on unlabeled images. These pseudo-labels are then used to train a student model, and this process is repeated to obtain better models by updating the teacher model with the previous student model. The ASTOD~\citep{vandeghen2022adaptive} framework applies an iterative self-training process for object detection, using multiple image views to produce high-quality pseudo-labels. ST++\citep{yang2022st++} is a recent self-training algorithm for segmentation tasks, which uses confidence scores to filter out incorrect pseudo-labels. \citep{zoph2020rethinking} has demonstrated that self-training outperforms traditional pre-training methods in various scenarios, including low and high data regimes, and can even succeed when pre-training methods fail. Unlike these complex self-training schemes that use an iterative approach to refine pseudo-labels, we propose a Simple Self-training scheme that generates pseudo-labels only once by keeping a fixed number of the most confident predictions.

%%%%%%%%%%%%%%%%%%%%%%%%%%%%%%%%%%%%%%%%%%%
\section{Approach}
\label{methods}
%%%%%%%%%%%%%%%%%%%%%%%%%%%%%%%%%%%%%%%%%%%
In this work, we focus on studying how to perform pre-training over the Transformer encoder and decoder parts within DETR for object detection tasks following~\citep{dai2021up,bar2022detreg}. The goal of DETR pre-training is to design an effective pretext task that can make the best use of a large-scale unlabeled dataset that has no ground-truth bounding box annotations.

\vspace{-3mm}
\textbf{\subsection{Formulation}}
\label{formulation}
The conventional DETR model has three components, the backbone extracting the image feature, the encoder enhancing the feature with a self-attention mechanism, and the decoder turning query inputs into object class and location predictions through cross-attention with image features. The existing self-supervised pre-training methods share a similar scheme that optimizes the encoder and decoder network parameters on the pre-training dataset while freezing a pre-trained backbone. After pre-training, all three components are tuned together on the downstream dataset. The pipeline is illustrated in Figure~\ref{fig:detr_pretrain}.
	
\vspace{1mm}
\paragraph{Preliminary.}
In the following article, we formulate the general self-supervised pre-training process as several equations. We use $f_{\theta_{\sf{B}}}$, $f_{\theta_{\sf{E}}}$, $f_{\theta_{\sf{D}}}$ to represent the backbone, Transformer encoder, and Transformer decoder within a DETR network parameterized by $\theta_{\sf{B}}$, $\theta_{\sf{E}}$, and $\theta_{\sf{D}}$. The input images from the pre-training and downstream dataset are denoted as $\overline{\mathbb{X}}=\{\overline{\mathbf{x}}_1, \cdots, \overline{\mathbf{x}}_N\}$ and $\mathbb{X}=\{\mathbf{x}_1, \cdots, \mathbf{x}_M\}$ respectively, where $N${$\gg$}$M$. The ground-truth label of downstream data is $\mathbb{Y}{=}\{\mathbf{y}_1,\cdots,\mathbf{y}_M \vert \mathbf{y}_i{=}(\mathbf{c}_i, \mathbf{b}_i)\}$, where $\mathbf{c}_i$ is the category label and $\mathbf{b}_i$ is the box location label.
Typically, the domain-specific pre-training data labels are lacking and most works choose to generate the pseudo-labels, i.e., $\overline{\mathbb{Y}}=\{\overline{\mathbf{y}}_1, \cdots, \overline{\mathbf{y}}_N\}$ instead. 
    
\vspace{1mm}
\paragraph{Pre-train.}
We illustrate the mathematical formulations of the DETR pre-training with Equation~\ref{eq1} and~\ref{eq2}.
Specifically, the pre-training input $\overline{\mathbf{x}}_i$ is forwarded through the backbone $f_{\theta_{\sf{B}}}$, encoder $f_{\theta_{\sf{E}}}$, and decoder $f_{\theta_{\sf{D}}}$ to get the prediction $\overline{\mathbf{z}}_{i}$. 
Here $\theta_{\sf{B}}$, $\theta_{\sf{E}}$, $\theta_{\sf{D}}$ represent the learnable parameters for the three network components respectively. $\theta_{\sf{B}}$ is initialized with SwAV~\citep{caron2020unsupervised} self-supervised pre-training method and frozen during pre-training. $\theta_{\sf{E}}$ and $\theta_{\sf{D}}$ are randomly initialized and then optimized to minimize the pre-training loss $\mathcal{L}_{\textrm{pre}}(\cdot)$, which is calculated with network output $\overline{\mathbf{z}}_{i}$ and pre-training target $\overline{\mathbf{y}}_i$.
    
\begin{align}
&\overline{\mathbf{z}}_{i}=f_{\theta_{\sf{D}}}(f_{\theta_{\sf{E}}}(f_{\theta_{\sf{B}}}(\overline{\mathbf{x}}_i)), \mathbb{Q}), \label{eq1} \\
&\widehat{\theta}_{\sf{D}}, \widehat{\theta}_{\sf{E}}, \widehat{\mathbb{Q}} = \underset{\theta_{\sf{D}},\theta_{\sf{E}},\mathbb{Q}}{\mathrm{argmin}} \;\sum_{i=1}^{N}\mathcal{L}_{\textrm{pre}}(\overline{\mathbf{z}}_{i}, \overline{\mathbf{y}}_i), \label{eq2}
\end{align}
where $\mathbb{Q}=\{\mathbf{q}_1,\cdots,\mathbf{q}_k\}$ represents the learnable object query of decoder and will also be jointly optimized with the encoder/decoder parameters. $\widehat{\theta}_{\sf{D}}$, $\widehat{\theta}_{\sf{E}}$, $\widehat{\mathbb{Q}}$ represent the decoder parameters, encoder parameters, and object query after pre-training.
In the following section~\ref{instantiations}, we will illustrate the formulation of $\mathcal{L}_{\textrm{pre}}$ in different methods.

\vspace{1mm}
\paragraph{Fine-tune.}
We obtain the optimized encoder and decoder parameter $\widehat{\theta}_{\sf{E}}$, $\widehat{\theta}_{\sf{D}}$ during pre-training. Then we tune the same network on the downstream data $\mathbf{x}_i$. Here, we initialize the backbone, encoder and decoder parameter with $\theta_{\sf{B}}$, $\widehat{\theta}_{\sf{E}}$, $\widehat{\theta}_{\sf{D}}$, and denote the network output as $\mathbf{z}_i$. All parameters of the three components and learnable query $\mathbb{Q}$ are optimized to minimize the downstream loss $\mathcal{L}_{\textrm{ds}}(\cdot)$ between $\mathbf{z}_i$ and downstream label $\mathbf{y}_i$.

    \begin{align}
    &\mathbf{z}_i=f_{\widehat{\theta}_{\sf{D}}}(f_{\widehat{\theta}_{\sf{E}}}(f_{\theta_{\sf{B}}}(\mathbf{x}_i)),  \widehat{\mathbb{Q}}), \label{eq3} \\
    &\widetilde{\theta}_{\sf{D}}, \widetilde{\theta}_{\sf{E}},\widetilde{\theta}_{\sf{B}}, \widetilde{\mathbb{Q}}=\underset{\widehat{\theta}_{\sf{D}},\widehat{\theta}_{\sf{E}},\theta_{\sf{B}}, \widehat{\mathbb{Q}}}{\mathrm{argmin}} \; \sum_{i=1}^{M}\mathcal{L}_{\textrm{ds}}(\mathbf{z}_i, \mathbf{y}_i), \label{eq4}
    \end{align}
where $\widetilde{\theta}_{\sf{D}}, \widetilde{\theta}_{\sf{E}},\widetilde{\theta}_{\sf{B}}, \widetilde{\mathbb{Q}}$ are optimized decoder, encoder, backbone parameters, and object query after downstream finetuning.

\vspace{-5mm}
\textbf{\subsection{Instantiations} \label{instantiations}}
Assume the target of the $i$-th pre-training input can be denoted as $\overline{\mathbf{y}}_i=\{\overline{\mathbf{y}}_{i1},\cdots,\overline{\mathbf{y}}_{im}\}$, where $m$ is the number of objects in each target. The network output consists of $k$ bounding box predictions, which is the same as the number of object queries. We denote the corresponding prediction as $\overline{\mathbf{z}}_i=\{\overline{\mathbf{z}}_{i1},\cdots,\overline{\mathbf{z}}_{ik}\}$.
Typically, the number of targets in $\overline{\mathbf{y}}_i$ is less than $30$, while we set our DETR network to output $100$ or $300$ predictions, so $m<k$. Thus we pad the targets with no-object category $\varnothing$ following DETR~\citep{carion2020end} to be of size $k$.
Then, DETR performs one-to-one alignment via Hungarian bipartite matching algorithm~\citep{kuhn1955hungarian} over $\overline{\mathbf{y}}_i$ and $\overline{\mathbf{z}}_i$. We illustrate the mathematical formulation in Equation~\ref{eq6}, which computes the optimal label assignment for each prediction by minimizing the matching cost function $\mathcal{L}_{\textrm{match}}(\cdot)$:
    
\begin{equation} \label{eq6}
\sigma_{i} = \underset{\sigma_{i} \in \Sigma_k}{\mathrm{argmin}} \;\sum_{j=1}^{k}\mathcal{L}_{\textrm{match}}(\overline{\mathbf{y}}_{ij}, \overline{\mathbf{z}}_{i\sigma_{i}(j)}),
\end{equation}
where $\Sigma_k$ represents all permutations over $k$ elements and $\sigma_i(j)$ maps the targeted box $j$ to the most similar predicted box within the $i$-th input. The matching cost function $\mathcal{L}_{\textrm{match}}(\cdot)$ measures the predictions from two aspects including the localization accuracy and classification accuracy following DETR~\citep{carion2020end}.
    
\begin{figure*}[t]
\begin{minipage}[t]{\linewidth}
\centering
\begin{subfigure}[b]{\textwidth}
\centering
\includegraphics[width=125mm,height=32mm]{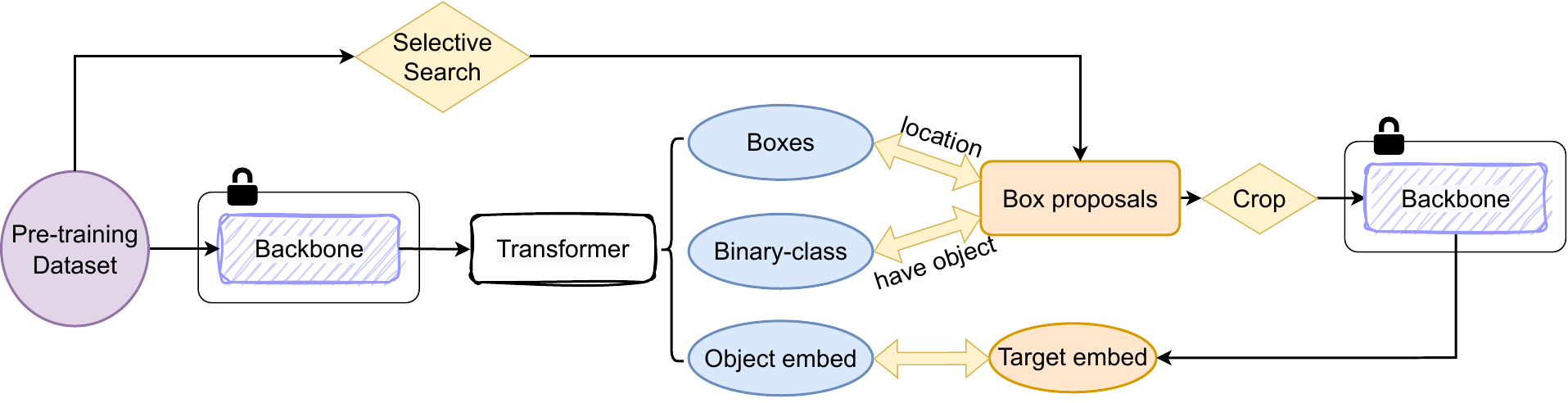}
\caption{\footnotesize{DETReg}}
\label{fig:pretrain_detreg_pipeline}
\end{subfigure}
\end{minipage}
\begin{minipage}[t]{\linewidth}
\centering
\begin{subfigure}[b]{\textwidth}
\centering
\includegraphics[width=125mm,height=35mm]{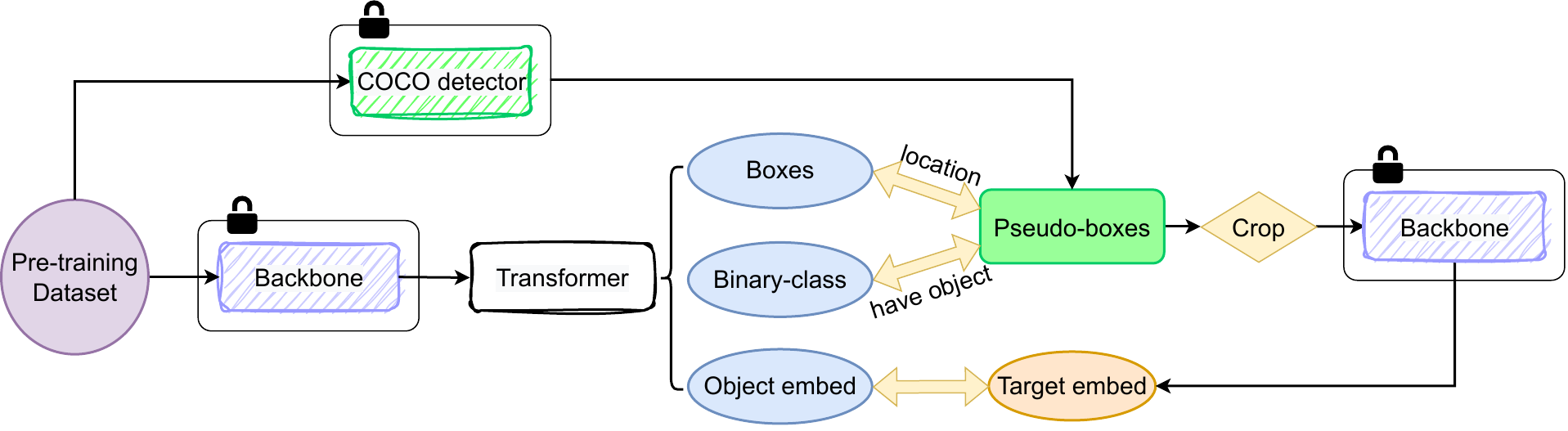}
\caption{\footnotesize{DETReg+pseudo-box}}
\label{fig:pretrain_detreg_pseudo_box_pipeline}
\end{subfigure}
\end{minipage}
\begin{minipage}[t]{\linewidth}
\centering
\begin{subfigure}[b]{\textwidth}
\centering
\includegraphics[width=125mm,height=27mm]{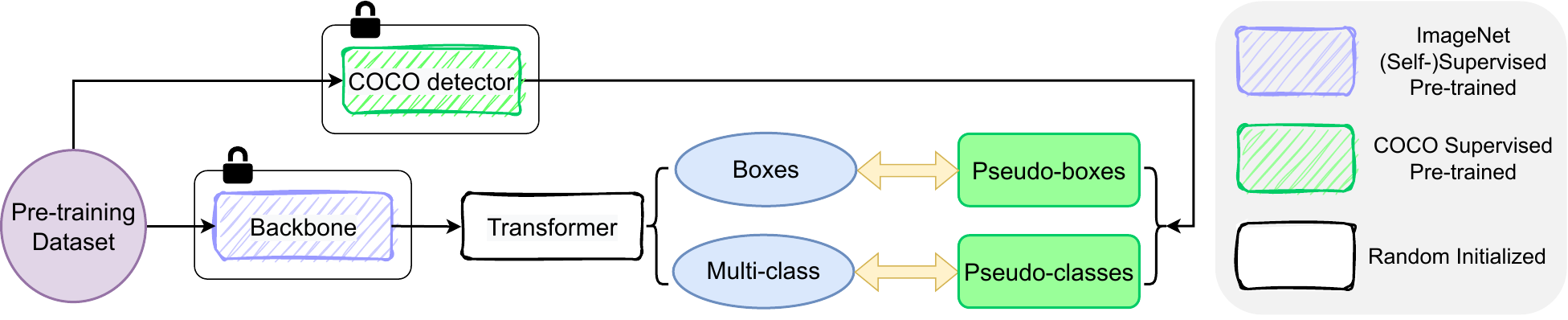}
\caption{\footnotesize{Simple Self-training}}
\label{fig:self_training_pipeline}
\end{subfigure}
\end{minipage}
\caption{\small{\textbf{The pre-training pipelines of DETReg, DETReg+pseudo-box, and Simple Self-training.}
In DETReg and DETReg+pseudo-box, we use an extra frozen backbone branch to get the target object embeddings from the image crops. The binary-class outputs of the Transformer predict whether the detected boxes contain an object.}}
\label{fig:detailed_pipeline}
\end{figure*}

Most self-supervised pre-training methods differentiate through the design of pretext tasks, which results in different structures for the pre-training target $\overline{\mathbf{y}}_{i}$ and implementations of the pre-training loss $\mathcal{L}_{\textrm{pre}}$. A good pretext task design can improve the final prediction performance. In the following, we first introduce the instantiation of a representative method called DETReg~\citep{bar2022detreg}. Then, we propose two more effective pre-training schemes: DETReg + Pseudo-box and Simple Self-training. Both methods focus on enhancing the localization and classification pre-training target quality. We compare the pre-training pipeline of three methods in Figure~\ref{fig:detailed_pipeline}.

\vspace{1mm}
\paragraph{DETReg.}
DETReg uses an unsupervised region proposal method named {\it Selective Search (ss)} to generate the target boxes. The $j$-th ``box proposal'' for the $i$-th input is denoted as $\overline{\mathbf{b}}^{ss}_{ij}\in[0,1]^{4}$.
We select the top $\overline{k}$ Selective Search box proposals $\{\overline{\mathbf{b}}^{ss}_{i1},\cdots,\overline{\mathbf{b}}^{ss}_{i\overline{k}}\}$ and pair them with the binary category target padded to the size of network query number $k$ ($k > \overline{k}$) $\{\overline{\mathbf{p}}^{ss}_{i1},\cdots,\overline{\mathbf{p}}^{ss}_{ik} \vert \overline{\mathbf{p}}^{ss}_{i1},\cdots,\overline{\mathbf{p}}^{ss}_{i\overline{k}}=1,\overline{\mathbf{p}}^{ss}_{i(\overline{k}+1)},\cdots,\overline{\mathbf{p}}^{ss}_{ik}=0\}$, where $\overline{\mathbf{p}}^{ss}_{ij}=1$ indicates the element is a box proposal while $\overline{\mathbf{p}}^{ss}_{ij}=0$ indicates a padded $\varnothing$. To compensate for the lack of semantic information in the binary category, the DETReg network incorporates another object embedding reconstruction branch to predict the object embeddings $\{\overline{\mathbf{f}}_{i1},\cdots,\overline{\mathbf{f}}_{ik} \vert \overline{\mathbf{f}}_{ij} \in \mathbb{R}^{d}\}$ of detected boxes, which is supervised by the target object descriptor $\{\overline{\mathbf{f}}^{\text{swav}}_{i1},\cdots,\overline{\mathbf{f}}^{\text{swav}}_{i\overline{k}}\}$ with $\overline{\mathbf{f}}^{\text{swav}}_{ij}$ indicating the object embedding extracted from the image patch in the $j$-th box proposal on the $i$-th input with a fixed SwAV backbone. Therefore, the pre-training target and network prediction are denoted as Equation~\ref{eq5}:
    
\begin{equation} \label{eq5}
\overline{\mathbf{y}}_{ij}=( \overline{\mathbf{p}}^{ss}_{ij}, \overline{\mathbf{b}}^{ss}_{ij}, \overline{\mathbf{f}}^{\text{swav}}_{ij}),\quad \overline{\mathbf{z}}_{ij}=( \overline{\mathbf{p}}_{ij}, \overline{\mathbf{b}}_{ij}, \overline{\mathbf{f}}_{ij}).
\end{equation}
    
The pre-training loss is the sum of binary classification loss $\mathcal{L}^{\textrm{bin}}_{\textrm{cls}}(\cdot)$, box loss $\mathcal{L}_{\textrm{box}}(\cdot)$, and embedding loss $\mathcal{L}_{\textrm{emb}}(\cdot)$ through all $k$ outputs as below:
    
\begin{equation} \label{eq7}
\begin{split}
\mathcal{L}_{\textrm{pre}}(\overline{\mathbf{y}}_{i}, \overline{\mathbf{z}}_{i}) & = \sum_{j=1}^{k} \lambda_{c}\mathcal{L}^{\textrm{bin}}_{\textrm{cls}}(\overline{\mathbf{p}}^{ss}_{ij}, \overline{\mathbf{p}}_{i\sigma_i(j)}) \\
 & + \lambda_{b}\mathds{1}_{\{ \overline{\mathbf{p}}^{ss}_{ij} \neq 0 \}} \mathcal{L}_{\textrm{box}}(\overline{\mathbf{b}}^{ss}_{ij}, \overline{\mathbf{b}}_{i\sigma_i(j)}) \\
 & + \lambda_{e}\mathcal{L}_{\textrm{emb}}(\overline{\mathbf{f}}^{\text{swav}}_{ij}, \overline{\mathbf{f}}_{i\sigma_i(j)}),
\end{split}
\end{equation}
where $\mathcal{L}^{\textrm{bin}}_{\textrm{cls}}(\cdot)$ is the binary classification loss which can be implemented as Cross Entropy Loss or Focal Loss. $\mathcal{L}_{\textrm{box}}(\cdot)$ is the sum of  L1 and GIoU Loss, and $\mathcal{L}_{\textrm{emb}}(\cdot)$ is the L1 Loss. $\lambda_{c}$, $\lambda_{b}$, and $\lambda_{e}$ are loss coefficients and $\sigma_i(j)$ maps the target box $j$ to the assigned predicted box $\sigma_i(j)$ with lowest cost within the $i$-th input.
	
\vspace{1mm}
\paragraph{DETReg + Pseudo-box.}
The unsupervised box proposals like Selective Search boxes are of very low quality. To handle this, we employ two off-the-shelf well-trained COCO object detectors to predict the pseudo-boxes for the pre-training data to replace the Selective Search proposals. Specifically, we replace the $(\overline{\mathbf{p}}^{ss}_{ij}, \overline{\mathbf{b}}^{ss}_{ij})$ in Equation~\ref{eq5} and~\ref{eq7} with $(\overline{\mathbf{p}}^{\text{pseudo}}_{ij}, \overline{\mathbf{b}}^{\text{pseudo}}_{ij})$. We use $\mathcal{H}$-Deformable-DETR with ResNet50 or Swin-L backbone as our detector network. We first train them on COCO, then use the trained detector to predict pseudo-boxes on the pre-training dataset, and the top~30 predictions are selected as $\overline{k}$. 

\vspace{1mm}
\paragraph{Simple Self-training.}
We further replace the binary category target $\overline{\mathbf{p}}^\text{pseudo}_{ij}$ with category predictions $\overline{\mathbf{c}}^{\text{pseudo}}_{ij}\in\{\varnothing,c_{1},\cdots,c_{n}\}$ of aforementioned COCO object detectors as the classification target and remove $\overline{\mathbf{f}}^{\text{swav}}_{ij}$ since we already have detailed class information. Due to that the detector is trained on COCO and the pseudo-category labels it predicts are the 80 COCO categories, the binary classification turns into a multiclass classification. The formulation is shown below:

\begin{equation} \label{eq9}
\overline{\mathbf{y}}_{ij}=( \overline{\mathbf{c}}^{\text{pseudo}}_{ij}, \overline{\mathbf{b}}^{\text{pseudo}}_{ij}),\quad 
\overline{\mathbf{z}}_{ij}=( \overline{\mathbf{c}}_{ij}, \overline{\mathbf{b}}_{ij}),
\end{equation}
    
\begin{equation} \label{eq10}
\begin{split}
\mathcal{L}_{\textrm{pre}}(\overline{\mathbf{y}}_i, \overline{\mathbf{z}}_{i}) & = \sum_{j=1}^{k} \lambda_{c}\mathcal{L}^{ \textrm{mul}}_{\textrm{cls}}(\overline{\mathbf{c}}^{\text{pseudo}}_{ij}, \overline{\mathbf{c}}_{i\sigma_i(j)}) \\
 & + \lambda_{b}\mathds{1}_{\{ \overline{\mathbf{c}}^{\text{pseudo}}_{ij} \neq \varnothing \}} \mathcal{L}_{\textrm{box}}(\overline{\mathbf{b}}^{\text{pseudo}}_{ij}, \overline{\mathbf{b}}_{i\sigma_i(j)}),
\end{split}
\end{equation}
where $\mathcal{L}^{ \textrm{mul}}_{\textrm{cls}}(\cdot)$ is the multiclass classification loss.

\begin{figure}[t]
\begin{minipage}[t]{0.24\linewidth}
\centering
\begin{subfigure}[b]{\textwidth}
\centering
\includegraphics[width = \textwidth,height=15mm]{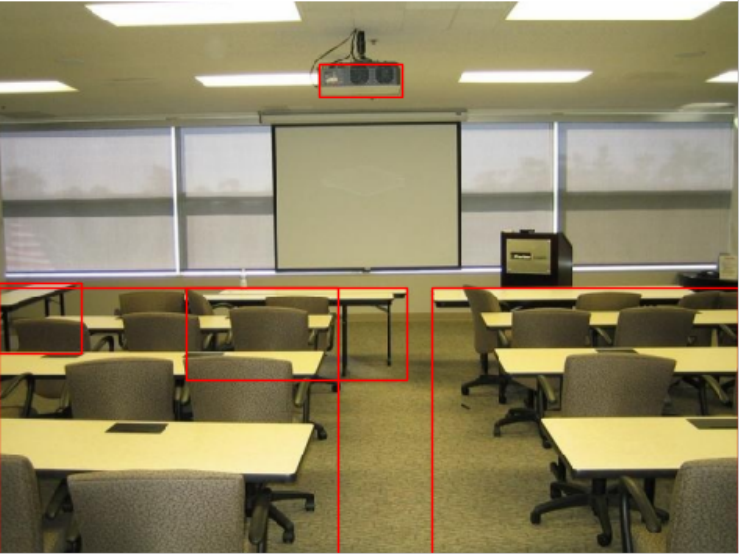}\\\vspace{1mm}
\includegraphics[width = \textwidth,height=15mm]{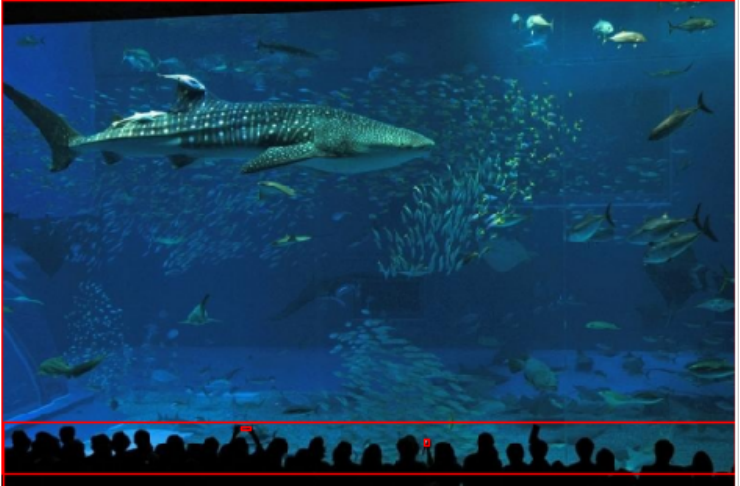}
\caption*{\centering\tiny{Ground-Truth}}
\end{subfigure}
\end{minipage}
\begin{minipage}[t]{0.24\linewidth}
\centering
\begin{subfigure}[b]{\textwidth}
\centering
\includegraphics[width = \textwidth,height=15mm]{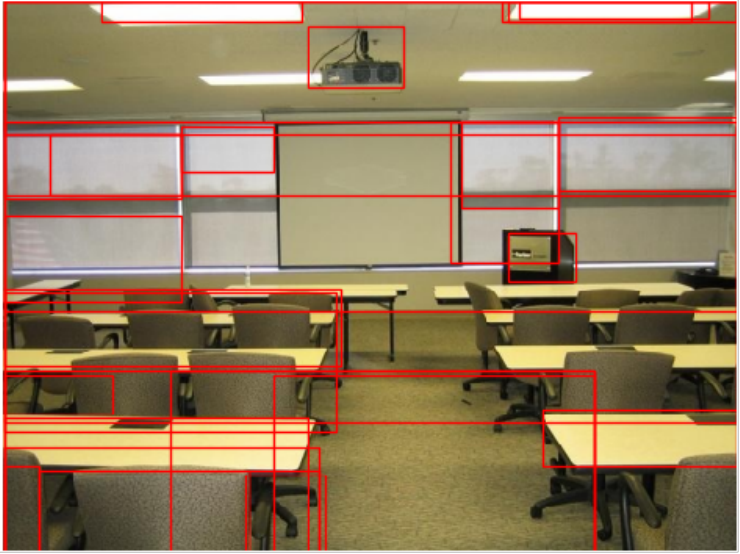}\\\vspace{1mm}
\includegraphics[width = \textwidth,height=15mm]{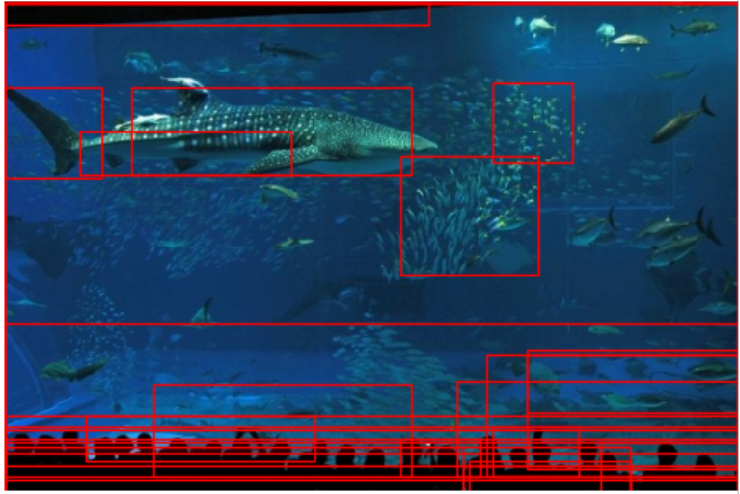}
\caption*{\centering\tiny{Selective Search}}
\end{subfigure}
\end{minipage}
\begin{minipage}[t]{0.24\linewidth}
\centering
\begin{subfigure}[b]{\textwidth}
\centering
\includegraphics[width = \textwidth,height=15mm]{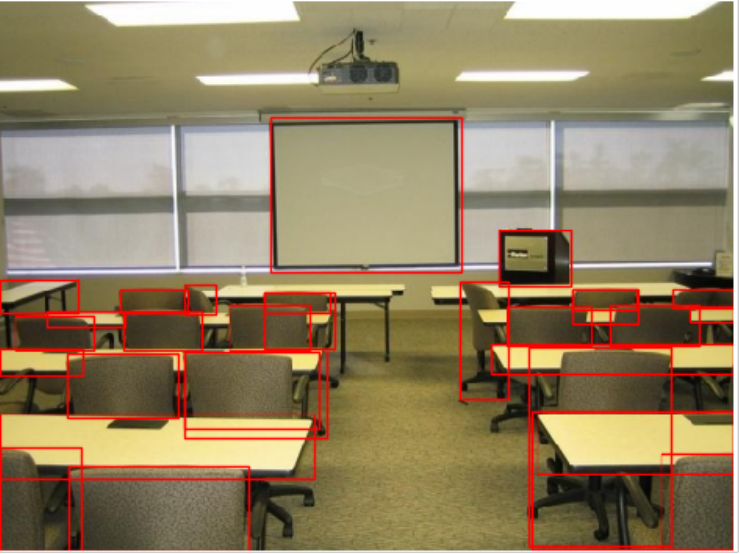}\\\vspace{1mm}
\includegraphics[width = \textwidth,height=15mm]{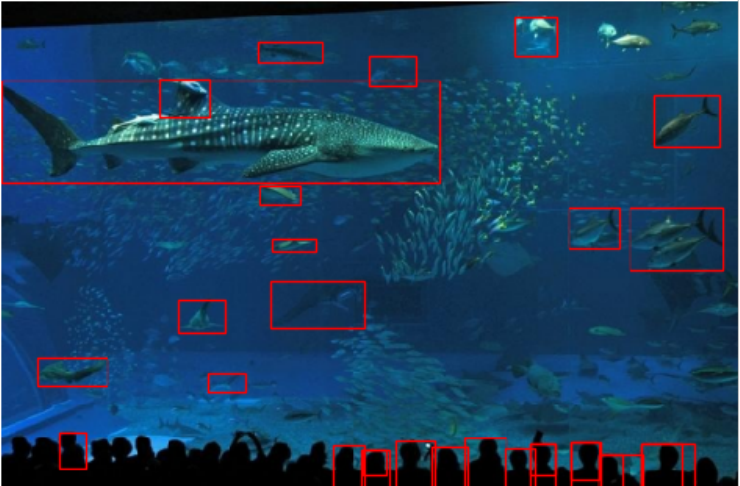}
\caption*{\centering\tiny{$\mathcal{H}$-Def-DETR + R50}}
\end{subfigure}
\end{minipage}
\begin{minipage}[t]{0.24\linewidth}
\centering
\begin{subfigure}[b]{\textwidth}
\centering
\includegraphics[width = \textwidth,height=15mm]{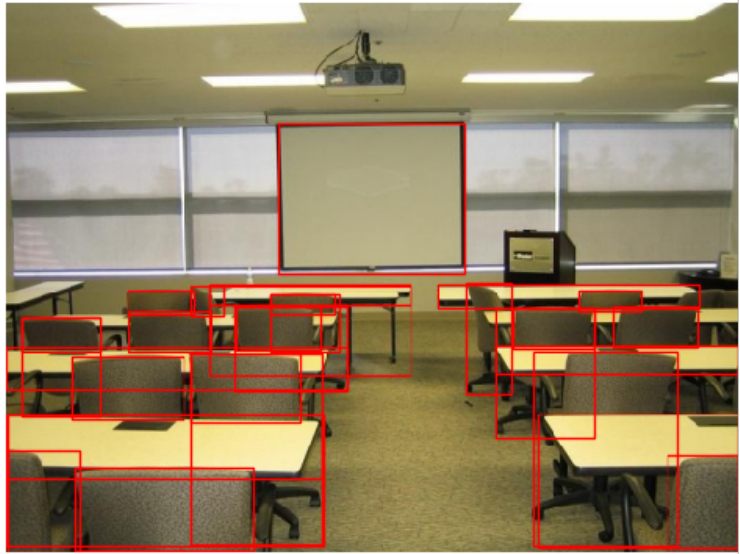}\\\vspace{1mm}
\includegraphics[width = \textwidth,height=15mm]{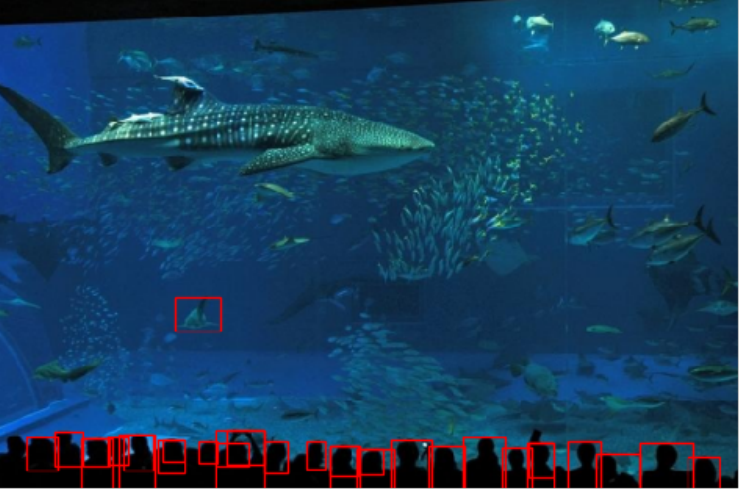}
\caption*{\centering\tiny{$\mathcal{H}$-Def-DETR + Swin-L}}
\end{subfigure}
\end{minipage}
\caption{\small{\textbf{Qualitative comparisons of the top $30$ generated bounding boxes of different methods on Objects$365$.} The methods include Selective Search and trained $\mathcal{H}$-Deformable-DETR detectors with R$50$ or Swin-L backbones.}}
\label{fig:compare_bbox_quality}
\end{figure}

\vspace{-5mm}
\textbf{\subsection{Discussion}}
\label{Discussion}
We utilize ImageNet and Objects365 as the two pre-training benchmarks. To display the quality of Selective Search proposals and pseudo-boxes generated by two off-the-shelf COCO object detectors, we report their boxes' Average Precision and Average Recall on Objects365 validation set in Table~\ref{tab:ap_ar_result}. As can be seen, pseudo-boxes generated by COCO object detectors are far more accurate than Selective Search boxes. We also visualize their box proposals in Figure~\ref{fig:compare_bbox_quality}.

Unlike the conventional self-training scheme~\citep{zoph2020rethinking,xie2020self} that relies on applying complicated augmentation strategy to boost the quality of pseudo-labels, adjusting NMS threshold carefully, and re-generating more accurate pseudo-labels based on the fine-tuned models in an iterative manner, our Simple Self-training method directly generate the pseudo-labels for one time without those tricks, resulting in a much simpler approach.
	
\begin{table}[!t]
\centering\setlength{\tabcolsep}{2pt}
\footnotesize
\renewcommand{\arraystretch}{1.35}
\resizebox{1\linewidth}{!}
{
\begin{tabular}{l|l|c|c|c|c|c|c|c|c}
    Localization method & AP & AP$_{50}$ & AP$_{75}$ & AP$_{S}$ & AP$_{M}$ & AP$_{L}$ & AR@$10$ & AR@$30$ & AR@$100$ \\ \shline
    Selective Search & $0.5$ & $1.6$ & $0.2$ & $0.2$ & $0.3$ & $1.2$ & $3.7$ & $8.3$ & $15.5$ \\
    $\mathcal{H}$-Deformable-DETR + R$50$ & $28.4$ & $40.4$ & $30.2$ & $12.7$ & $26.7$ & $43.1$ & $26.5$ & $37.4$ & $\bf{47.7}$ \\ 
    $\mathcal{H}$-Deformable-DETR + Swin-L  & $\bf{30.7}$ & $\bf{41.3}$ & $\bf{33.0}$ & $\bf{15.2}$ & $\bf{29.0}$ & $\bf{44.9}$ & $\bf{28.1}$ & $\bf{38.5}$ & $47.4$
\end{tabular}
}
\caption{\small{Objects$356$ AP and AR score for Selective Search box proposals, and pseudo-box predictions of $\mathcal{H}$-Deformable-DETR-based COCO detectors with R50 and Swin-L backbone.}}
\label{tab:ap_ar_result}
\vspace{-1mm}
\end{table}

%%%%%%%%%%%%%%%%%%%%%%%%%%%%%%%%
\vspace{5mm}
\section{Experiment}
\textbf{\subsection{Implementation Details}}
%%%%%%%%%%%%%%%%%%%%%%%%%%%%%%%%	
\paragraph{Datasets.} Our object detection network is pre-trained on the ImageNet or Objects$365$~\citep{shao2019objects365} benchmark, then fine-tuned on COCO train2017 and evaluated on COCO val2017, or fine-tuned on PASCAL VOC trainval07+12 and evaluated on PASCAL VOC test2007. For the pre-training benchmarks, ImageNet has $1.2$ Million images which mostly contain one object since the dataset is created for classification. Objects$365$ is a large-scale dataset for object detection with $2$ Million images. The image scene is more complicated with around $15$ ground-truth bounding boxes per image on average. We use Objects$365$ as the default pre-training benchmark for all experiments in sections~\ref{section:sota} and~\ref{section:ablate}, as its complex scenes bring better pre-training performance for the Simple Self-training approach.

\vspace{1mm}
\paragraph{Architectures.} We use two kinds of DETR backbones including ResNet$50$ which is self-supervised pre-trained by SwAV on ImageNet and Swin-L which is supervised pre-trained on ImageNet. We pre-train three DETR-based architectures in Section~\ref{section:other_detr} including vanilla DETR~\citep{carion2020end}, Deformable-DETR~\citep{zhu2020deformable}, and $\mathcal{H}$-Deformable-DETR~\citep{jia2023detrs}, which is a recent state-of-the-art object detector based on a combination of an improved Deformable-DETR and an effective hybrid matching scheme. The Transformer module in those architectures is composed of 6 encoder layers and 6 decoder layers. The vanilla DETR and Deformable-DETR are plain without tricks, while $\mathcal{H}$-Deformable-DETR is improved with iterative bounding box refinement, two-stage~\citep{zhu2020deformable}, mixed query selection, and look forward twice scheme~\citep{zhang2022dino}. By default, we use $\mathcal{H}$-Deformable-DETR with ResNet$50$ backbone for the ablation study.

\begin{table}[t]
\begin{minipage}[t]{1\linewidth}
\providecommand{\blue}[1]{\textcolor{blue}{#1}}
    \definecolor{deepgreen}{rgb}{0.07, 0.53, 0.03}
    \centering
    \setlength{\tabcolsep}{1pt}
    \renewcommand{\arraystretch}{1.5}
    \footnotesize
    \resizebox{1\linewidth}{!}
    {
        \begin{tabular}{l|c|c|c|cccccc}
            Method     & Framework & Backbone & \#epoch & AP & AP$_{50}$ & AP$_{75}$ & AP$_{S}$ & AP$_{M}$ & AP$_{L}$\\
            \shline
            Swin ~\citep{liu2021swin} & HTC & Swin-L & $36$ &  {$57.1$} & $75.6$ & $62.5$ & $42.4$ & $60.7$ & $71.1$\\
            Group-DETR~\citep{chen2022group}  & DETR & Swin-L & $36$ & {$58.4$} & - & - & {$41.0$} & {$62.5$} & {$73.9$} \\
            DINO-DETR~\citep{zhang2022dino}  & DETR & Swin-L & $36$ & {$58.5$} & {$77.0$} & {$64.1$} & {$41.5$} & {$62.3$} & $\bf{74.0}$ \\
            $\mathcal{H}$-Deformable-DETR~\citep{jia2023detrs} & DETR & Swin-L  & $36$  & {$57.9$} & $76.9$ & $63.7$ & $42.4$ & $61.9$ & $73.4$ \\\hline
            Ours (pre-trained $\mathcal{H}$-Deformable-DETR) & DETR & Swin-L & $24$  & $\bf{59.3}$ & $\bf{77.9}$ & $\bf{65.1}$ & $\bf{44.1}$ & $\bf{62.9}$ & $73.6$ \\
        \end{tabular}
    }
    \caption{\small{System-level comparisons with the state-of-the-art DETR-based single-scale evaluation results on COCO \texttt{val} set.}
    \label{tab:compare_to_sota_2ddet_coco}
    }
\end{minipage}
\end{table}

\begin{table}[t]
\begin{minipage}[t]{1\linewidth}
\providecommand{\blue}[1]{\textcolor{blue}{#1}}
    \definecolor{deepgreen}{rgb}{0.07, 0.53, 0.03}
    \centering
    \setlength{\tabcolsep}{1pt}
    \renewcommand{\arraystretch}{1.5}
    \footnotesize
    \resizebox{1\linewidth}{!}
    {
        \begin{tabular}{l|c|c|c|c|cccccc}
            Method & DETR model & Pretrain & \#query & \#epoch & $\operatorname{AP}$ & $\operatorname{AP}_{50}$ & $\operatorname{AP}_{75}$ & $\operatorname{AP}_{S}$ & $\operatorname{AP}_{M}$ & $\operatorname{AP}_{L}$ \\
            \shline
            \textit{from scratch} & DETR & - & $100$ & $150$ & $40.3$ & $61.3$ & $42.2$ & $18.2$ & $44.6$ & $60.5$ \\
            DETReg  & DETR & ImageNet & $100$ & $150$ & $40.2$ & $60.7$ & $42.3$ & $17.6$ & $44.3$ & $59.6$ \\
            ours & DETR & ImageNet & $100$ & $150$ & $\bf{41.9}$ & $\bf{62.7}$ & $\bf{44.0}$ & $\bf{20.7}$ & $\bf{46.0}$ & $\bf{62.8}$ \\
            \hline
            \textit{from scratch} & DDETR-MS & - & $300$ & $50$ & $45.2$ & $64.2$ & $49.4$ & $\bf{27.2}$ & $49.3$ & $59.1$ \\
            DETReg  & DDETR-MS & ImageNet & $300$ & $50$ & $43.5$ & $61.4$ & $47.3$ & $24.2$ & $47.1$ & $58.7$ \\
            ours & DDETR-MS & ImageNet & $300$ & $50$ & $\bf{46.0}$ & $\bf{64.4}$ & $\bf{50.0}$ & $26.6$ & $\bf{49.8}$ & $\bf{61.5}$ \\
            \hline
            \textit{from scratch} & $\mathcal{H}$-DDETR-MS & - & $300$ & $12$ & $49.6$ & $67.5$ & $54.1$ & $31.9$ & $53.3$ & $64.1$ \\
            DETReg  & $\mathcal{H}$-DDETR-MS & ImageNet & $300$ & $12$ & $49.5$ & $66.8$ & $53.9$ & $30.5$ & $53.5$ & $63.6$ \\
            ours & $\mathcal{H}$-DDETR-MS & ImageNet & $300$ & $12$ & $\bf{51.6}$ & $\bf{69.4}$ & $\bf{56.4}$ & $\bf{35.0}$ & $\bf{55.3}$ & $\bf{66.8}$ \\
        \end{tabular}
    }
    \caption{\small{Comparisons with self-supervised pre-training method DETReg on the COCO downstream benchmark.}
    \label{tab:compare_unsupervised_coco}
    }
\end{minipage}
\end{table}

\begin{table}[t]
\begin{minipage}[t]{1\linewidth}
\providecommand{\blue}[1]{\textcolor{blue}{#1}}
    \definecolor{deepgreen}{rgb}{0.07, 0.53, 0.03}
    \centering
    \setlength{\tabcolsep}{1pt}
    \renewcommand{\arraystretch}{1.5}
    \footnotesize
    \resizebox{1\linewidth}{!}
    {
        \begin{tabular}{l|c|c|c|c|cccccc}
            Method & DETR model & Pretrain & \#query & \#epoch & $\operatorname{AP}$ & $\operatorname{AP}_{50}$ & $\operatorname{AP}_{75}$ & $\operatorname{AP}_{S}$ & $\operatorname{AP}_{M}$ & $\operatorname{AP}_{L}$ \\
            \shline
            \textit{from scratch} & DETR & - & $100$ & $150$ & $56.3$ & $80.3$ & $60.6$ & $10.2$ & $36.0$ & $65.9$ \\
            DETReg  & DETR & ImageNet & $100$ & $150$ & $60.9$ & $82.0$ & $65.9$ & $15.1$ & $40.8$ & $69.8$ \\
            ours & DETR & ImageNet & $100$ & $150$ & $\bf{63.5}$ & $\bf{83.8}$ & $\bf{68.6}$ & $\bf{22.5}$ & $\bf{44.3}$ & $\bf{72.1}$ \\
            \hline
            \textit{from scratch} & DDETR-MS & - & $300$ & $50$ & $61.1$ & $83.1$ & $68.0$ & $25.5$ & $47.4$ & $67.7$ \\
            DETReg  & DDETR-MS & ImageNet & $300$ & $50$ & $63.6$ & $82.6$ & $70.2$ & $27.5$ & $49.7$ & $70.2$ \\
            ours & DDETR-MS & ImageNet & $300$ & $50$ & $\bf{67.8}$ & $\bf{85.4}$ & $\bf{75.5}$ & $\bf{30.9}$ & $\bf{54.7}$ & $\bf{74.4}$ \\
            \hline
            \textit{from scratch} & $\mathcal{H}$-DDETR-MS & - & $300$ & $12$ & $63.8$ & $82.4$ & $70.0$ & $26.5$ & $50.0$ & $70.4$ \\
            DETReg  & $\mathcal{H}$-DDETR-MS & ImageNet & $300$ & $12$ & $67.7$ & $84.5$ & $74.9$ & $\bf{35.1}$ & $55.1$ & $74.7$ \\
            ours & $\mathcal{H}$-DDETR-MS & ImageNet & $300$ & $12$ & $\bf{71.6}$ & $\bf{87.0}$ & $\bf{79.2}$ & $33.1$ & $\bf{60.3}$ & $\bf{78.2}$ \\
        \end{tabular}
    }
    \caption{\small{Comparisons with self-supervised pre-training method DETReg on the PASCAL VOC downstream benchmark.}
    \label{tab:compare_unsupervised_pascal}
    }
\end{minipage}
\end{table}

\begin{table}
\begin{minipage}[t]{1\linewidth}
\centering
\renewcommand{\arraystretch}{1.25}
\footnotesize
\resizebox{\columnwidth}{!}{
\begin{tabular}{l|c|cccccc}
    Method & Pre-training dataset & $\operatorname{AP}$ & $\operatorname{AP}_{50}$ & $\operatorname{AP}_{75}$ & $\operatorname{AP}_{S}$ & $\operatorname{AP}_{M}$ & $\operatorname{AP}_{L}$  \\
    \shline
    \multirow{2}*{DETReg} & ImageNet & $\bf{49.5}$ & $\bf{66.8}$ & $\bf{53.9}$ & $30.5$ & $\bf{53.5}$ & $\bf{63.6}$ \\
    & O365 & $49.2$ & $66.5$ & $53.6$ & $\bf{31.4}$ & $53.2$ & $63.5$ \\
    \hline
    \multirow{2}*{DETReg+pseudo-box} & ImageNet & $50.9$ & $68.3$ & $55.7$ & $33.6$ & $54.6$ & $64.9$ \\ 
    & O$365$  & $\bf{52.0}$ & $\bf{69.6}$ & $\bf{56.7}$ & $\bf{36.1}$ & $\bf{55.9}$ & $\bf{65.3}$ \\ 
    \hline
    \multirow{2}*{Simple Self-training} & ImageNet & $51.6$ & $69.4$ & $56.4$ & $35.0$ & $55.3$ & $66.8$ \\ 
    & O$365$  & $\bf{52.8}$ & $\bf{70.9}$ & $\bf{57.6}$ & $\bf{37.0}$ & $\bf{56.6}$ & $\bf{67.3}$ \\
\end{tabular}
}
\caption{\small{Effect of pre-training dataset choices}.}
\label{tab:ablate_pretrain_dataset}
\end{minipage}
\end{table}

\begin{table}[t]
\begin{minipage}[t]{1\linewidth}
\providecommand{\blue}[1]{\textcolor{blue}{#1}}
    \definecolor{deepgreen}{rgb}{0.07, 0.53, 0.03}
    \centering
    \setlength{\tabcolsep}{0.25pt}
    \renewcommand{\arraystretch}{1.7}
    \footnotesize
    \resizebox{1\linewidth}{!}
    {
    \begin{tabular}{l|c|c|cccccc}
       Method & Localization target & Classification target & $\operatorname{AP}$ & $\operatorname{AP}_{50}$ & $\operatorname{AP}_{75}$ & $\operatorname{AP}_{S}$ & $\operatorname{AP}_{M}$ & $\operatorname{AP}_{L}$  \\
        \shline
        \textit{from scratch} & - & - & $49.6$ & $67.5$ & $54.1$ & $31.9$ & $53.3$ & $64.1$ \\
        DETReg & Selective Search & Object-embedding loss & $49.2$ & $66.5$ & $53.6$ & $31.4$ & $53.2$ & $63.5$ \\ 
        DETReg+pseudo-box & Pseudo-box prediction  & Object-embedding loss & $52.0$ & $69.6$ & $56.7$ & $36.1$ & $55.9$ & $65.3$ \\
        Simple Self-training & Pseudo-box prediction  & Pseudo-class prediction  & $52.8$ & $70.9$ & $57.6$ & $37.0$ & $56.6$ & $67.3$ \\ 
        Supervised & Ground-truth & Ground-truth & $\bf{53.2}$ & $\bf{71.5}$ & $\bf{58.1}$ & $\bf{37.3}$ & $\bf{57.0}$ & $\bf{67.4}$ \\
    \end{tabular}
    }
    \caption{\small{Fine-tuning results on COCO after pre-training with different methods using various localization and classification pre-training targets on Objects365.
    \label{tab:pretrain_target_coco}
    }}
\end{minipage}
\end{table}

\begin{table}
    \centering
    \setlength{\tabcolsep}{12pt}
    \renewcommand{\arraystretch}{1.25}
    \resizebox{\columnwidth}{!}{
    \begin{tabular}{l|cccccc}
       Method & $\operatorname{AP}$ & $\operatorname{AP}_{50}$ & $\operatorname{AP}_{75}$ & $\operatorname{AP}_{S}$ & $\operatorname{AP}_{M}$ & $\operatorname{AP}_{L}$   \\
        \shline
       \textit{from scratch} & $63.8$ & $82.4$ & $70.0$ & $26.5$ & $50.0$ & $70.4$ \\ 
       DETReg & $67.7$ & $84.7$ & $74.1$ & $34.8$ & $55.9$ & $74.3$ \\ 
       DETReg+pseudo-box & $71.6$ & $87.0$ & $79.1$ & $36.1$ & $59.0$ & $77.9$  \\ 
       Simple Self-training & $71.6$ & $87.9$ & $79.7$ & $33.5$ & $60.2$ & $\bf{78.7}$ \\ 
       Supervised & $\bf{72.6}$ & $\bf{88.0}$ & $\bf{80.7}$ & $\bf{37.6}$ & $\bf{62.6}$ & $78.6$ \\ 
    \end{tabular}
    }
    \caption{\small{Fine-tuning results on PASCAL VOC after pre-training with different methods on Objects365.}}
    \label{tab:pretrain_target_voc}
\end{table}

\begin{table}
\begin{minipage}[t]{1\linewidth}
\centering
\setlength{\tabcolsep}{13pt}
\renewcommand{\arraystretch}{1.25}
\footnotesize
    \resizebox{\columnwidth}{!}{
    \begin{tabular}{l|cccccc}
       \#pseudo-box & $\operatorname{AP}$ & $\operatorname{AP}_{50}$ & $\operatorname{AP}_{75}$ & $\operatorname{AP}_{S}$ & $\operatorname{AP}_{M}$ & $\operatorname{AP}_{L}$  \\
        \shline 
       $30$ & $\bf{52.0}$ & $\bf{69.6}$ & $\bf{56.7}$ & $\bf{36.1}$ & $\bf{55.9}$ & $65.3$ \\ 
       $60$   & $51.6$ & $69.1$ & $56.6$ & $34.8$ & $55.4$ & $\bf{65.5}$ \\ 
       $100$   & $51.5$ & $68.9$ & $56.3$ & $34.9$ & $54.7$ & $65.4$ \\ 
    \end{tabular}
    }
    \caption{\small{Ablation experiments on the number of pseudo-boxes for the DETReg+pseudo-box method.}}
    \label{tab:ablate_box_num}
\end{minipage}
\end{table}

\begin{table}
    \centering
    \resizebox{\columnwidth}{!}{
    \begin{tabular}{l|c|c|cccccc}
        Method & Encoder & Decoder & $\operatorname{AP}$ & $\operatorname{AP}_{50}$ & $\operatorname{AP}_{75}$ & $\operatorname{AP}_{S}$ & $\operatorname{AP}_{M}$ & $\operatorname{AP}_{L}$  \\
        \shline
        \multirow{3}*{DETReg+pseudo-box} & \checkmark & \checkmark & $\bf{52.0}$ & $\bf{69.6}$ & $\bf{56.7}$ & $\bf{36.1}$ & $\bf{55.9}$ & $\bf{65.3}$ \\ 
        & & \checkmark & $49.4$ & $67.1$ & $53.5$ & $32.0$ & $53.2$ & $63.2$ \\ 
        & \checkmark & & $51.5$ & $\bf{69.6}$ & $56.1$ & $35.4$ & $55.3$ & $\bf{65.5}$ \\ 
        \hline
        \multirow{3}*{Simple Self-training} & \checkmark & \checkmark & $\bf{52.8}$ & $\bf{70.9}$ & $\bf{57.6}$ & $\bf{37.0}$ & $\bf{56.6}$ & $\bf{67.3}$ \\ 
        & & \checkmark & $50.2$ & $68.2$ & $54.3$ & $32.4$ & $54.1$ & $63.6$ \\
        & \checkmark & & $51.8$ & $69.6$ & $56.4$ & $34.9$ & $55.4$ & $66.6$ \\
    \end{tabular}
    }
    \caption{\small{Effect of Transformer encoder or decoder pre-training.}}
    \label{tab:ablate_enc_dec}
\end{table}

\vspace{1mm}
\paragraph{Training.} We pre-train the network on ImageNet for $5$ epochs following DETReg or on Objects$365$ for $3$ epochs to ensure the same iteration number according to their different dataset sizes. For fine-tuning, we train for 150 epochs with vanilla DETR, 50 epochs with Deformable-DETR, and 12 epochs with $\mathcal{H}$-Deformable-DETR, or 24 epochs with $\mathcal{H}$-Deformable-DETR in Section~\ref{section:sota} for better performance. The learning rate drops at $120/150$, $40/50$, $11/12$, and $20/24$ respectively. The batch size for pre-training and fine-tuning are both 16. 

\vspace{1mm}
\paragraph{Metrics.} We measure the object detection accuracy for the top $100$ detected bounding boxes. Specifically, we compute $\operatorname{AP}$, $\operatorname{AP}_{50}$ and $\operatorname{AP}_{75}$ as the average precision when using IoU thresholds across the range of $0.50$ to $0.95$, and exactly of $0.50$ or $0.75$; and also $\operatorname{AP}_{S}$, $\operatorname{AP}_{M}$, $\operatorname{AP}_{L}$ as the AP for small, medium, large bounding boxes.

\vspace{-5mm}
\textbf{\subsection{Comparison to the State-of-the-art}\label{section:sota}}
Table~\ref{tab:compare_to_sota_2ddet_coco} shows the object detection result on COCO validation set of $\mathcal{H}$-Deformable-DETR network pre-trained on Objects$365$ benchmark with our method in comparison with other state-of-the-art object detection systems. Our Simple Self-training approach significantly boosts the performance of $\mathcal{H}$-Deformable-DETR from $57.9\%$ to $59.3\%$ with fewer training epochs. We expect our approach to achieve better results with bigger batch size and epoch number, for instance, batch size of 256 and epoch of 60 are used for the self-supervised pre-training in Siamese DETR~\citep{huang2023siamese}.

\textbf{\subsection{Results on different DETR architectures}\label{section:other_detr}}

As shown in Table~\ref{tab:compare_unsupervised_coco} and Table~\ref{tab:compare_unsupervised_pascal}, we display the results of DETReg and Simple Self-training with different DETR architectures on the COCO and PASCAL VOC benchmarks. The line of \textit{from scratch} shows the baseline results without pre-training as the ResNet$50$ backbone is initialized with SwAV and the Transformer is randomly initialized. The results show that with the reported experiment setting, the DETReg pre-training fails to bring improvement to the \textit{from scratch} baseline on the COCO benchmark, while can get small gains on the PASCAL VOC benchmark. Our Simple Self-training can effectively improve the baseline performance for all three DETR architectures on both benchmarks.

\textbf{\subsection{Ablation Experiments and Analysis}\label{section:ablate}}
    
\paragraph{Choice of pre-training dataset.} \label{subsection:dataset} 
We also investigate the impact of pre-training datasets with the $\mathcal{H}$-Deformable-DETR architecture in Table~\ref{tab:ablate_pretrain_dataset}. Compared to ImageNet, pre-training with the Objects$365$ benchmark yields better performance with the DETReg+pseudo-box and Simple Self-training approach, which is not the case with the DETReg approach. As DETReg+pseudo-box and Simple Self-training employ accurate pseudo-boxes of COCO detectors as the pre-training targets, they can benefit from a more complex image scene that contains richer objects like Objects$365$, while Selective Search's chaotic proposals on Objects$365$ may not be better localization targets than its proposals on ImageNet. 
It has been proved that ImageNet is a good benchmark for pre-training general representation ability, which can be transferred to multiple downstream tasks. However, for pre-training a specific detection network, a large-scale object detection benchmark like Objects$365$ is more helpful if the pseudo-box has good quality. Therefore, we use Objects$365$ as the default pre-training benchmark for the following studies.
    
\begin{figure}[t]
\centering
\includegraphics[width=0.8\linewidth]{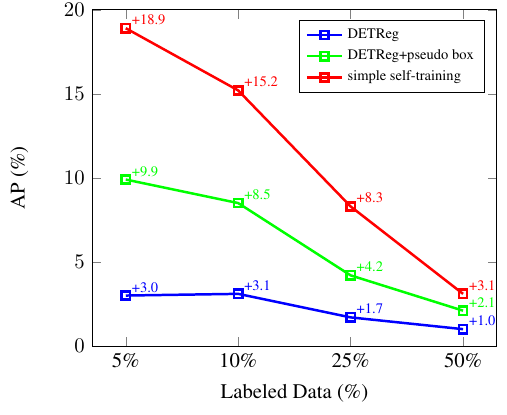}
\caption{\small{\textbf{Ablation experiments on low-data regimes.} The value shows the performance improvement of three pre-training schemes compared to the \textit{from scratch} baseline.}
}
\label{fig:low_data}
\end{figure}

\begin{table*}[t]
\begin{minipage}[t]{1\linewidth}
\providecommand{\blue}[1]{\textcolor{blue}{#1}}
    \definecolor{deepgreen}{rgb}{0.07, 0.53, 0.03}
    \centering
    \setlength{\tabcolsep}{5pt}
    \renewcommand{\arraystretch}{2}
    \footnotesize
    \resizebox{1\linewidth}{!}
    {
        \begin{tabular}{l|c|c|c|c|ccc|cccc}
            \multirow{2}{*}{Text prompt} & \multirow{2}{*}{Gernerative model} & \multirow{2}{*}{Pretraining dataset} & \multirow{2}{*}{Localization target}  & \multirow{2}{*}{Classification target}  & \multicolumn{3}{c|}{COCO} & \multicolumn{3}{c}{PASCAL VOC} \\ \cline{6-11} 
            & & & & & $\operatorname{AP}$ & $\operatorname{AP}_{50}$ & $\operatorname{AP}_{75}$ & $\operatorname{AP}$ & $\operatorname{AP}_{50}$ & $\operatorname{AP}_{75}$ \\
            \shline
            - & - & O$365$ & Pseudo-box prediction & Pseudo-class prediction   & $52.8$ & $70.9$ & $57.6$ & $71.6$ & $\bf{87.9}$ & $79.7$  \\\hline
            COCO captions & ControlNet & Control-COCO 2M & Ground-truth & Ground-truth & $51.1$ &	$69.2$ &	$55.8$ & $71.7$ &	$87.8$ &	$79.2$ \\
            COCO captions & ControlNet & Control-COCO 2M & Pseudo-box prediction & Pseudo-class prediction & $52.6$ &	$70.6$ &	$57.5$ & $72.0$ &	$87.8$ &	$\bf{80.4}$ \\ \hline
            LLaVA captions & ControlNet & LLaVAControl-COCO 2M & Ground-truth & Ground-truth & $50.7$ &	$69.6$ &	$55.4$ &	 $71.6$ &	$87.5$ &	$79.5$ \\
            LLaVA captions & ControlNet & LLaVAControl-COCO 2M & Pseudo-box prediction & Pseudo-class prediction & $\bf{52.9}$ &	$70.8$ &	$57.9$ & $\bf{72.3}$ &	$87.7$ &	$\bf{80.4}$ \\\hline
            COCO captions & SDXL & SDXL-COCO 2M & Pseudo-box prediction & Pseudo-class prediction & $52.5$ &	$70.7$ & 	$57.3$ &	$72.1$ &	$87.6$ & 	$79.7$ \\
            LLaVA captions & SDXL &  SDXL-COCO 2M & Pseudo-box prediction & Pseudo-class prediction & $\bf{52.9}$ &	$\bf{71.0}$ &	$\bf{58.0}$ &	$72.0$ &	$87.6$ &	$80.1$	  \\
        \end{tabular}
    }
    \caption{\small{Evaluation results of pre-training with synthetic images similar to COCO generated by text-to-image generative models ControlNet and SDXL. The text prompts given to the generative models are COCO ground-truth captions (represented as COCO captions) or the generated captions by the large multimodal model LLaVA based on COCO images (represented as LLaVA captions).}
    \label{tab:result_of_synthetic_images}
    }
\end{minipage}
\end{table*}

\begin{table*}[t]
\begin{minipage}[t]{1\linewidth}
\providecommand{\blue}[1]{\textcolor{blue}{#1}}
    \definecolor{deepgreen}{rgb}{0.07, 0.53, 0.03}
    \centering
    \setlength{\tabcolsep}{5pt}
    \renewcommand{\arraystretch}{2}
    \footnotesize
    \resizebox{1\linewidth}{!}
    {
        \begin{tabular}{l|c|c|c|c|ccc|cccc}
            \multirow{2}{*}{Text prompt} & \multirow{2}{*}{Gernerative model} & \multirow{2}{*}{Pretraining dataset} & \multirow{2}{*}{Localization target}  & \multirow{2}{*}{Classification target}  & \multicolumn{3}{c|}{COCO} & \multicolumn{3}{c}{PASCAL VOC} \\ \cline{6-11} 
            & & & & & $\operatorname{AP}$ & $\operatorname{AP}_{50}$ & $\operatorname{AP}_{75}$ & $\operatorname{AP}$ & $\operatorname{AP}_{50}$ & $\operatorname{AP}_{75}$ \\
            \shline
            LLaVA captions & ControlNet & LLaVAControl-O$365$ 2M & Pseudo-box prediction & Pseudo-class prediction & $52.4$ &	$70.5$ & 	$57.2$ &	$71.8$ &	$87.6$ & 	$79.8$ \\
            LLaVA captions & SDXL &  SDXL-O$365$ 2M & Pseudo-box prediction & Pseudo-class prediction & $52.6$ &	$70.6$ &	$57.6$ &	$71.6$ &	$87.4$ &	$79.3$	
        \end{tabular}
    }
    \caption{\small{Evaluation results of pre-training with synthetic images similar to Objects$365$ generated by ControlNet and SDXL. Since Objects$365$ does not have ground-truth captions, the text prompts given to the generative models are generated captions by LLaVA based on Objects$365$ images (represented as LLaVA captions).}
    \label{tab:result_of_synthetic_images_o365}
    }
\end{minipage}
\end{table*}
    
\vspace{1mm}
\paragraph{Pre-training methods.} We present the downstream results on the COCO benchmark of the \textit{from scratch} and different pre-training methods in Table~\ref{tab:pretrain_target_coco} and results on the PASCAL VOC benchmark in Table~\ref{tab:pretrain_target_voc}. All methods except \textit{from scratch} are pre-trained on the Objects$365$ benchmark. The middle three pre-training methods do not utilize Objects$365$ ground-truth labels, while the last is supervised by ground-truth and thereby serves as an upper bound. 
The difference between the three unsupervised pre-training pipelines is illustrated in Figure~\ref{fig:detailed_pipeline}. As shown in the Table~\ref{tab:pretrain_target_coco} and~\ref{tab:pretrain_target_voc}, the DETReg+pseudo-box method builds upon DETReg and improves its localization targets by utilizing more accurate COCO detector pseudo-boxes, leading to significant improvement.
The Simple Self-training method discards the object-embedding loss and instead supervises the multi-class classification head with the class predictions of COCO detectors, resulting in further performance gains. 
For the supervised method, we replace the pseudo-box and pseudo-class targets in the Simple Self-training with ground-truth and achieve an upper-bound performance that is only slightly better than our Simple Self-training strategy. This step-by-step comparison demonstrates how we can progressively improve the pre-training performance by introducing better localization and classification targets. Additionally, we observe that better localization pre-training targets are more impactful than better classification targets for object detection tasks.

\vspace{1mm}
\paragraph{Pseudo-box Number.} In Table~\ref{tab:ablate_box_num}, we ablate with the number of pseudo-boxes in the DETReg + pseudo-box method. We observe that using more than 30 pseudo-boxes for pre-training does not improve the performance, despite more pseudo-boxes exhibiting higher recall on the ground-truth (as shown in Table~\ref{tab:ap_ar_result}, where AR@10, 30, 100 means AR with 10, 30, 100 proposed boxes) and providing more supervision signals. A possible explanation is that each Objects365 image contains approximately 15 box annotations, and the predictions beyond the top 30 may have low confidence and less meaningful information, as a result of incorporating noise into the pseudo-box target.

\vspace{1mm}
\paragraph{Encoder/Decoder Pre-training.} We evaluate the importance of Transformer encoder and decoder pre-training in the DETReg+pseudo-box and Simple Self-training approaches in Table~\ref{tab:ablate_enc_dec}. We first report the performance of using both the encoder and decoder pre-trained parameters, then we report the results of only loading the encoder or decoder pre-trained parameters and random initializing the other part. In both pre-training approaches, we observe that the encoder pre-training contributes more than the decoder pre-training, which is reasonable considering the high ratio of encoder GFLOPs shown in~\ref{fig:intro:gflops}.
    
\vspace{1mm}
\paragraph{Fine-tuning dataset size.} We investigate the effectiveness of three pre-training schemes compared to training \textit{from scratch} when only a limited amount of data is available for fine-tuning in Figure~\ref{fig:low_data}. Specifically, we fine-tune the pre-trained network on $5\%$, $10\%$, $25\%$, and $50\%$ of the COCO training set and evaluate it on the full COCO validation set. All three pre-training schemes greatly speed up the convergence. We observe that DETReg only yields slightly higher performance than random initialization. The Simple Self-training approach remains the most effective, particularly when only a very small amount of data ($5\%$) is available. 

\begin{figure*}[t]
\begin{minipage}[t]{\linewidth}
\centering
\begin{minipage}[t]{0.155\linewidth}
\centering
\begin{subfigure}[b]{\textwidth}
\centering
\includegraphics[width = \textwidth,height=18mm]{img/encoder_discriminability_score_vis_coco_val/ori0.pdf}\\\vspace{1mm}
\includegraphics[width = \textwidth,height=18mm]{img/encoder_discriminability_score_vis_coco_val/ori22.pdf}\\\vspace{1mm}
\includegraphics[width = \textwidth,height=18mm]{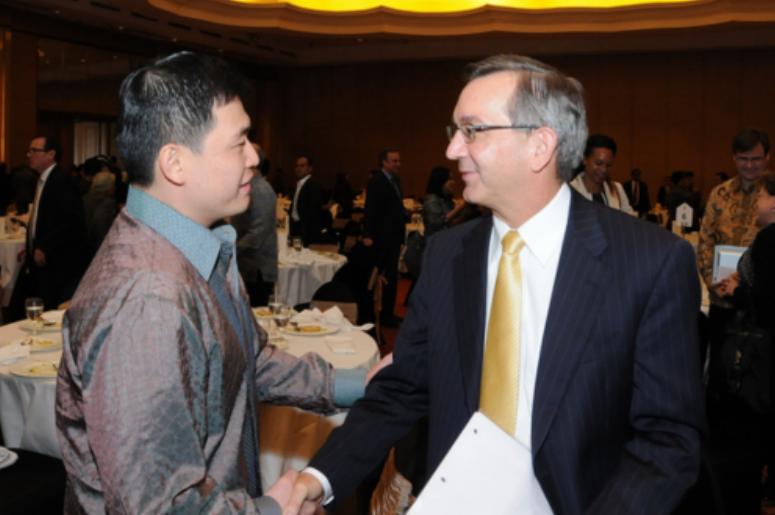}
\parbox[b][6mm][t]{24mm}{\caption*{\centering{\scriptsize{Image}}}}
\end{subfigure}
\end{minipage}
\begin{minipage}[t]{0.155\linewidth}
\centering
\begin{subfigure}[b]{\textwidth}
\centering
\includegraphics[width = \textwidth,height=18mm]{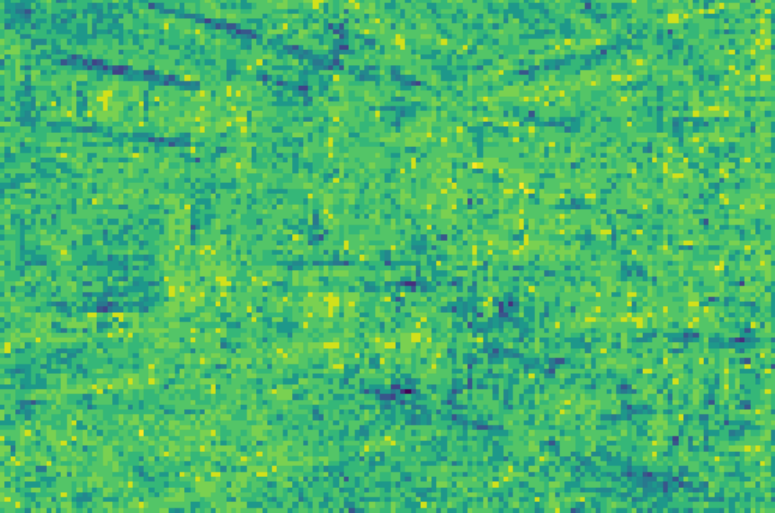}\\\vspace{1mm}
\includegraphics[width = \textwidth,height=18mm]{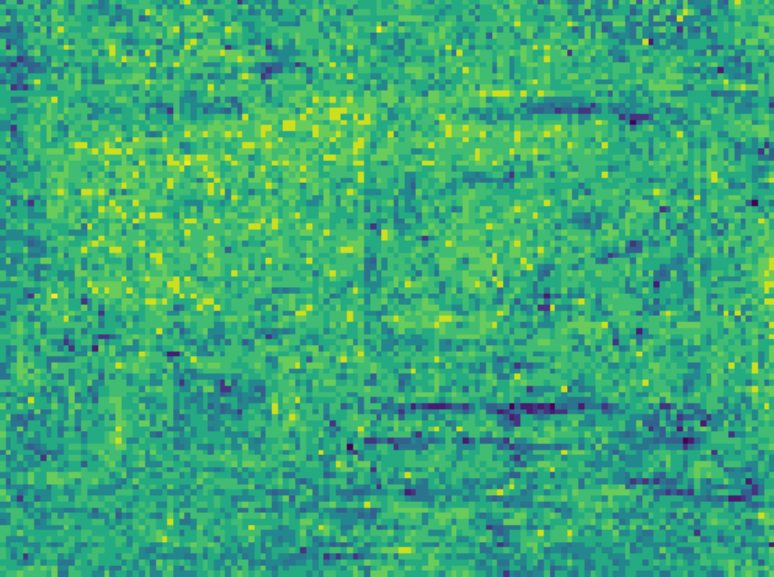}\\\vspace{1mm}
\includegraphics[width = \textwidth,height=18mm]{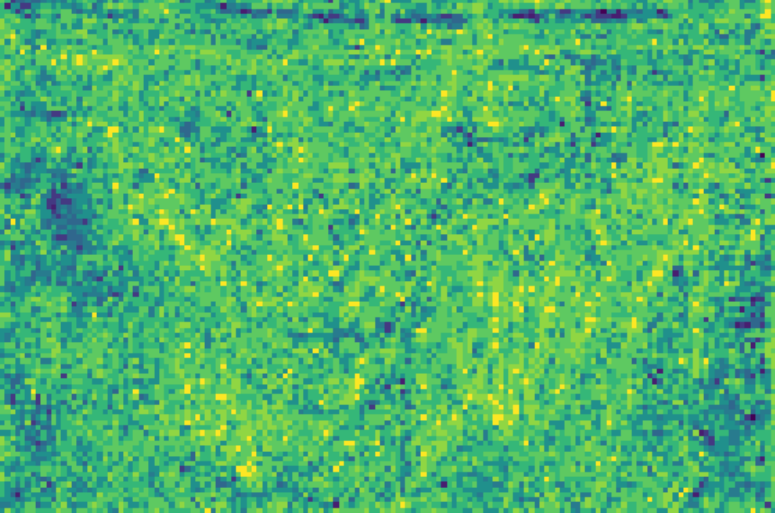}
\parbox[b][6mm][t]{24mm}{\caption*{\centering{\scriptsize{Random initialization}}}}
\end{subfigure}
\end{minipage}
\begin{minipage}[t]{0.155\linewidth}
\centering
\begin{subfigure}[b]{\textwidth}
\centering
\includegraphics[width = \textwidth,height=18mm]{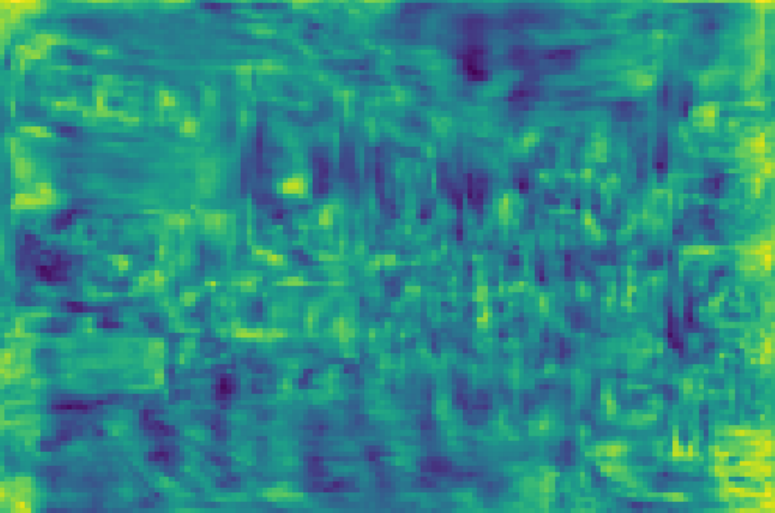}\\\vspace{1mm}
\includegraphics[width = \textwidth,height=18mm]{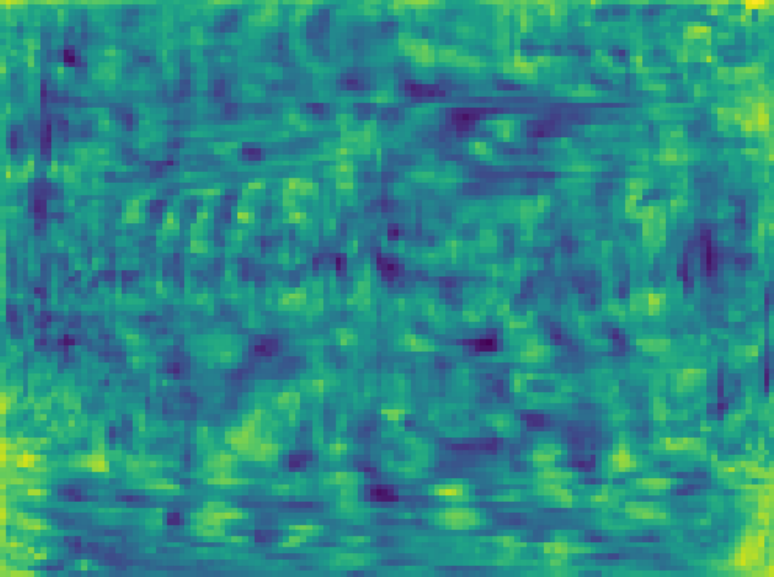}\\\vspace{1mm}
\includegraphics[width = \textwidth,height=18mm]{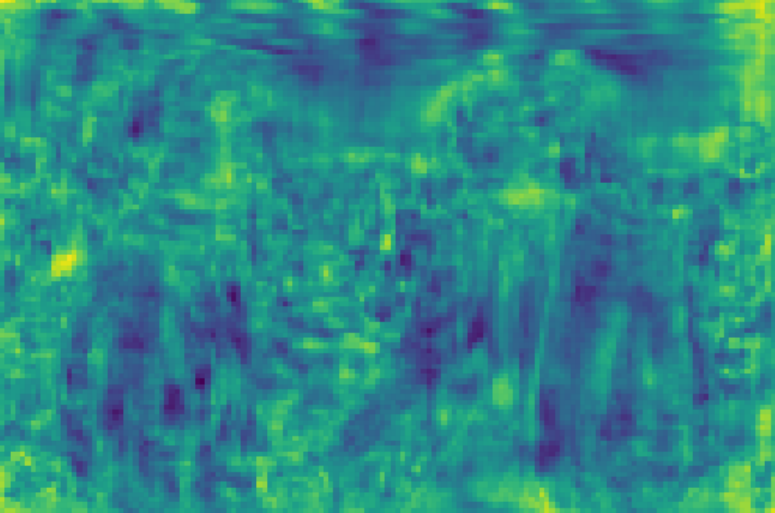}
\parbox[b][6mm][t]{24mm}{\caption*{\centering{\scriptsize{DETReg}}}}
\end{subfigure}
\end{minipage}
\begin{minipage}[t]{0.155\linewidth}
\centering
\begin{subfigure}[b]{\textwidth}
\centering
\includegraphics[width = \textwidth,height=18mm]{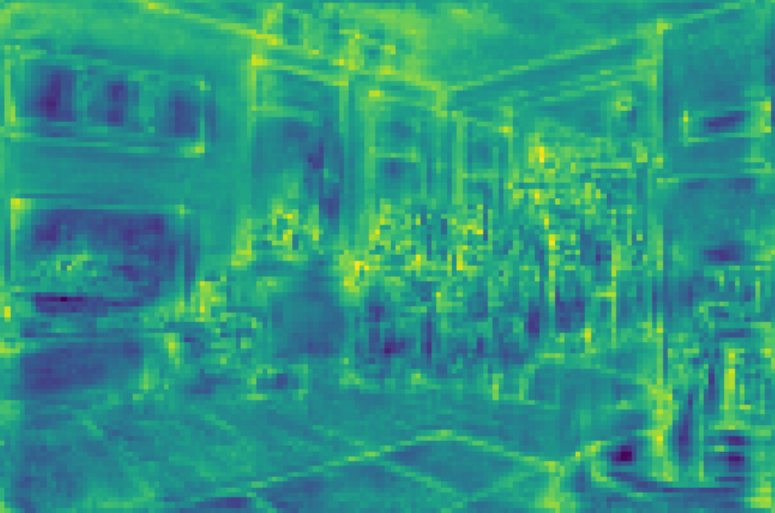}\\\vspace{1mm}
\includegraphics[width = \textwidth,height=18mm]{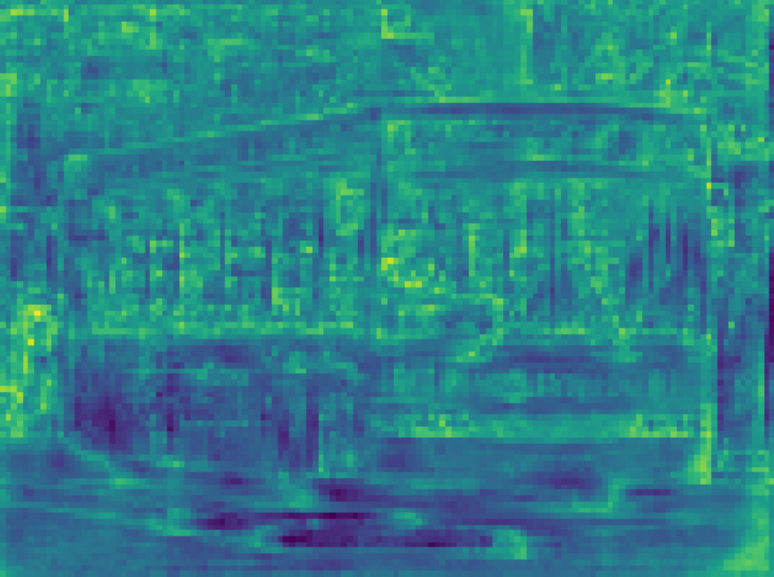}\\\vspace{1mm}
\includegraphics[width = \textwidth,height=18mm]{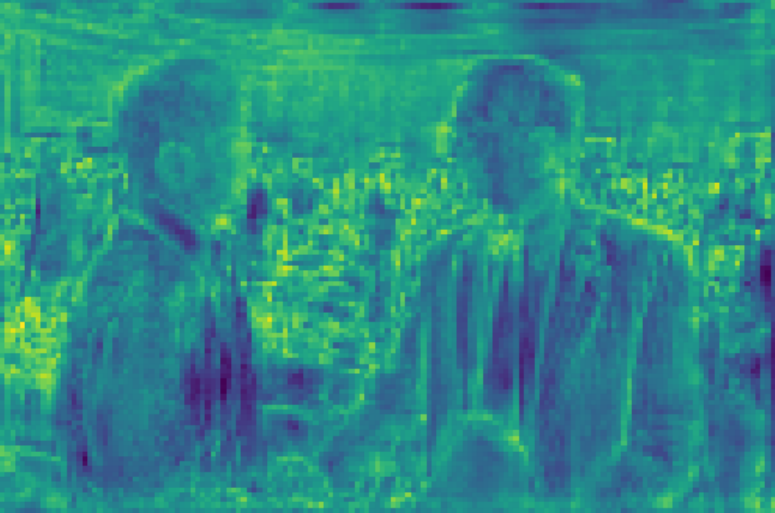}
\parbox[b][6mm][t]{24mm}{\caption*{\centering{\scriptsize{DETReg+ Pseudo-box}}}}
\end{subfigure}
\end{minipage}
\begin{minipage}[t]{0.155\linewidth}
\centering
\begin{subfigure}[b]{\textwidth}
\centering
\includegraphics[width = \textwidth,height=18mm]{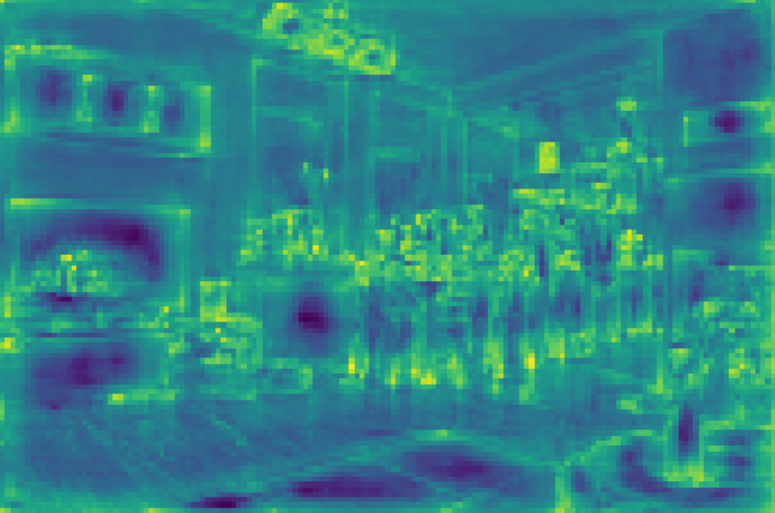}\\\vspace{1mm}
\includegraphics[width = \textwidth,height=18mm]{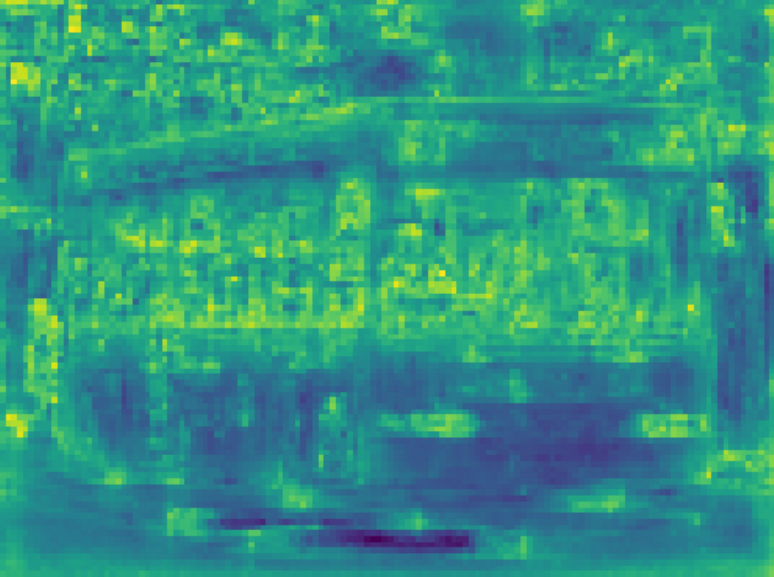}\\\vspace{1mm}
\includegraphics[width = \textwidth,height=18mm]{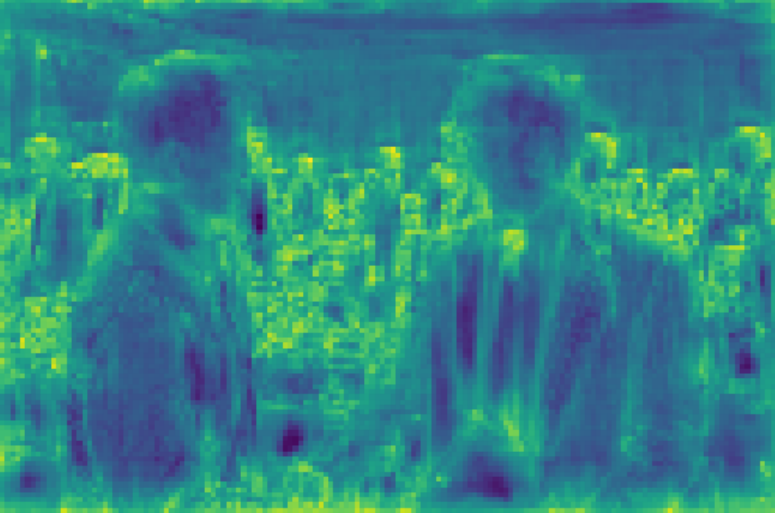}
\parbox[b][6mm][t]{24mm}{\caption*{\centering{\scriptsize{Simple Self-training}}}}
\end{subfigure}
\end{minipage}
\begin{minipage}[t]{0.155\linewidth}
\centering
\begin{subfigure}[b]{\textwidth}
\centering
\includegraphics[width = \textwidth,height=18mm]{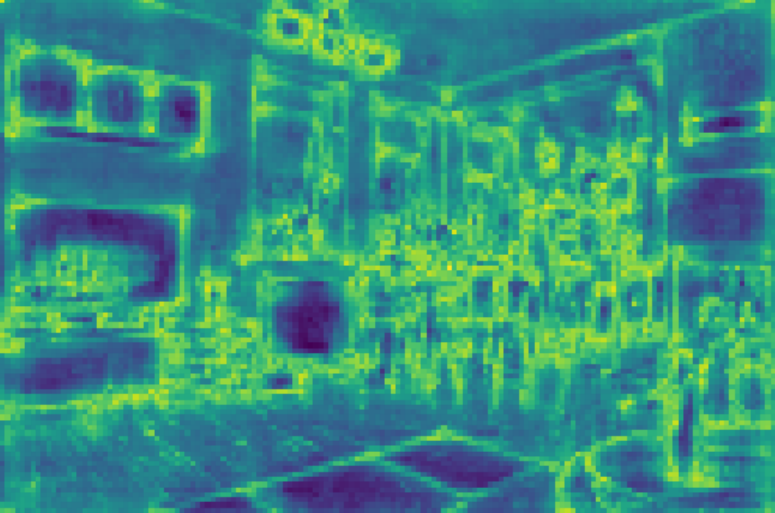}\\\vspace{1mm}
\includegraphics[width = \textwidth,height=18mm]{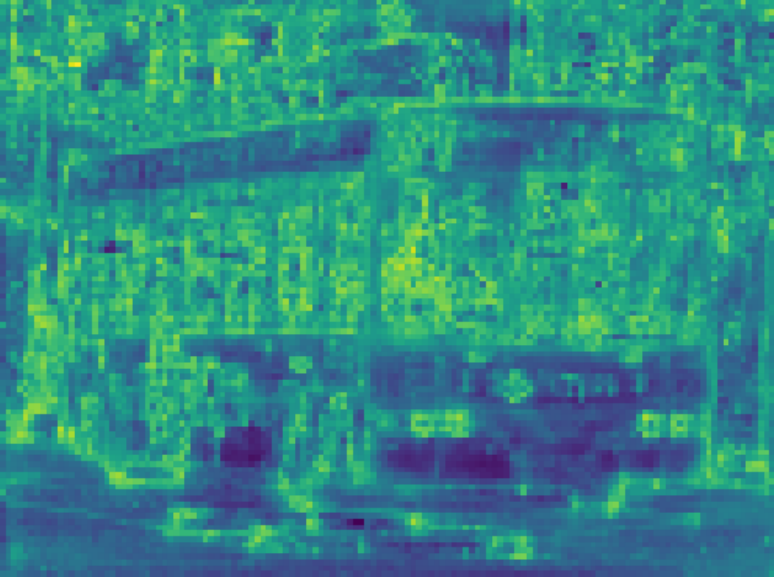}\\\vspace{1mm}
\includegraphics[width = \textwidth,height=18mm]{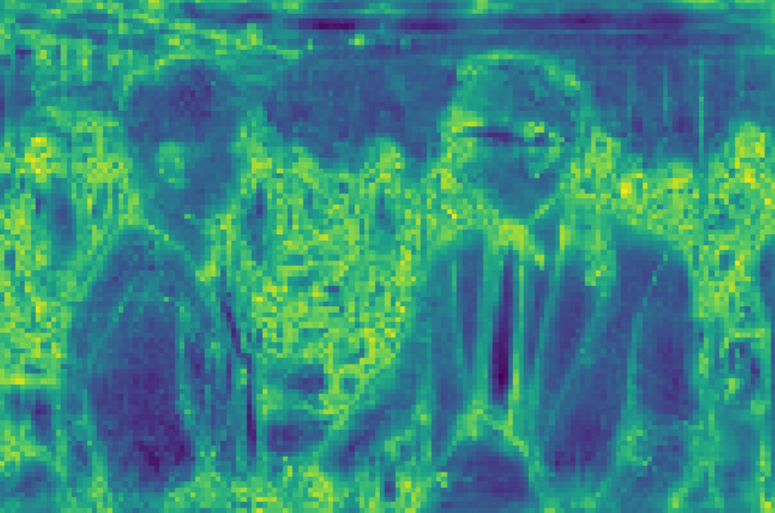}
\parbox[b][6mm][t]{24mm}{\caption*{\centering{\scriptsize{Supervised-pretraining}}}}
\end{subfigure}
\end{minipage}
\caption{\small{Visualizations of discriminability scores in the encoder on COCO \texttt{val} images.}}
\label{fig:vis_discriminability_scores_extra}
\end{minipage}
\end{figure*}

\begin{figure*}[t]
\begin{minipage}[t]{\linewidth}
% \vspace{5mm}
\centering
\begin{minipage}[t]{0.19\linewidth}
\centering
\begin{subfigure}[b]{\textwidth}
\centering
\includegraphics[width = \textwidth,height=22mm]{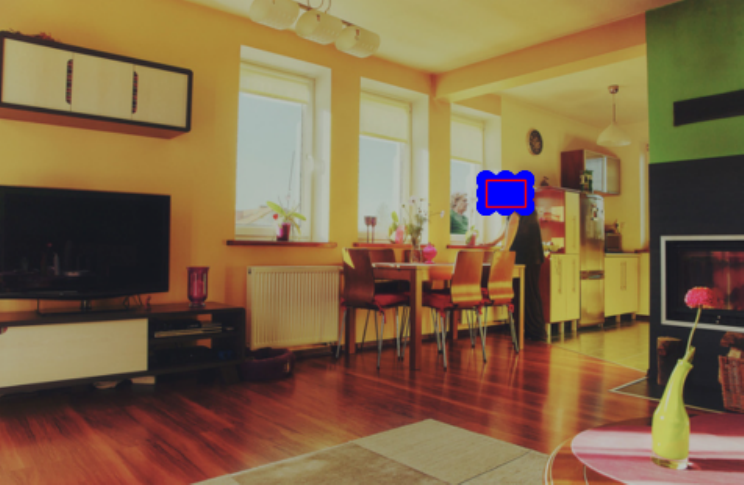}\\\vspace{1mm}
\includegraphics[width = \textwidth,height=22mm]{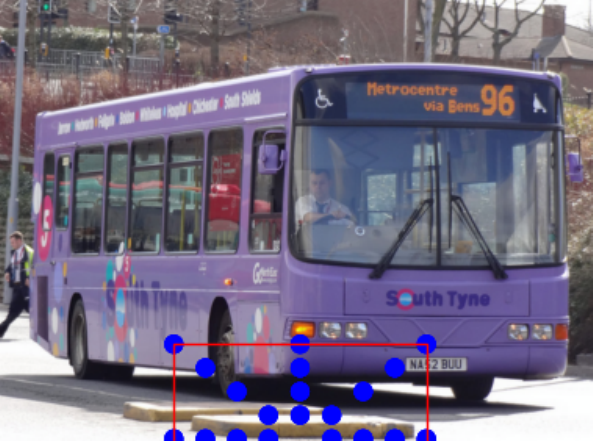}\\\vspace{1mm}
\includegraphics[width = \textwidth,height=22mm]{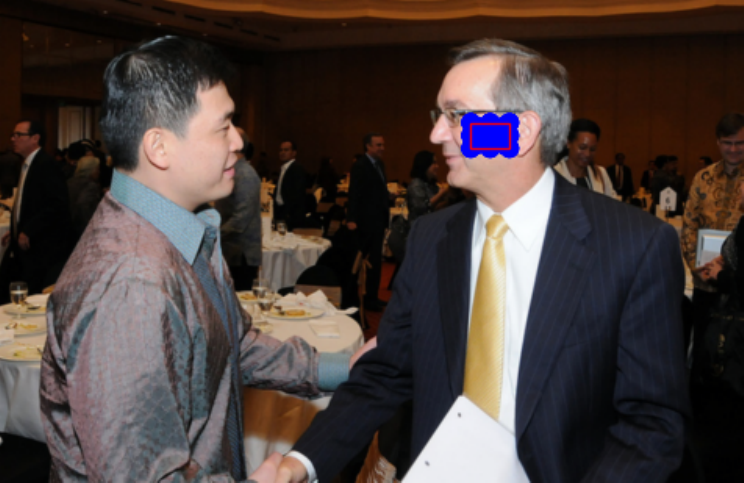}
\parbox[b][6mm][t]{24mm}{\caption*{\centering{\scriptsize{Random initialization}}}}
\end{subfigure}
\end{minipage}
\begin{minipage}[t]{0.19\linewidth}
\centering
\begin{subfigure}[b]{\textwidth}
\centering
\includegraphics[width = \textwidth,height=22mm]{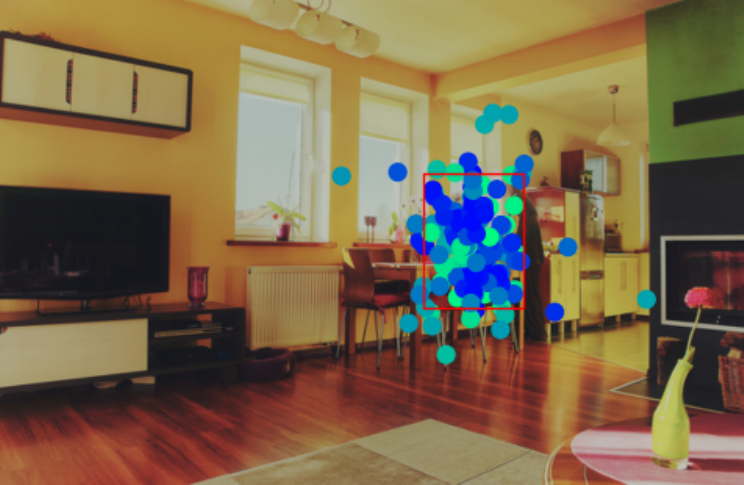}\\\vspace{1mm}
\includegraphics[width = \textwidth,height=22mm]{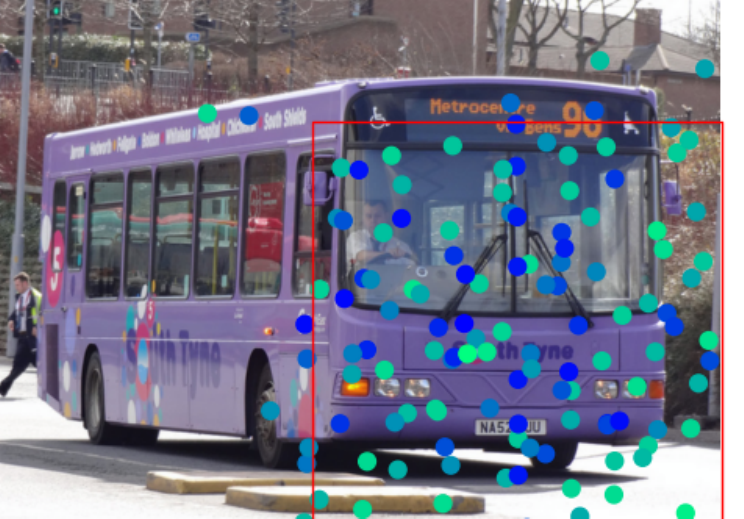}\\\vspace{1mm}
\includegraphics[width = \textwidth,height=22mm]{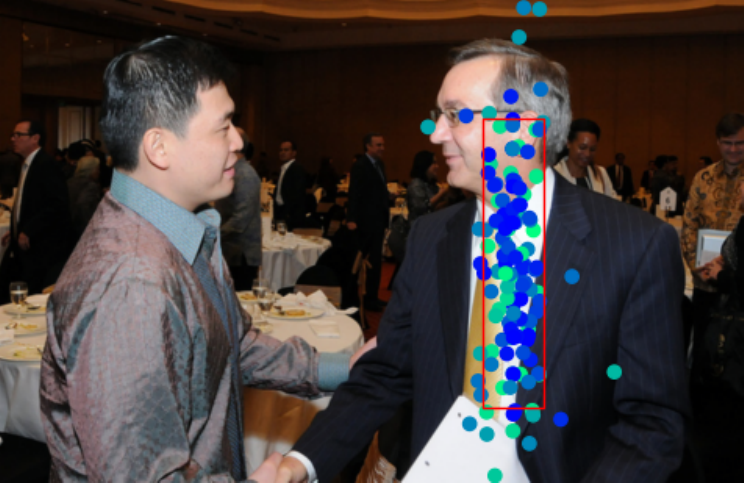}
\parbox[b][6mm][t]{24mm}{\caption*{\centering{\scriptsize{DETReg}}}}
\end{subfigure}
\end{minipage}
\begin{minipage}[t]{0.19\linewidth}
\centering
\begin{subfigure}[b]{\textwidth}
\centering
\includegraphics[width = \textwidth,height=22mm]{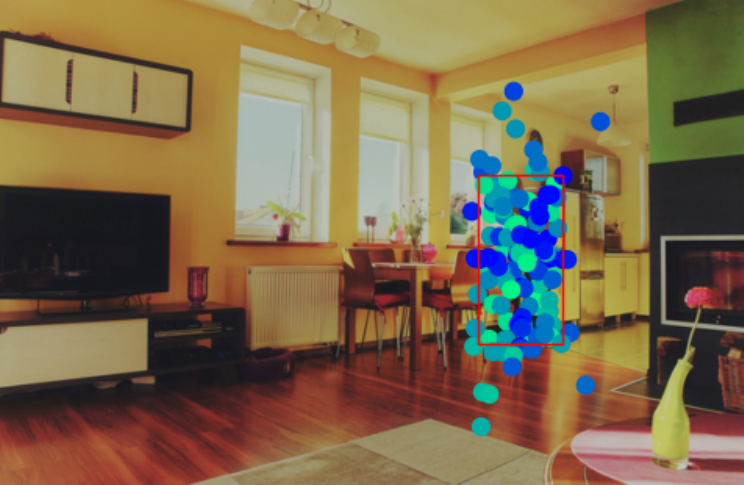}\\\vspace{1mm}
\includegraphics[width = \textwidth,height=22mm]{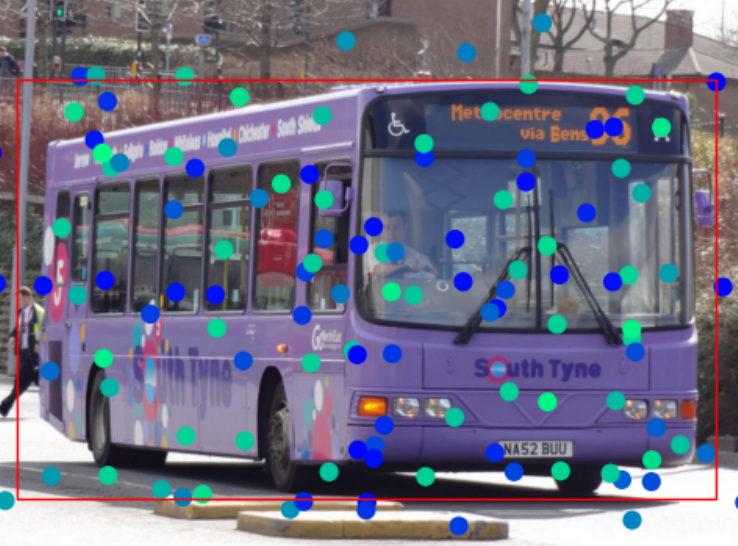}\\\vspace{1mm}
\includegraphics[width = \textwidth,height=22mm]{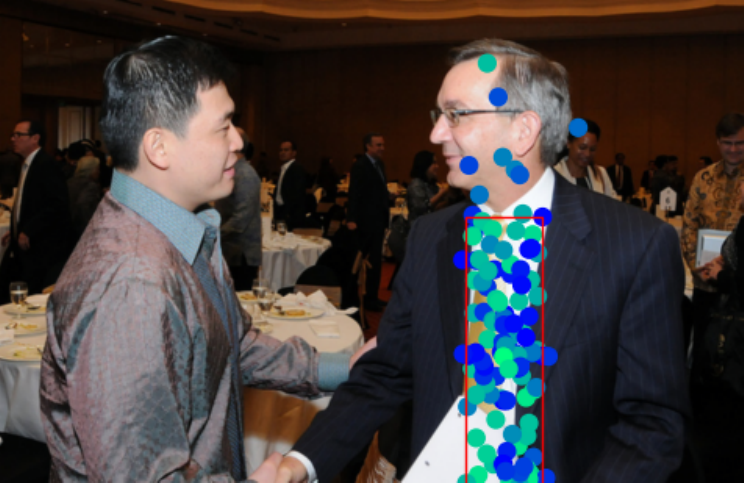}
\parbox[b][6mm][t]{24mm}{\caption*{\centering{\scriptsize{DETReg+ Pseudo-box}}}}
\end{subfigure}
\end{minipage}
\begin{minipage}[t]{0.19\linewidth}
\centering
\begin{subfigure}[b]{\textwidth}
\centering
\includegraphics[width = \textwidth,height=22mm]{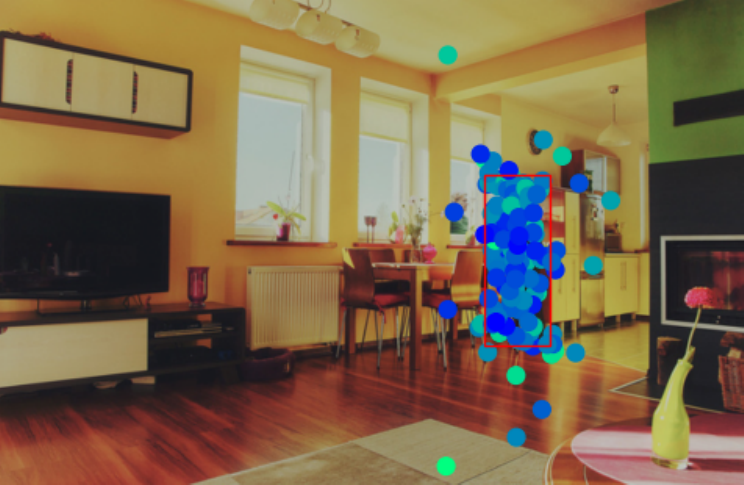}\\\vspace{1mm}
\includegraphics[width = \textwidth,height=22mm]{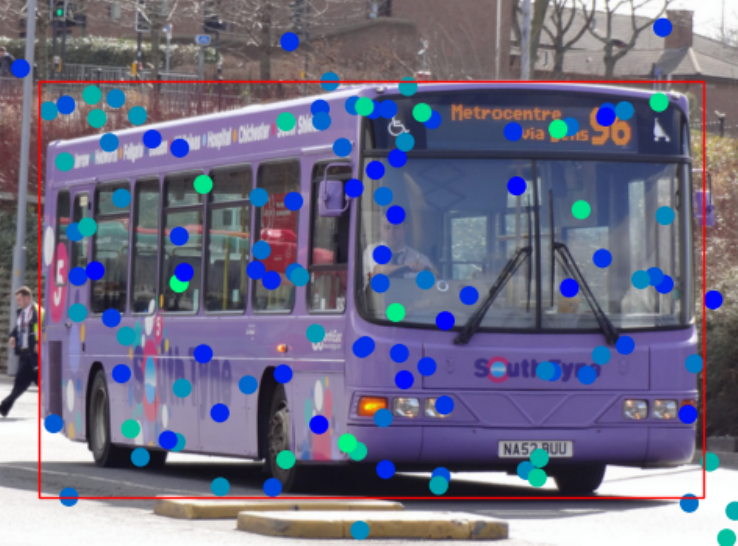}\\\vspace{1mm}
\includegraphics[width = \textwidth,height=22mm]{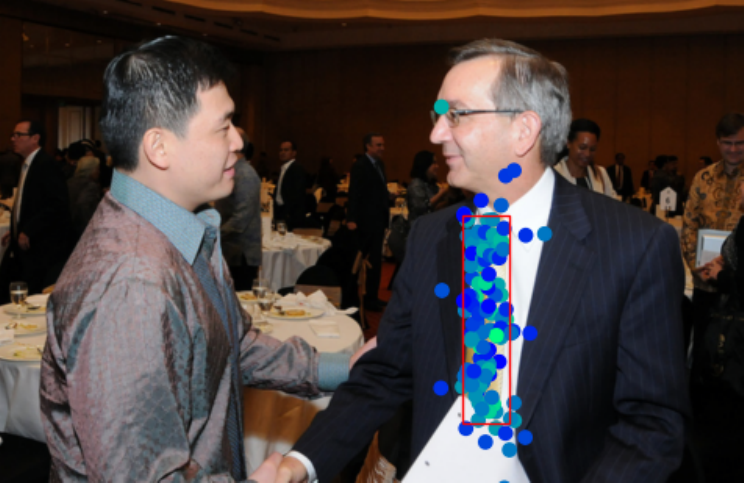}
\parbox[b][6mm][t]{24mm}{\caption*{\centering{\scriptsize{Simple Self-training}}}}
\end{subfigure}
\end{minipage}
\begin{minipage}[t]{0.19\linewidth}
\centering
\begin{subfigure}[b]{\textwidth}
\centering
\includegraphics[width = \textwidth,height=22mm]{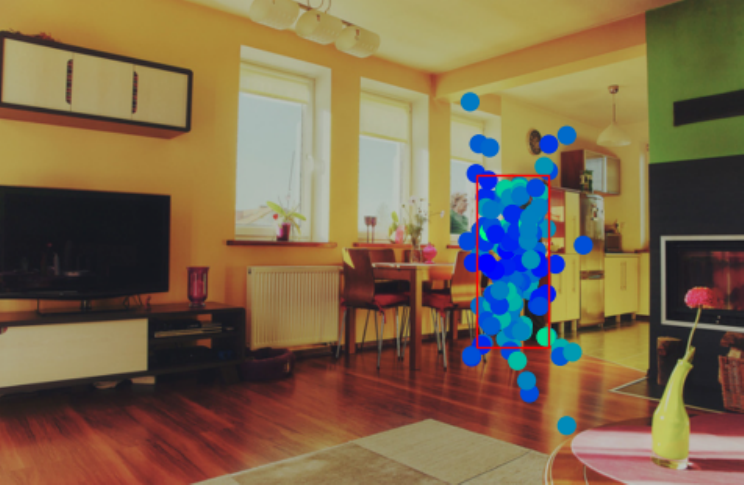}\\\vspace{1mm}
\includegraphics[width = \textwidth,height=22mm]{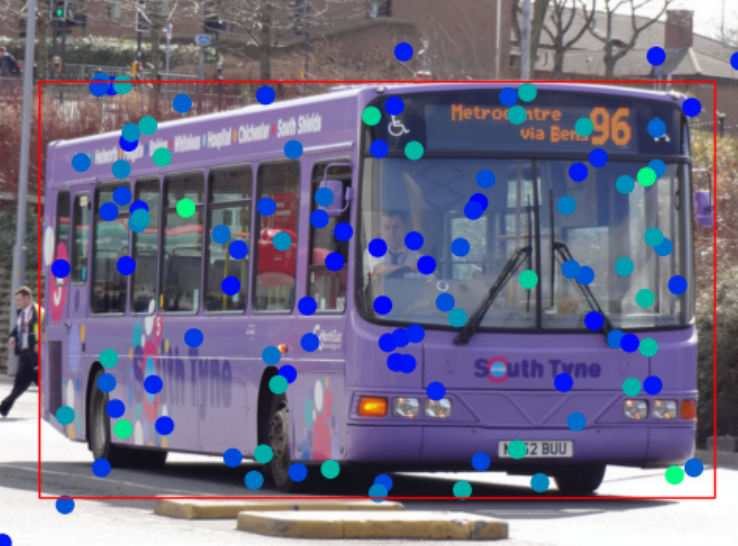}\\\vspace{1mm}
\includegraphics[width = \textwidth,height=22mm]{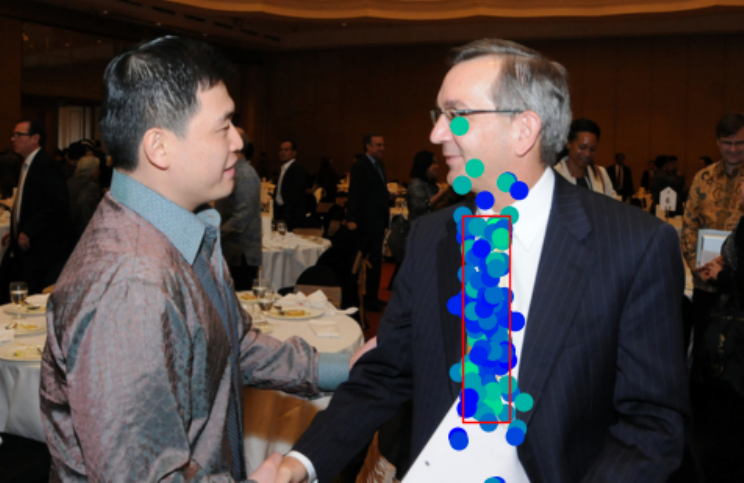}
\parbox[b][6mm][t]{24mm}{\caption*{\centering{\scriptsize{Supervised-pretraining}}}}
\end{subfigure}
\end{minipage}
\caption{\small{Visualizations of deformable cross-attention based on the last Transformer decoder layer on COCO \texttt{val} images.}}
\label{fig:vis_decoder_cross_attention_extra}
\end{minipage}
\end{figure*}
    
\vspace{1mm}
\paragraph{Qualitative analysis.} Without fine-tuning, we visualize the discriminability scores~\citep{zong2022detrs} of the pre-trained encoder in Figure~\ref{fig:vis_discriminability_scores_extra} to investigate what the encoder has learned in pre-training. From the figures, we can see that DETReg's feature discriminability is seriously disturbed by the background. However, when we utilize improved localization and classification targets in the DETReg+pseudo-box and Simple Self-training approach, finer details are captured.
Notably, the Simple Self-training method demonstrates performance that is almost on par with pre-training using ground-truth.

We also visualize the deformable cross-attention of the pre-trained decoder in Figure~\ref{fig:vis_decoder_cross_attention_extra}. The colored dots in the image represent the sampling points from all resolution scales, where the color indicates the attention weights, with a lighter color indicating higher attention. As random initialization shows, the initial key points are sampled radially from the center to the edge. All pre-training methods learn to scatter the sampling points across the entire object of interest with different patterns, while the Simple Self-training pre-trained decoder can sample key points from an accurate range of objects and distribute attention weight more effectively.

\begin{figure*}[t]
\begin{minipage}[t]{\linewidth}
\centering
\begin{minipage}[t]{0.19\linewidth}
\centering
\begin{subfigure}[b]{\textwidth}
\centering
\includegraphics[width = \textwidth,height=24mm]{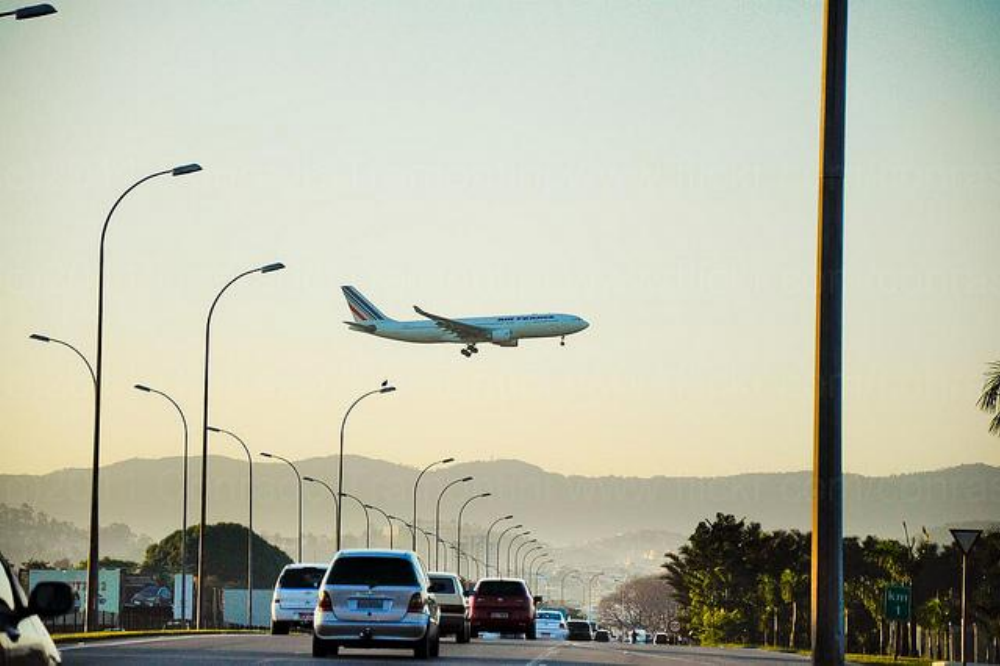}\\\vspace{1mm}
\includegraphics[width = \textwidth,height=24mm]{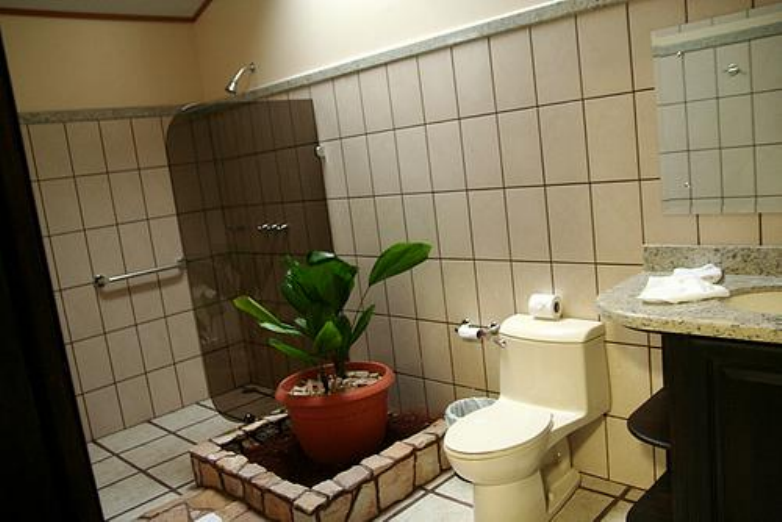}\\\vspace{1mm}
\includegraphics[width = \textwidth,height=24mm]{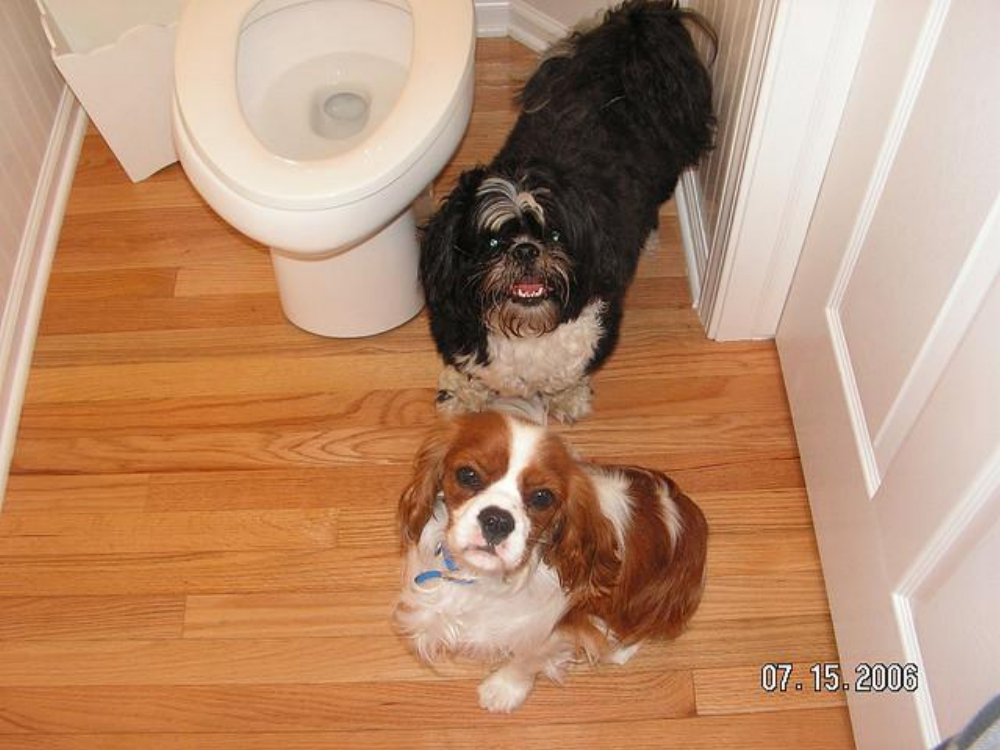}\\\vspace{1mm}
\includegraphics[width = \textwidth,height=24mm]{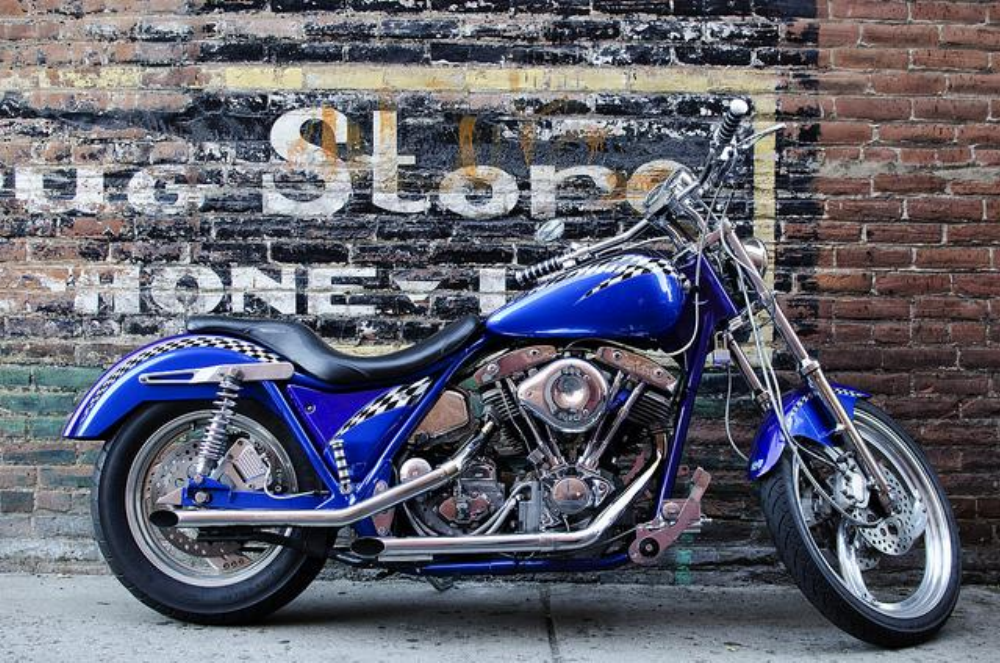}\\\vspace{1mm}
\includegraphics[width = \textwidth,height=24mm]{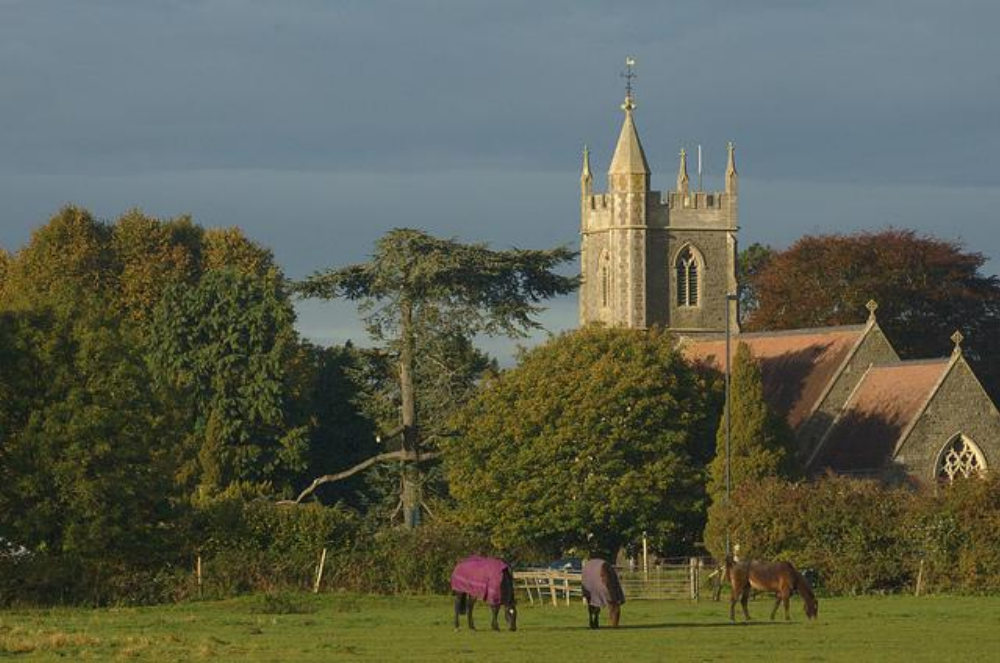}\\\vspace{1mm}
\includegraphics[width = \textwidth,height=24mm]{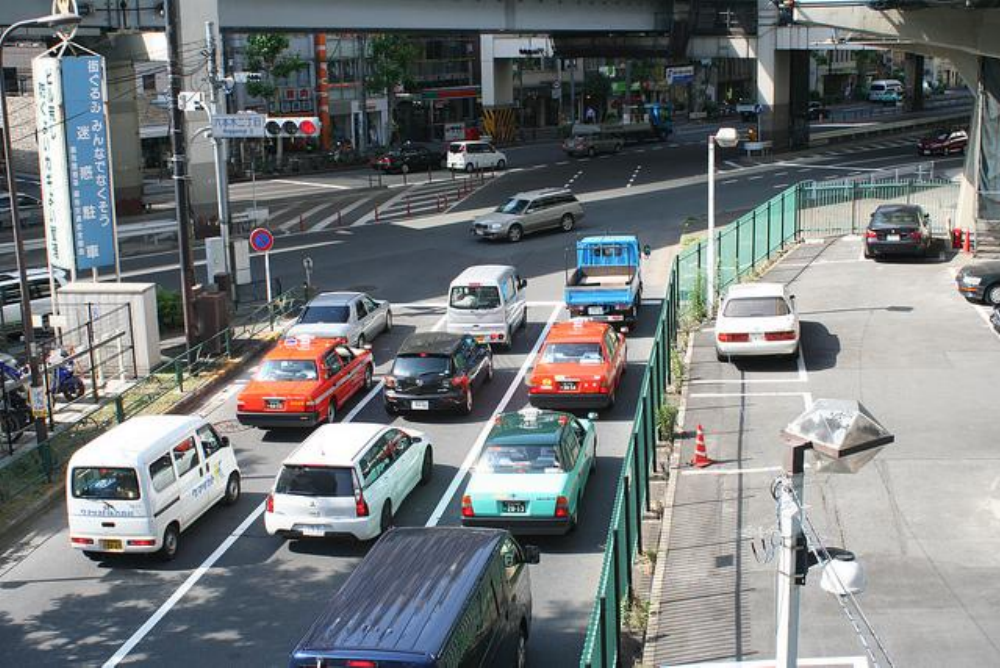}\\\vspace{1mm}
\includegraphics[width = \textwidth,height=24mm]{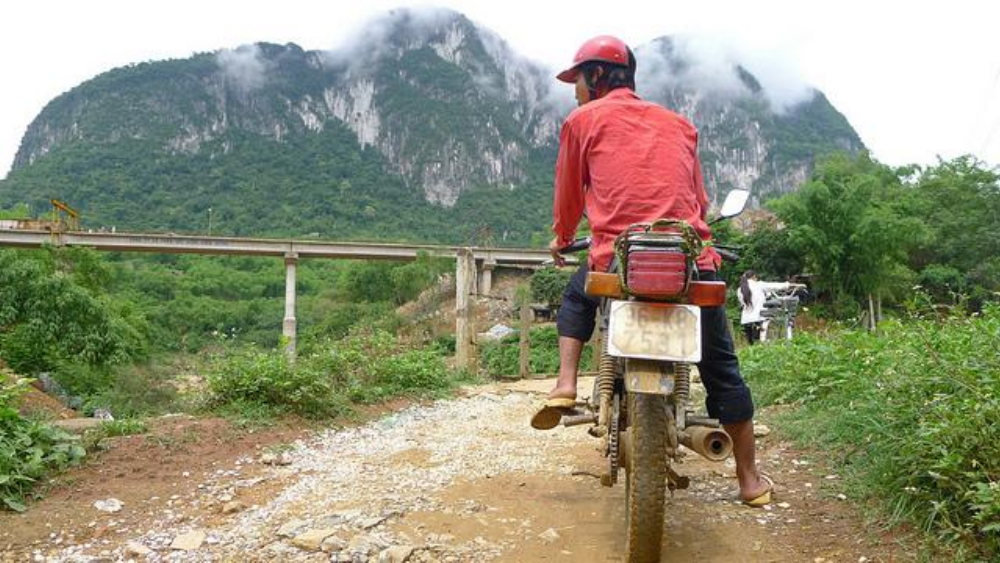}\\\vspace{1mm}
\includegraphics[width = \textwidth,height=24mm]{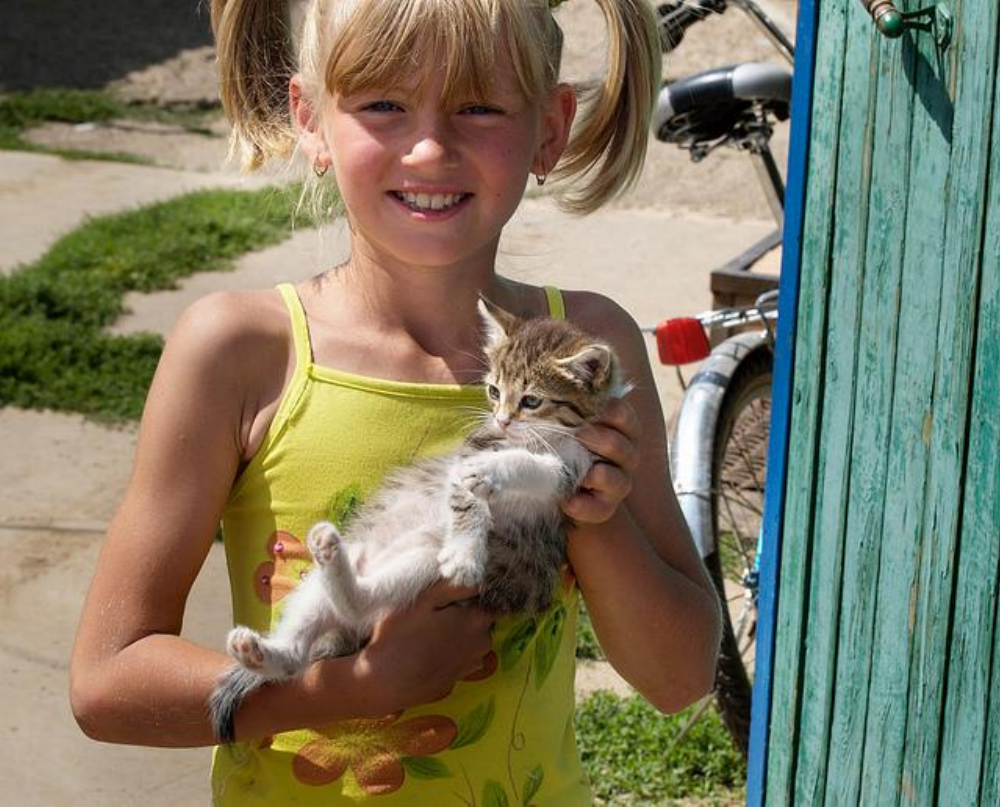}
\parbox[b][6mm][t]{20mm}{\caption*{\centering{Original images}}}
\end{subfigure}
\end{minipage}
\begin{minipage}[t]{0.19\linewidth}
\centering
\begin{subfigure}[b]{\textwidth}
\centering
\includegraphics[width = \textwidth,height=24mm]{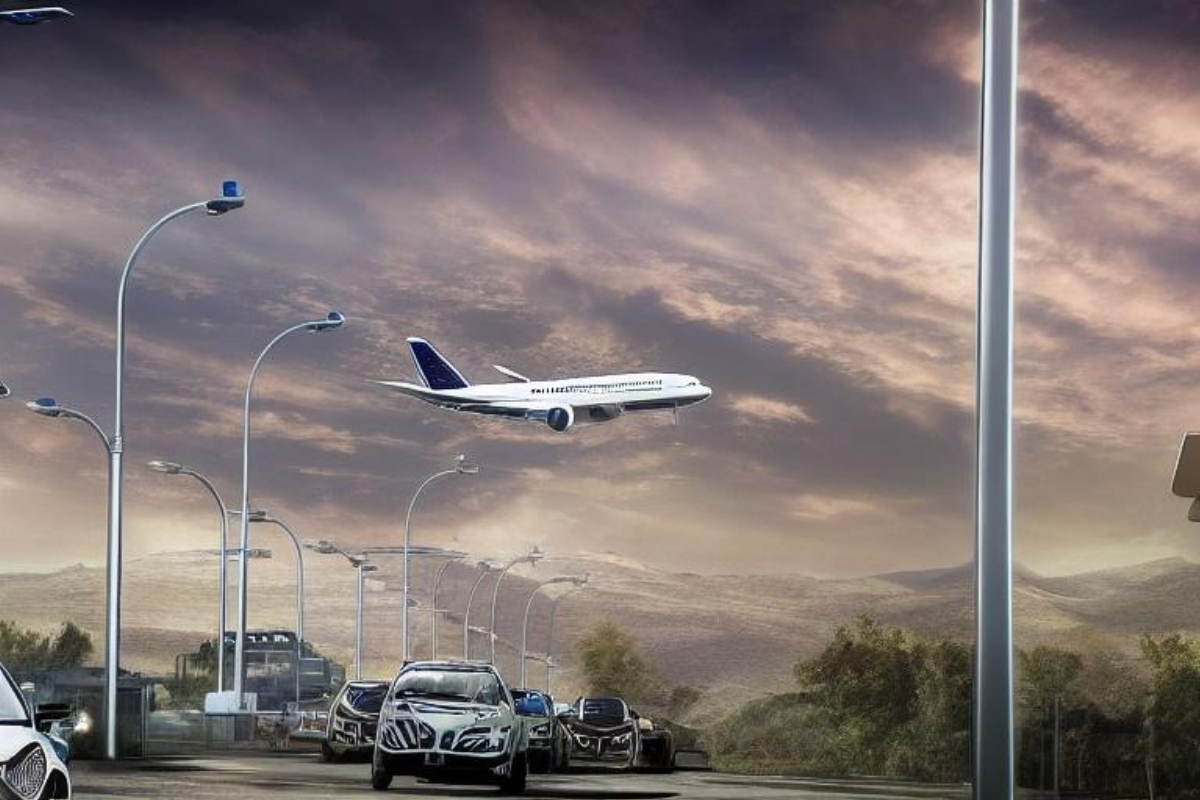}\\\vspace{1mm}
\includegraphics[width = \textwidth,height=24mm]{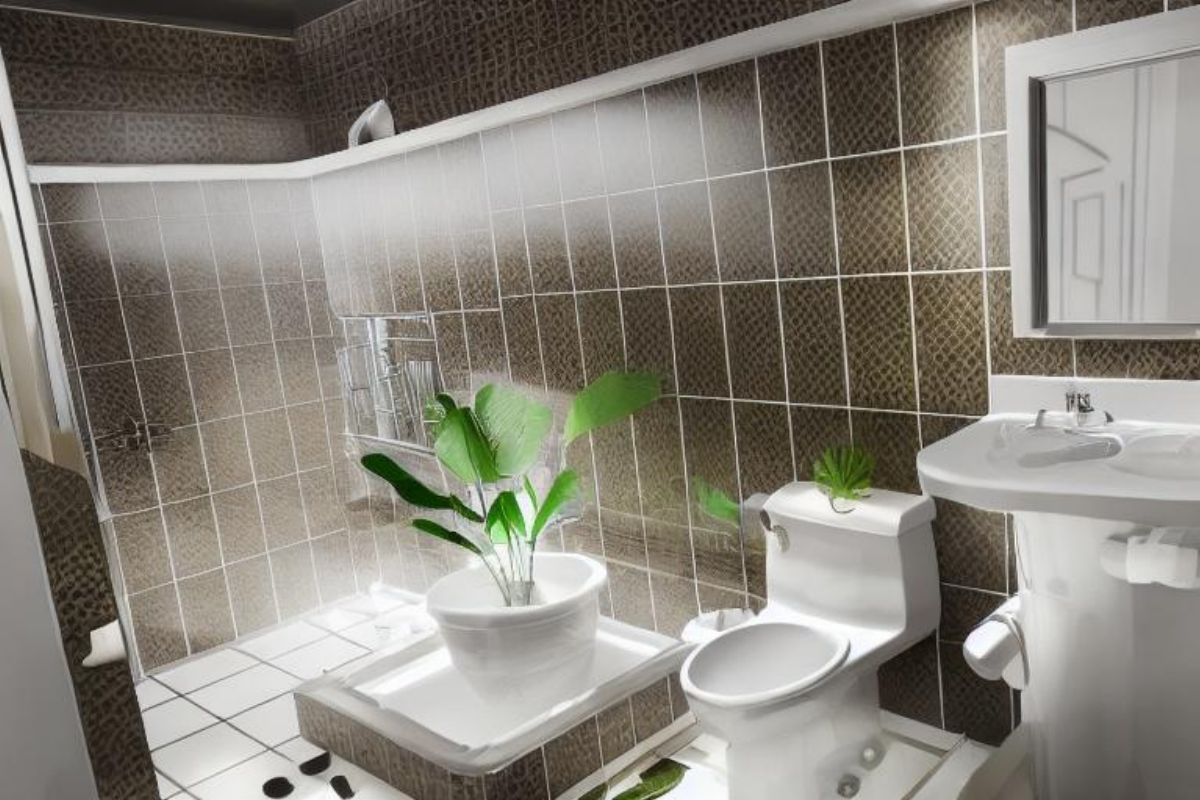}\\\vspace{1mm}
\includegraphics[width = \textwidth,height=24mm]{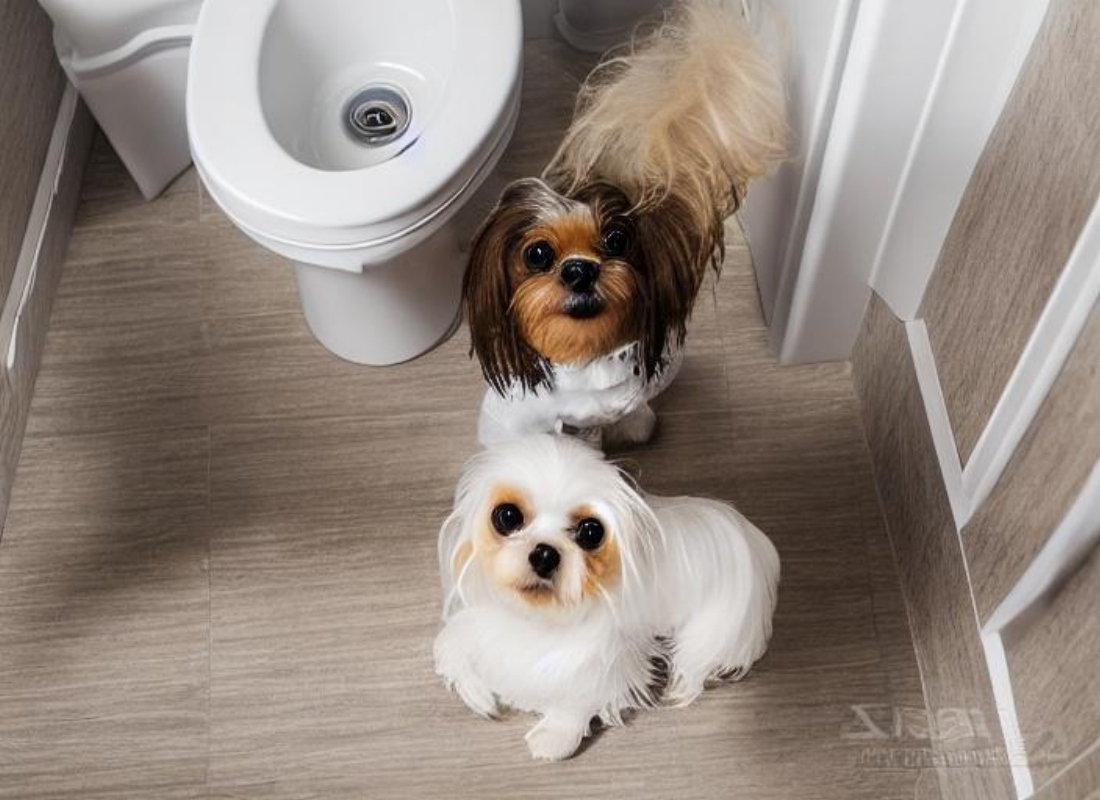}\\\vspace{1mm}
\includegraphics[width = \textwidth,height=24mm]{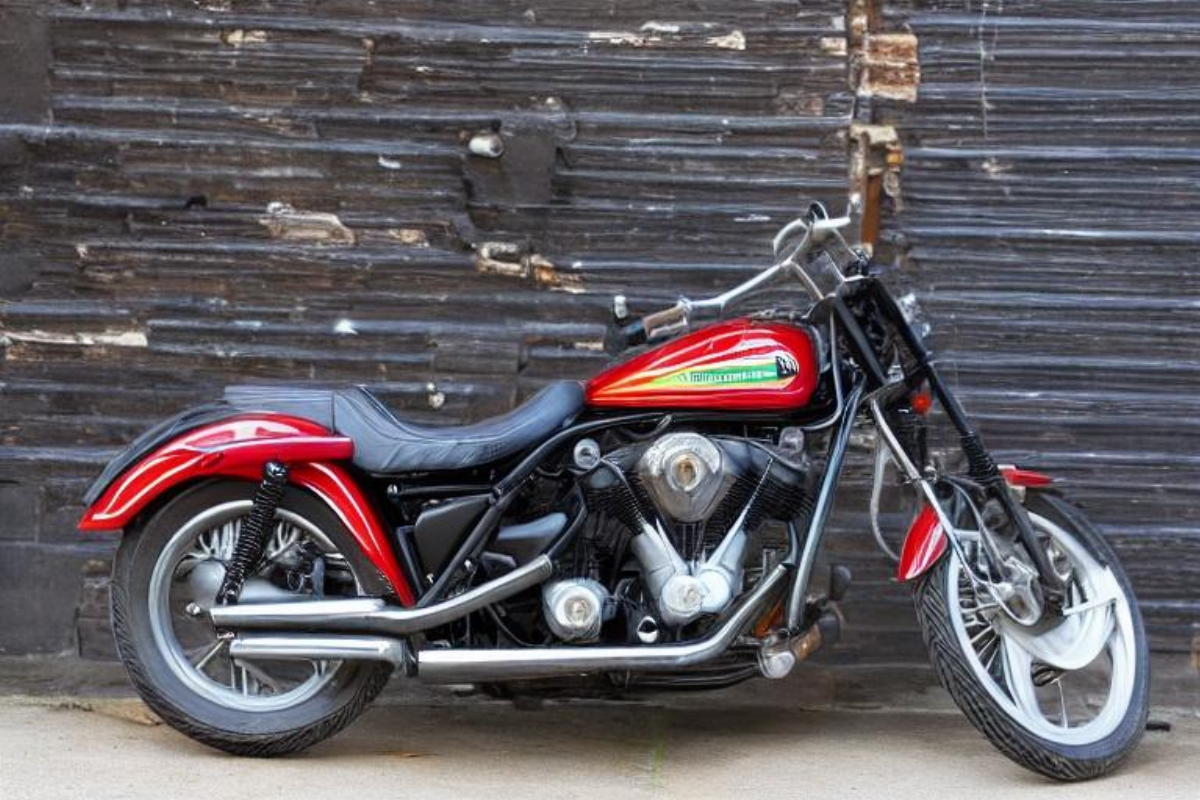}\\\vspace{1mm}
\includegraphics[width = \textwidth,height=24mm]{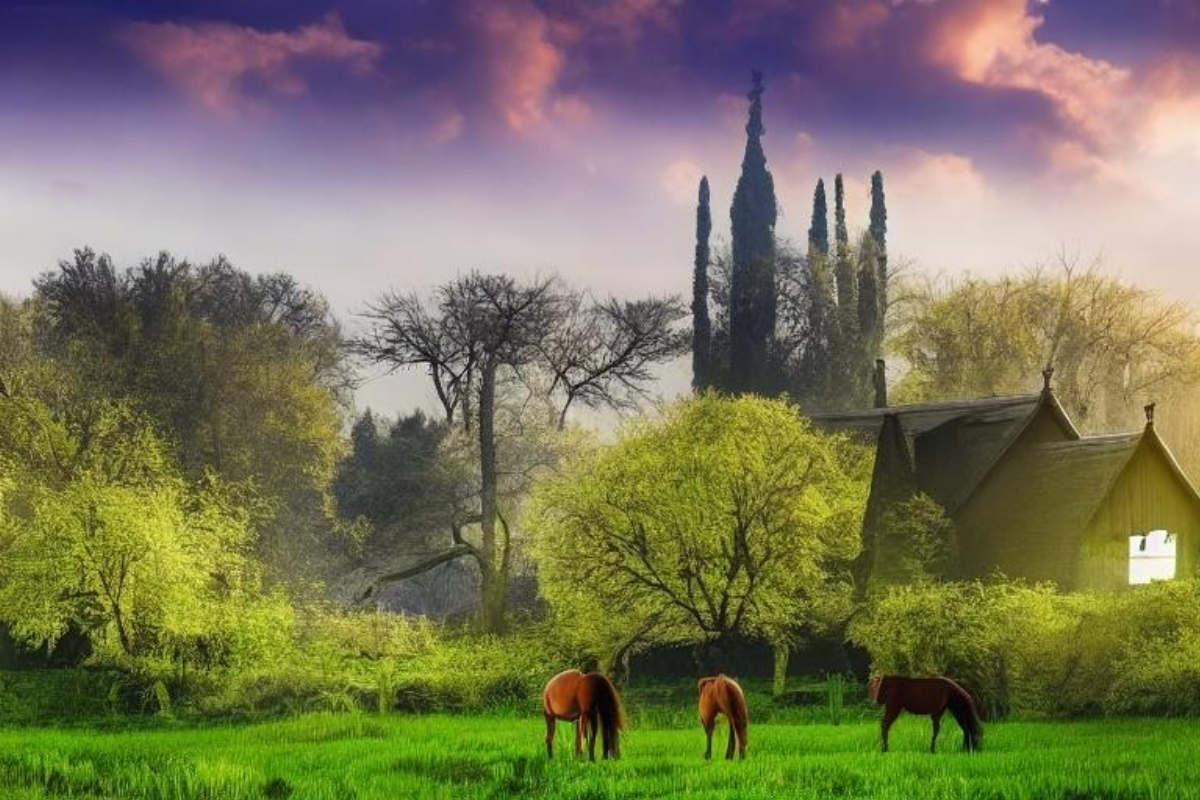}\\\vspace{1mm}
\includegraphics[width = \textwidth,height=24mm]{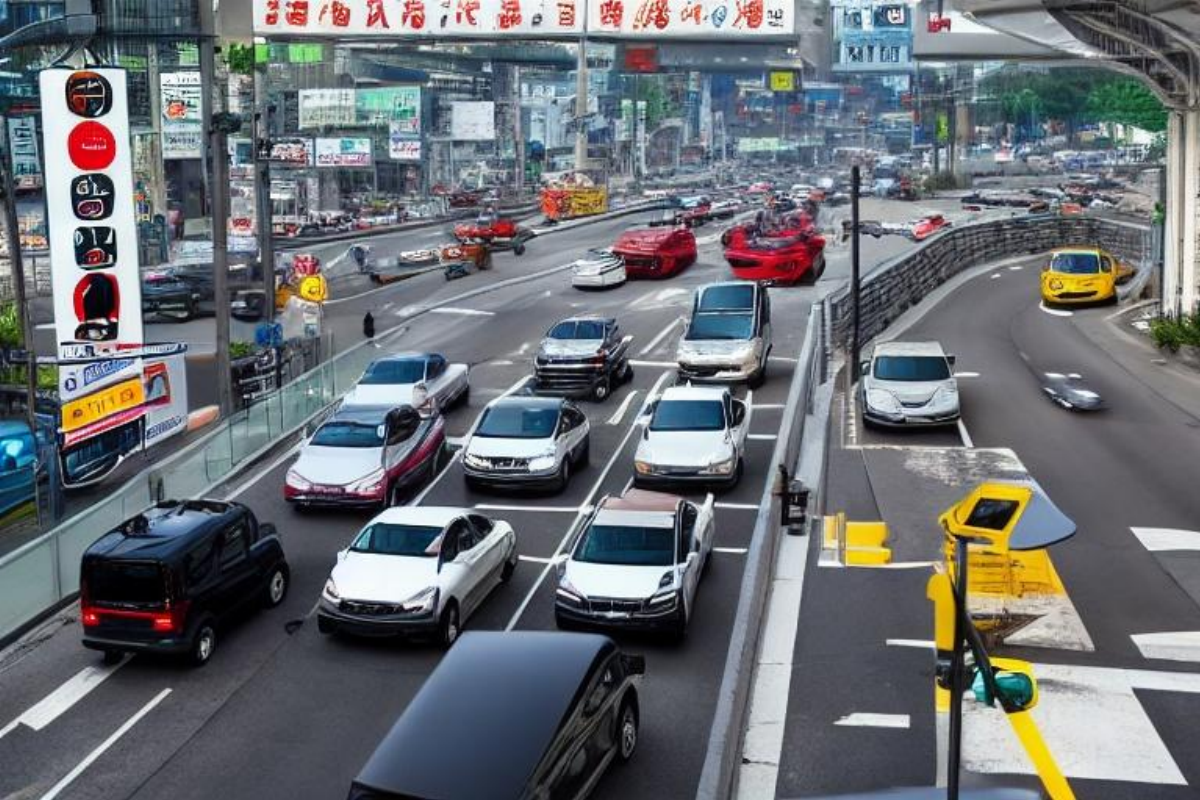}\\\vspace{1mm}
\includegraphics[width = \textwidth,height=24mm]{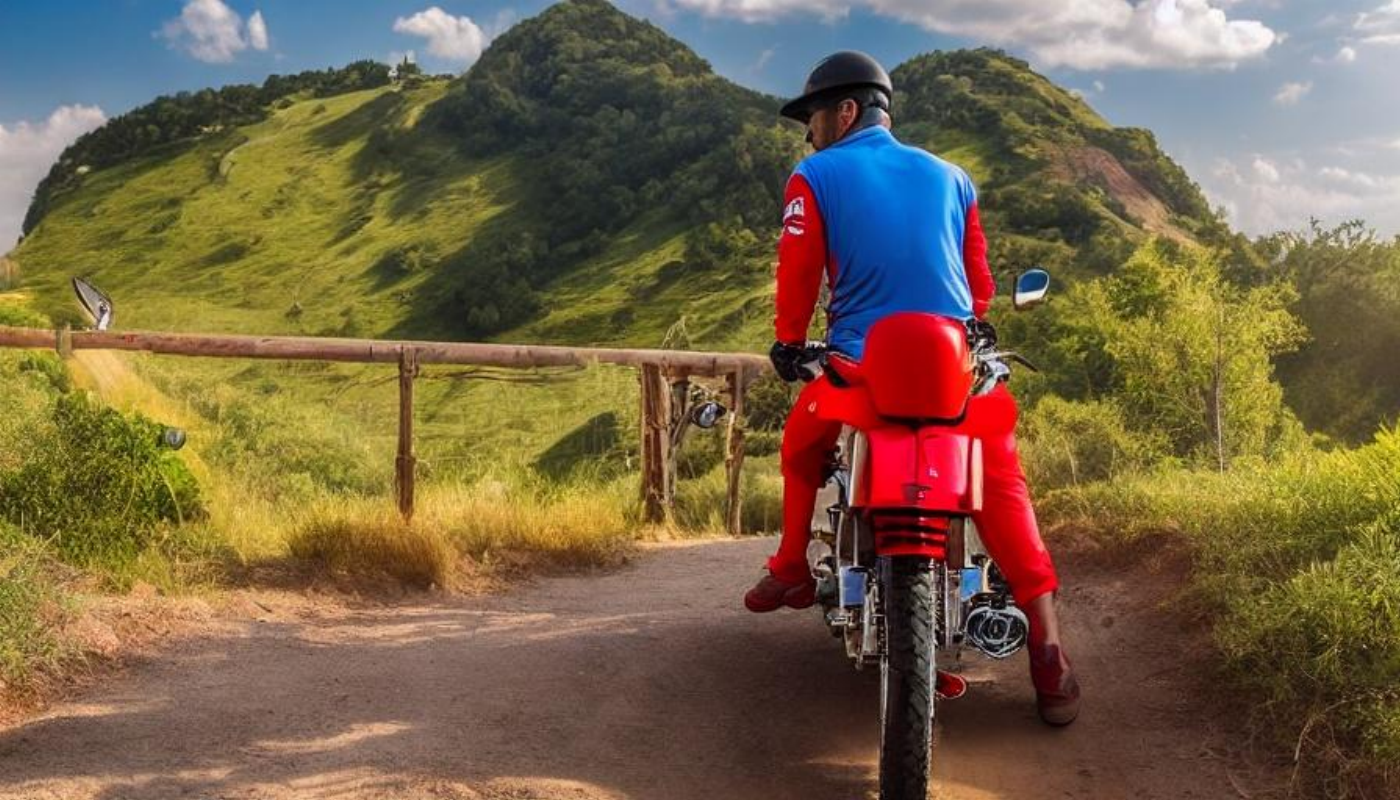}\\\vspace{1mm}
\includegraphics[width = \textwidth,height=24mm]{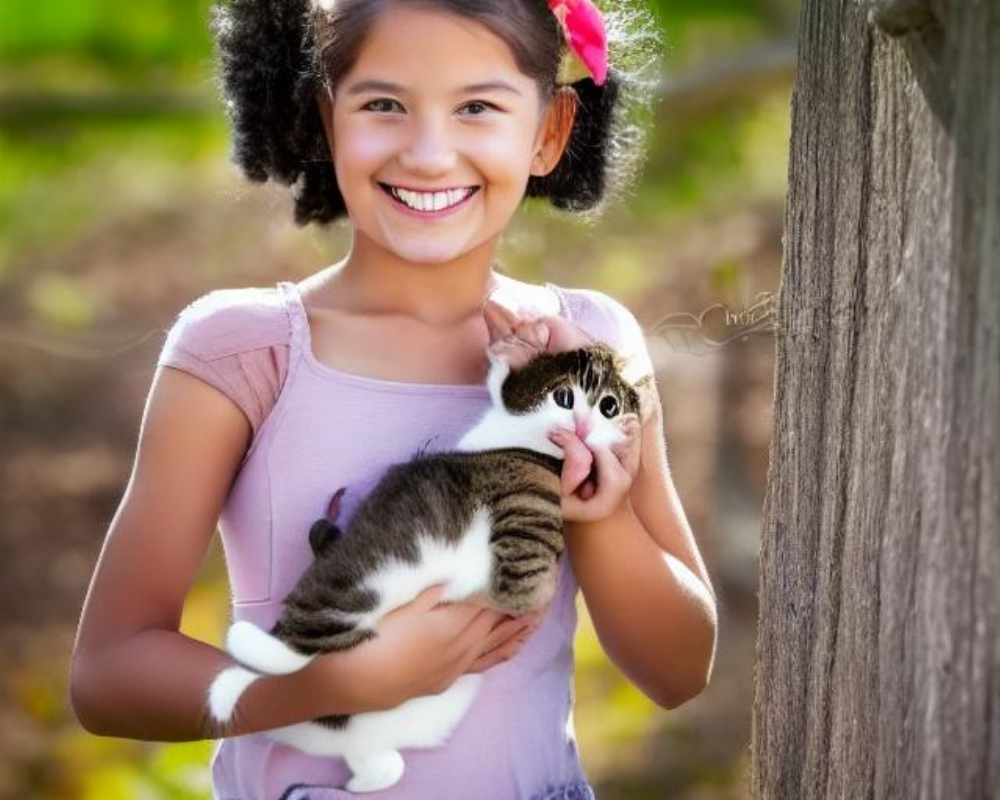}
\parbox[b][6mm][t]{20mm}{\caption*{\centering{COCO captions + ControlNet}}}
\end{subfigure}
\end{minipage}
\begin{minipage}[t]{0.19\linewidth}
\centering
\begin{subfigure}[b]{\textwidth}
\centering
\includegraphics[width = \textwidth,height=24mm]{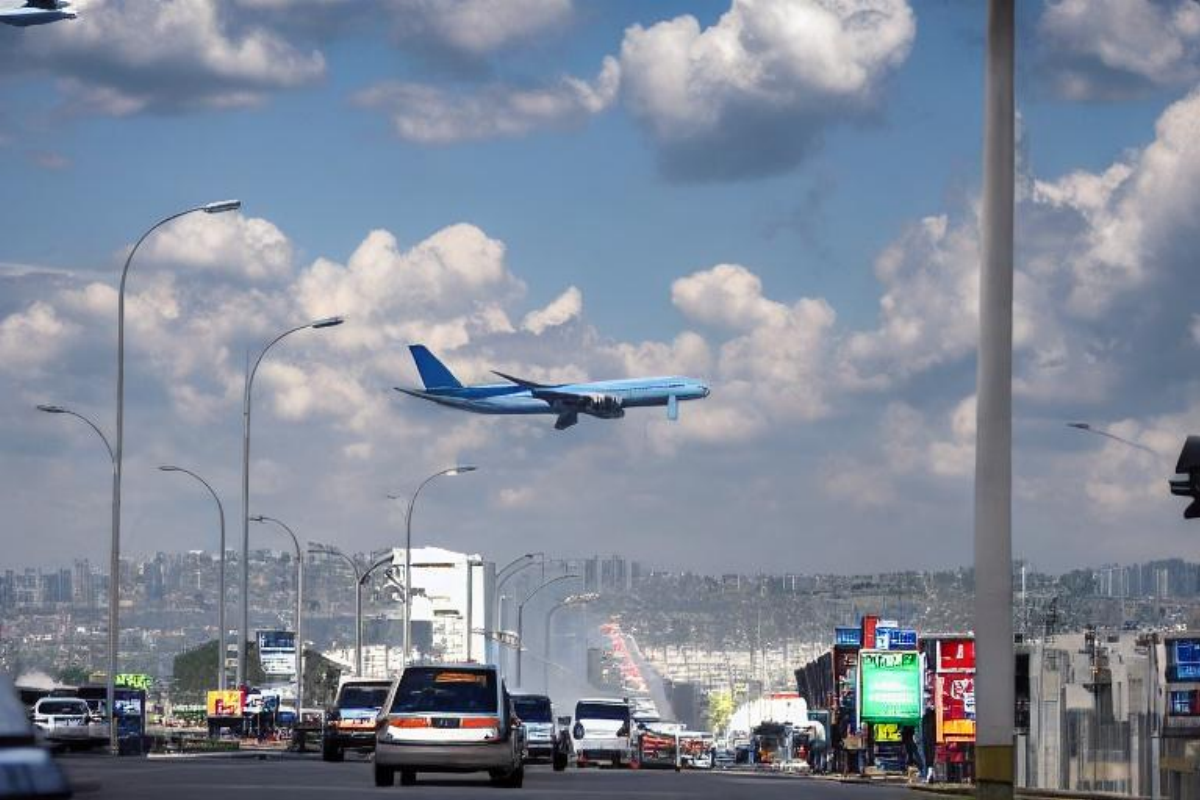}\\\vspace{1mm}
\includegraphics[width = \textwidth,height=24mm]{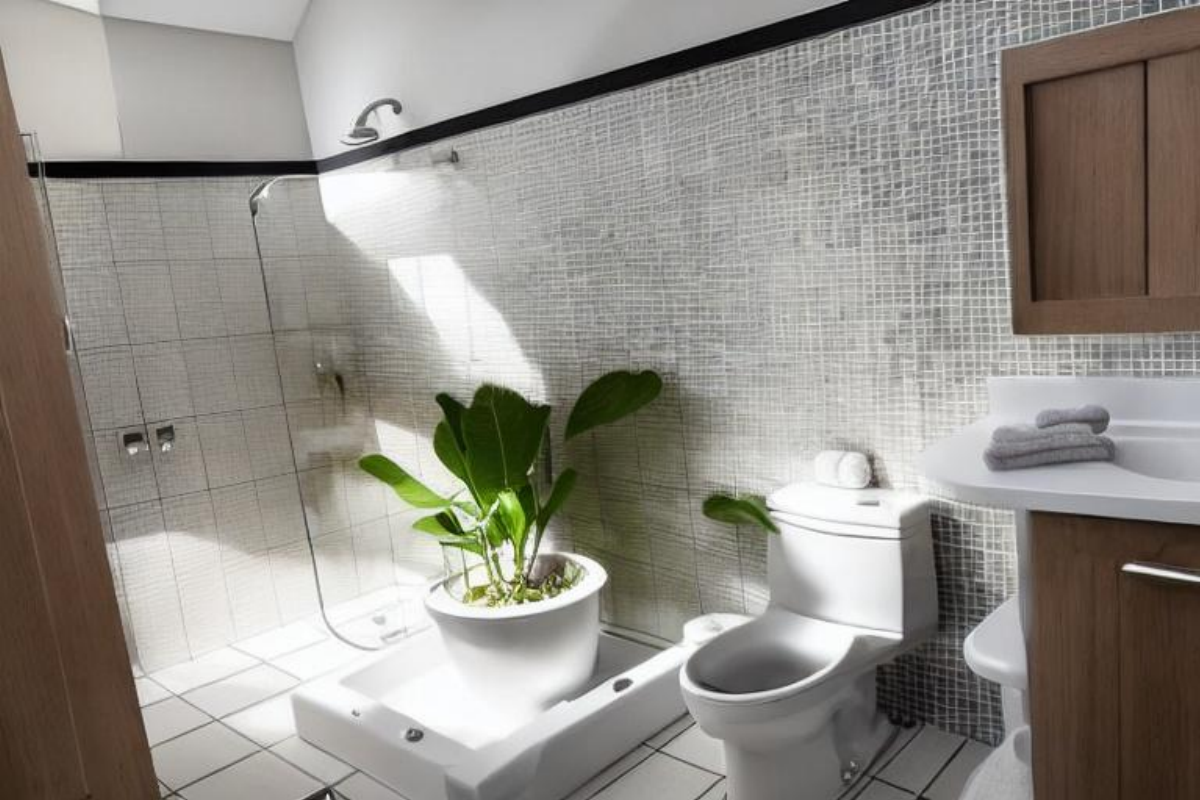}\\\vspace{1mm}
\includegraphics[width = \textwidth,height=24mm]{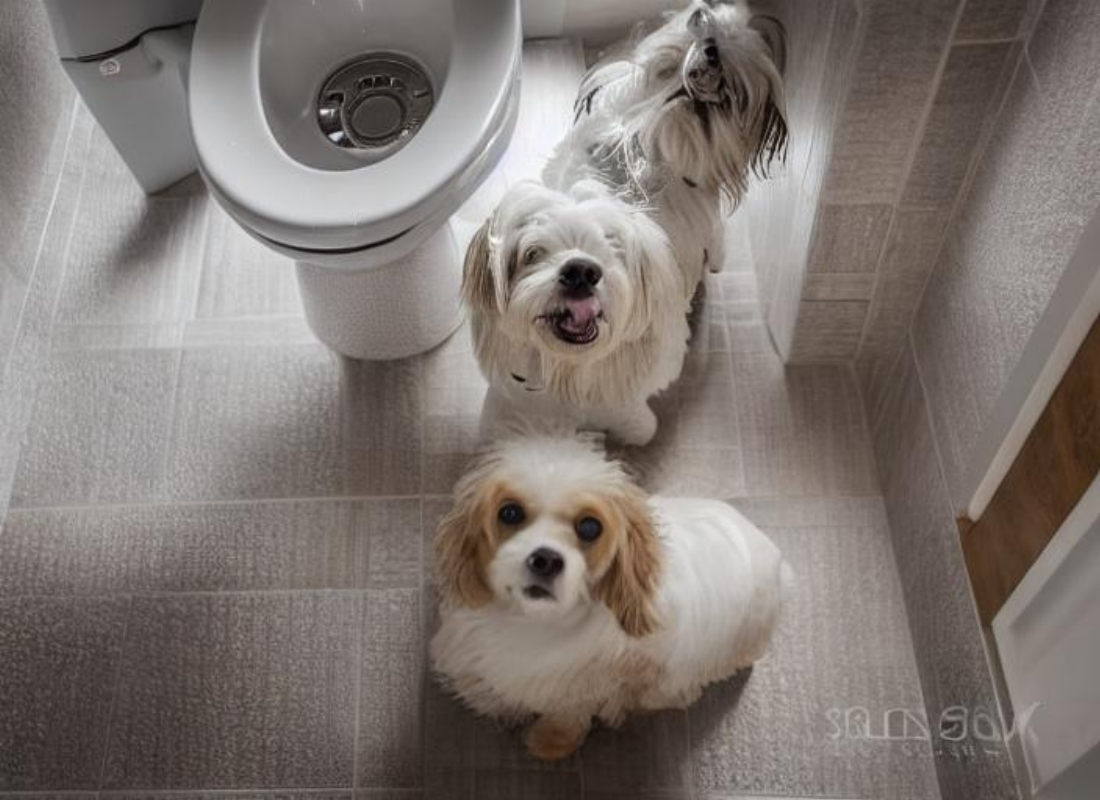}\\\vspace{1mm}
\includegraphics[width = \textwidth,height=24mm]{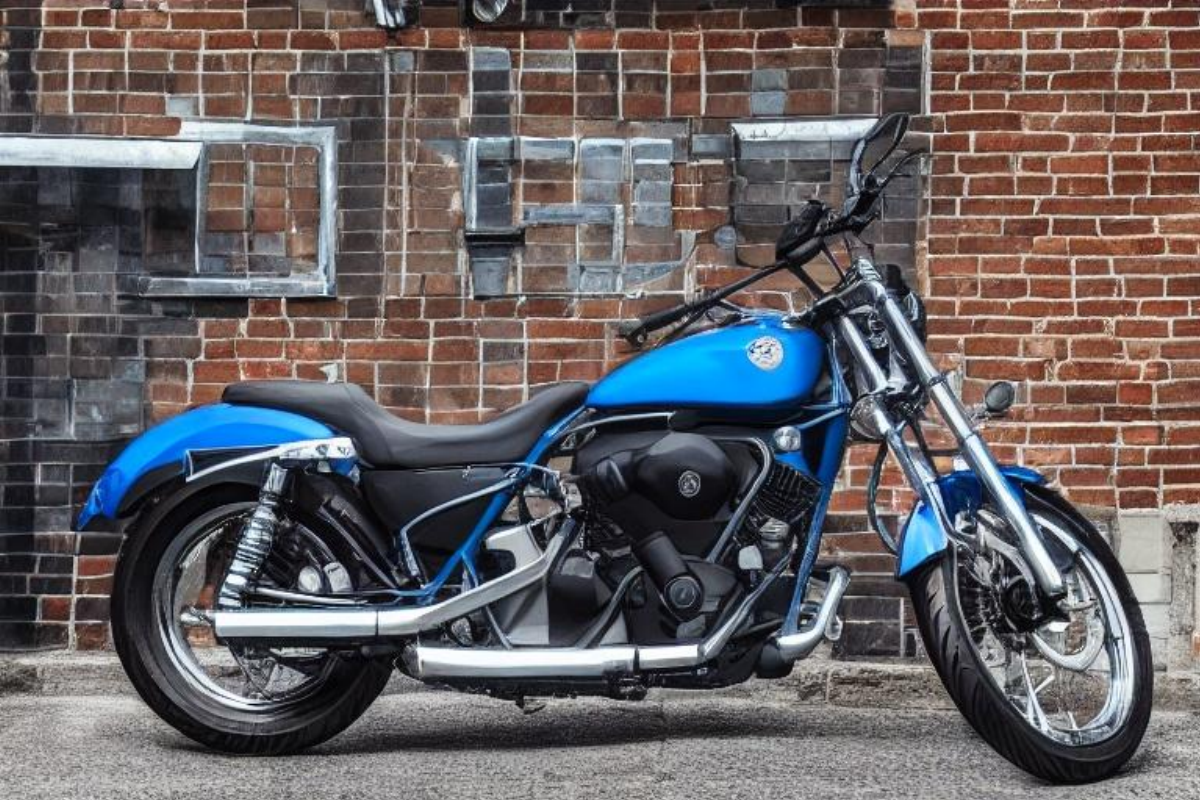}\\\vspace{1mm}
\includegraphics[width = \textwidth,height=24mm]{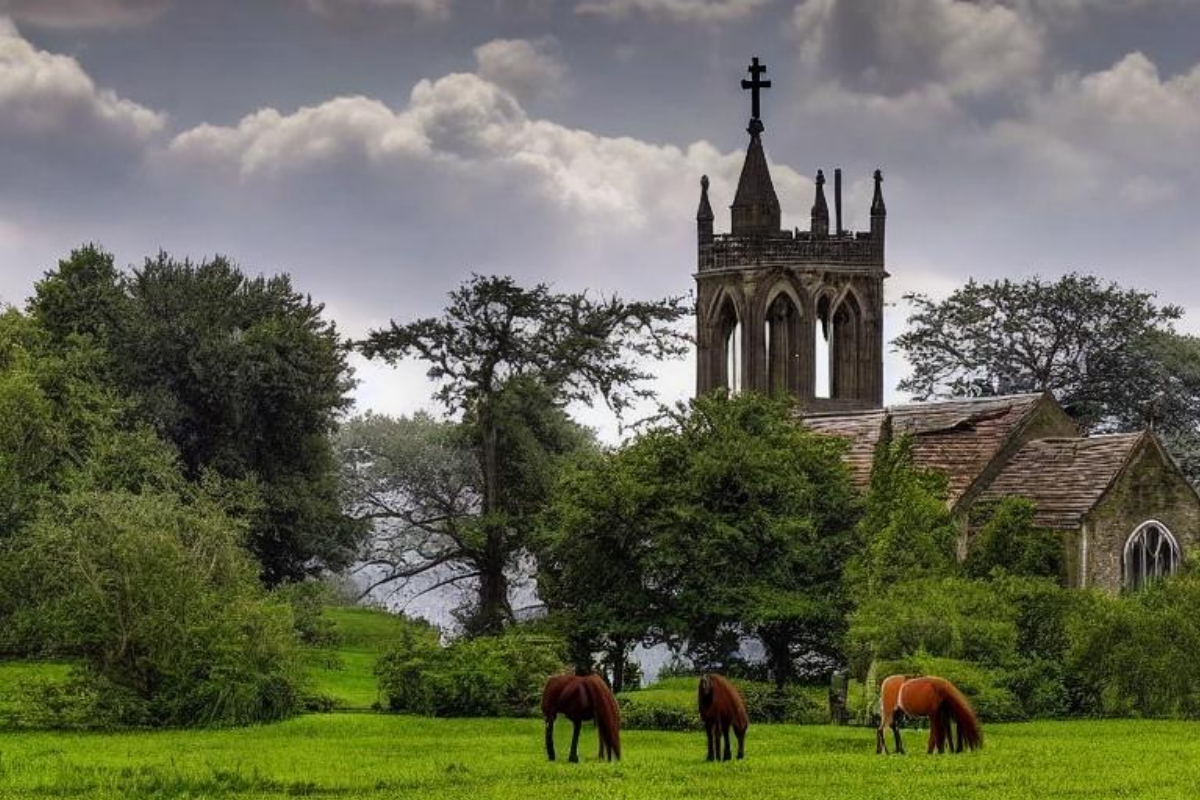}\\\vspace{1mm}
\includegraphics[width = \textwidth,height=24mm]{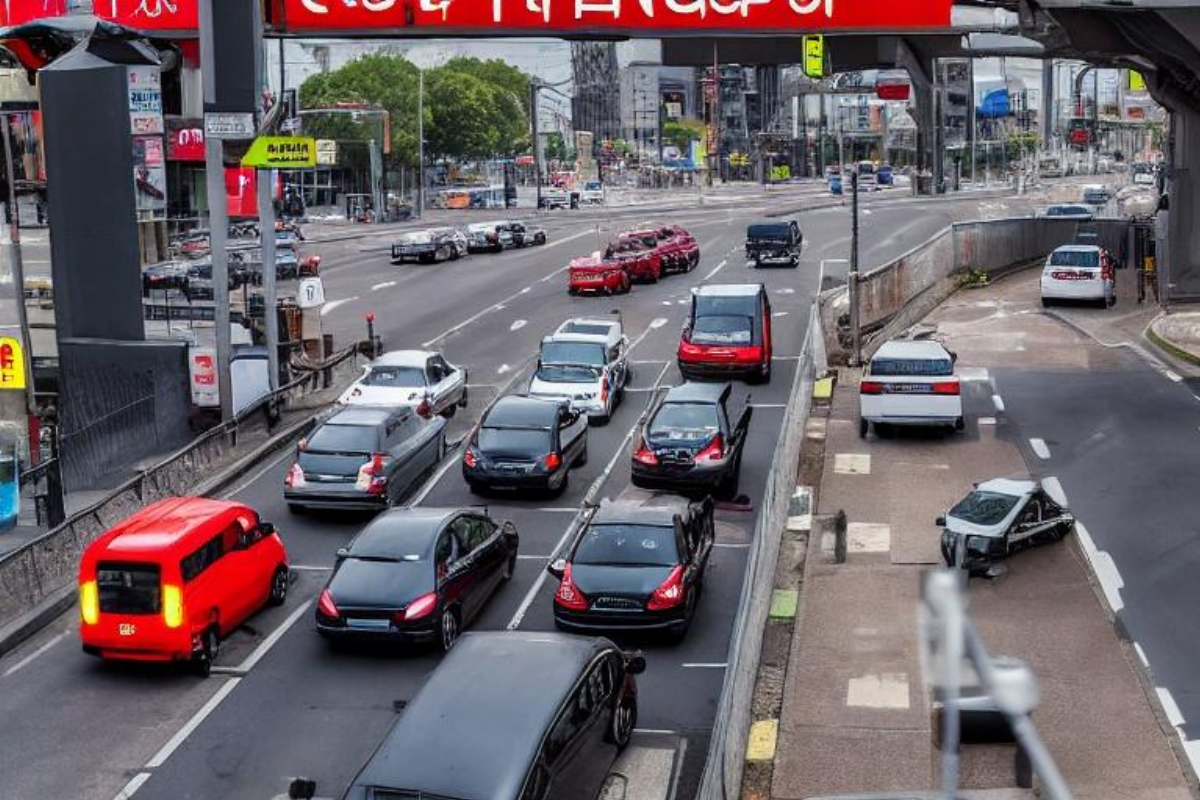}\\\vspace{1mm}
\includegraphics[width = \textwidth,height=24mm]{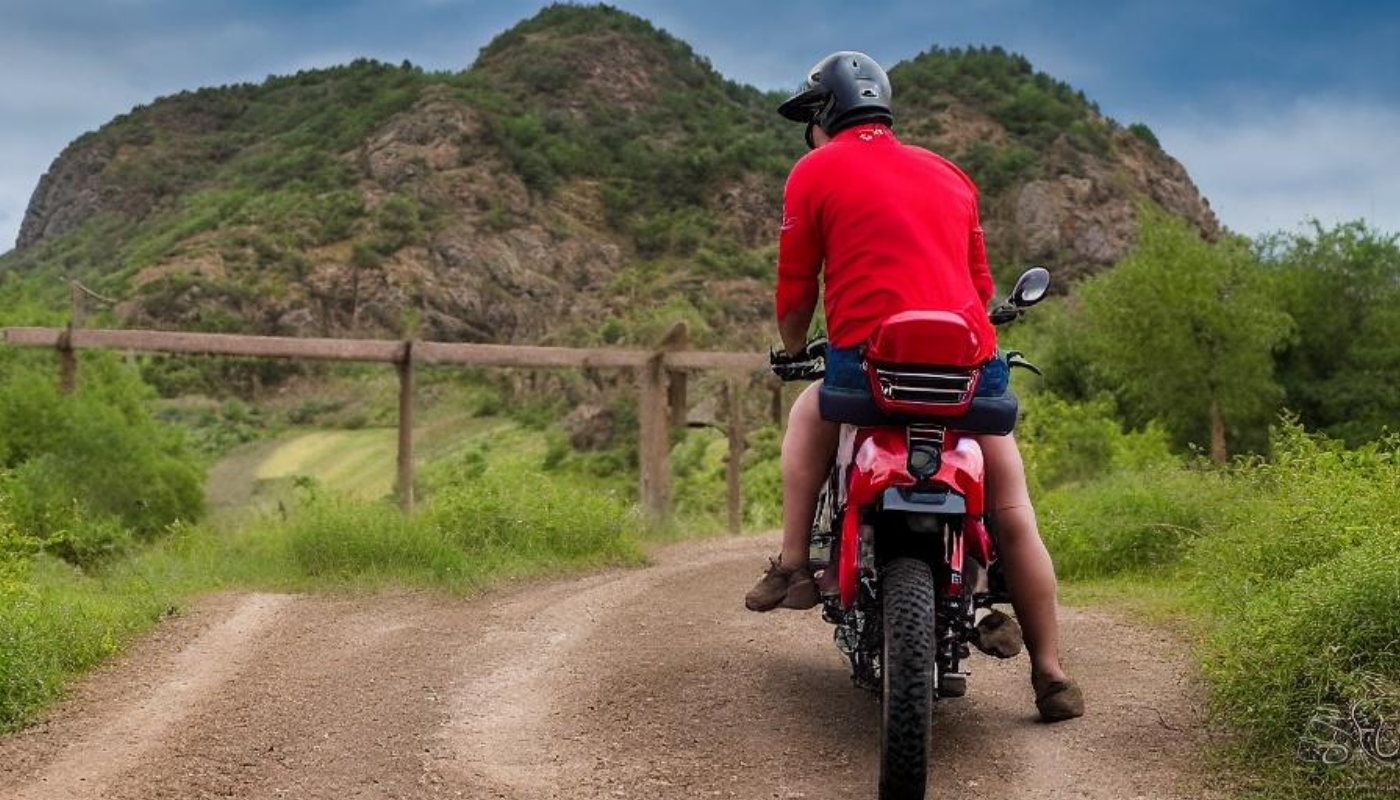}\\\vspace{1mm}
\includegraphics[width = \textwidth,height=24mm]{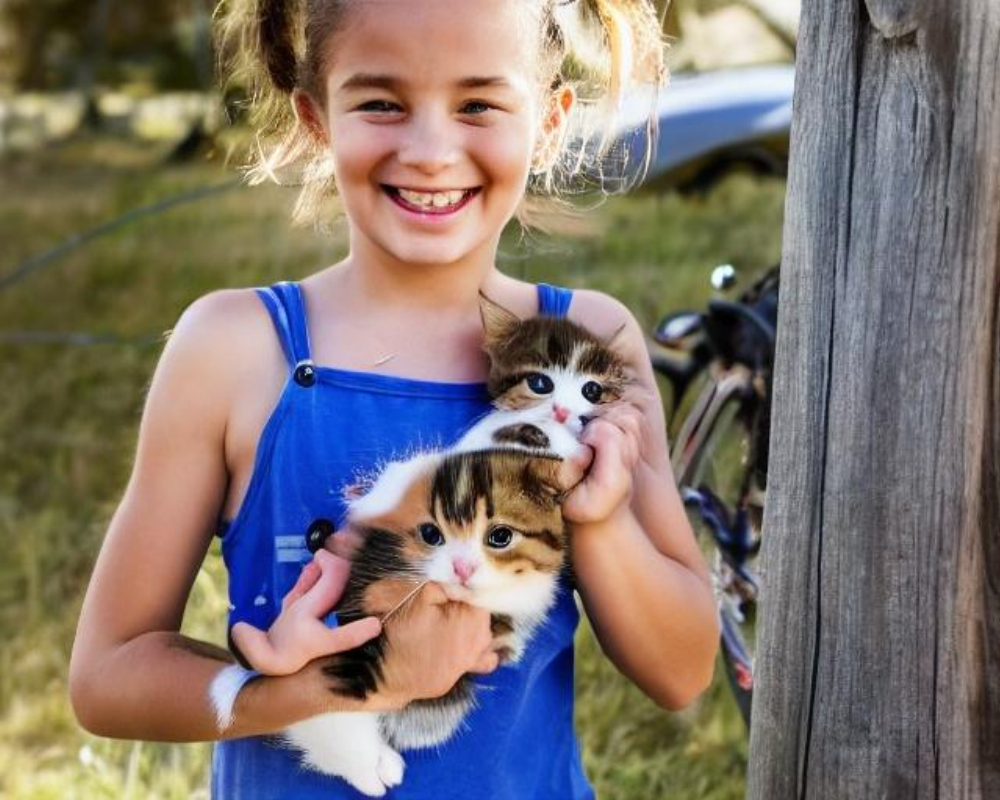}
\parbox[b][6mm][t]{20mm}{\caption*{\centering{LLaVA captions + ControlNet}}}
\end{subfigure}
\end{minipage}
\begin{minipage}[t]{0.19\linewidth}
\centering
\begin{subfigure}[b]{\textwidth}
\centering
\includegraphics[width = \textwidth,height=24mm]{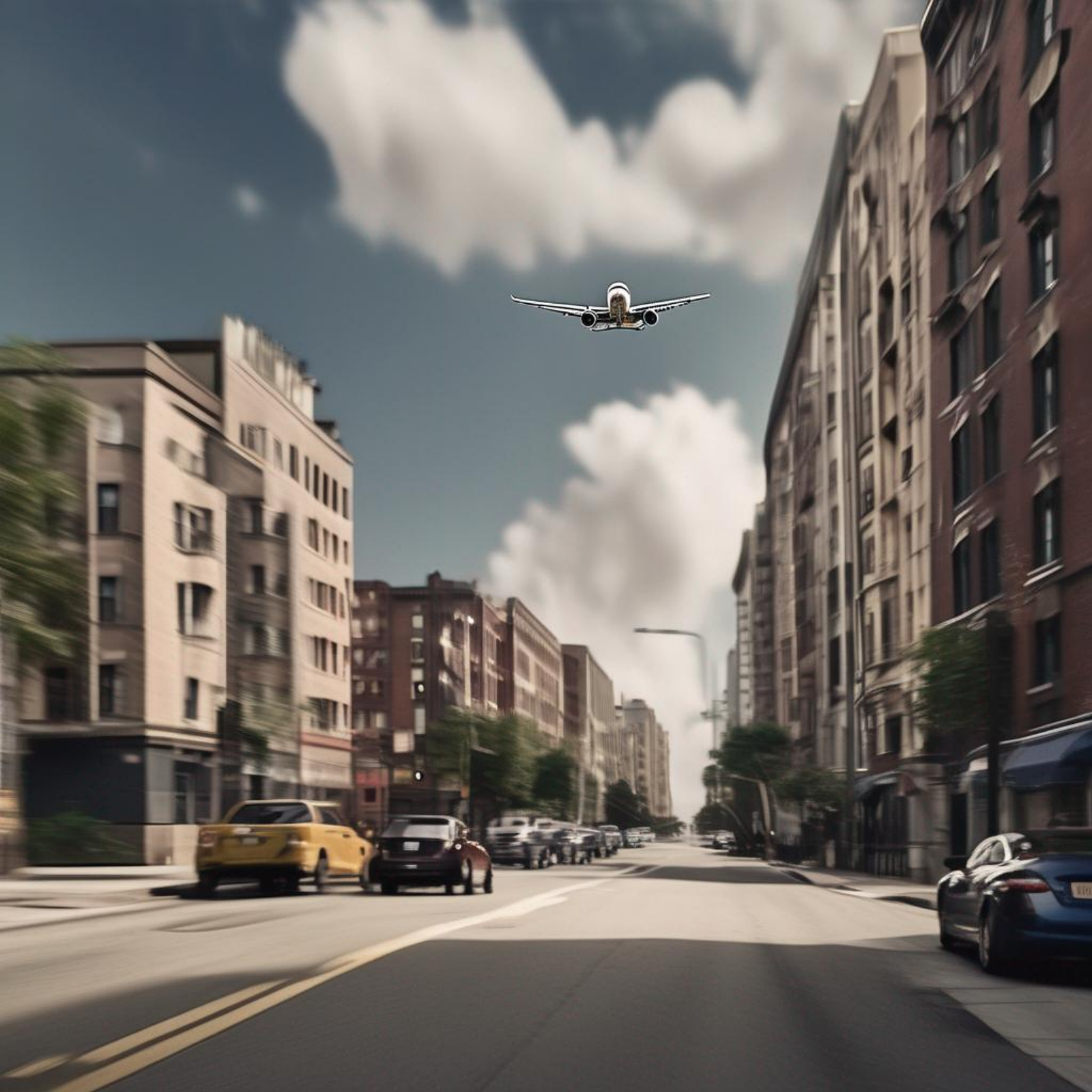}\\\vspace{1mm}
\includegraphics[width = \textwidth,height=24mm]{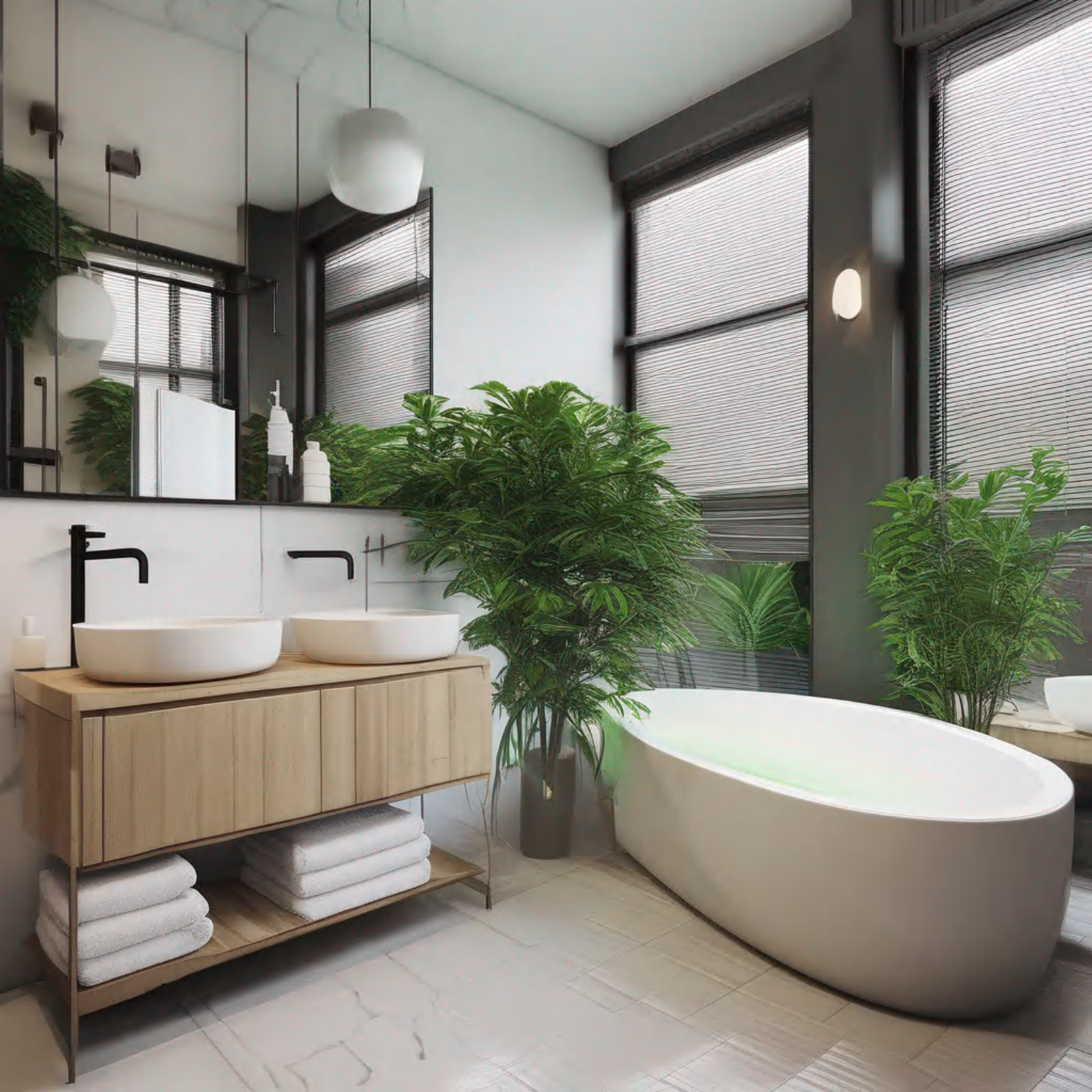}\\\vspace{1mm}
\includegraphics[width = \textwidth,height=24mm]{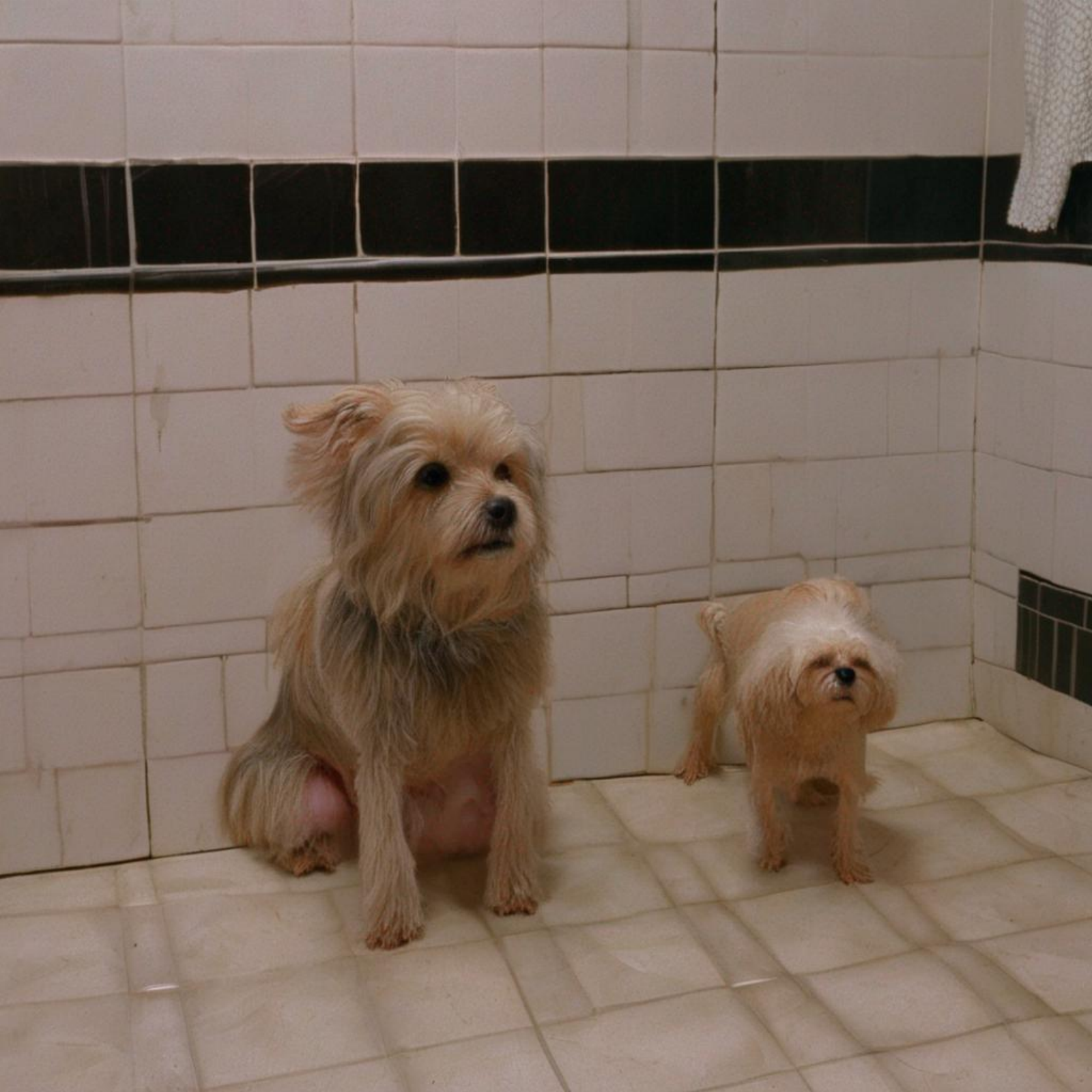}\\\vspace{1mm}
\includegraphics[width = \textwidth,height=24mm]{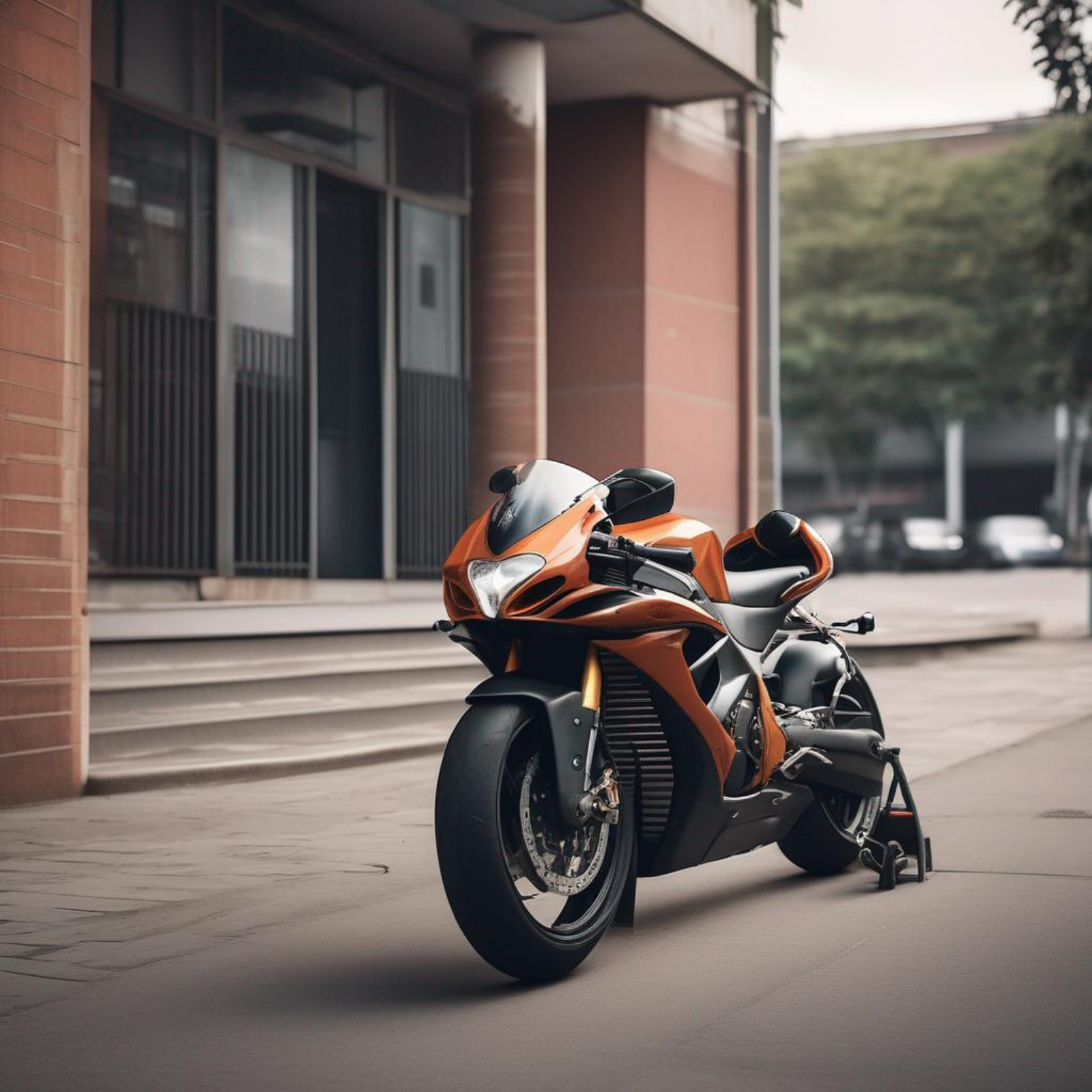}\\\vspace{1mm}
\includegraphics[width = \textwidth,height=24mm]{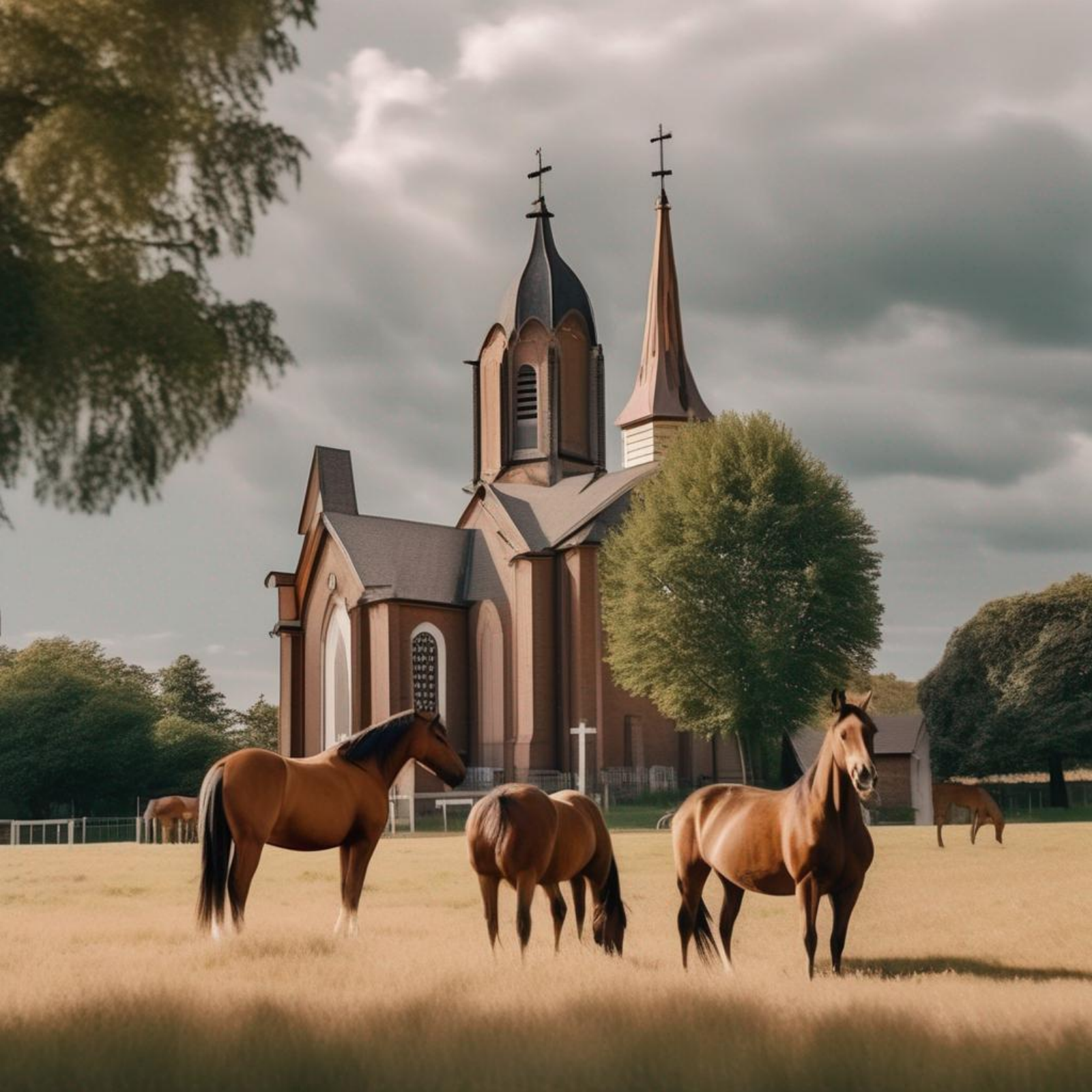}\\\vspace{1mm}
\includegraphics[width = \textwidth,height=24mm]{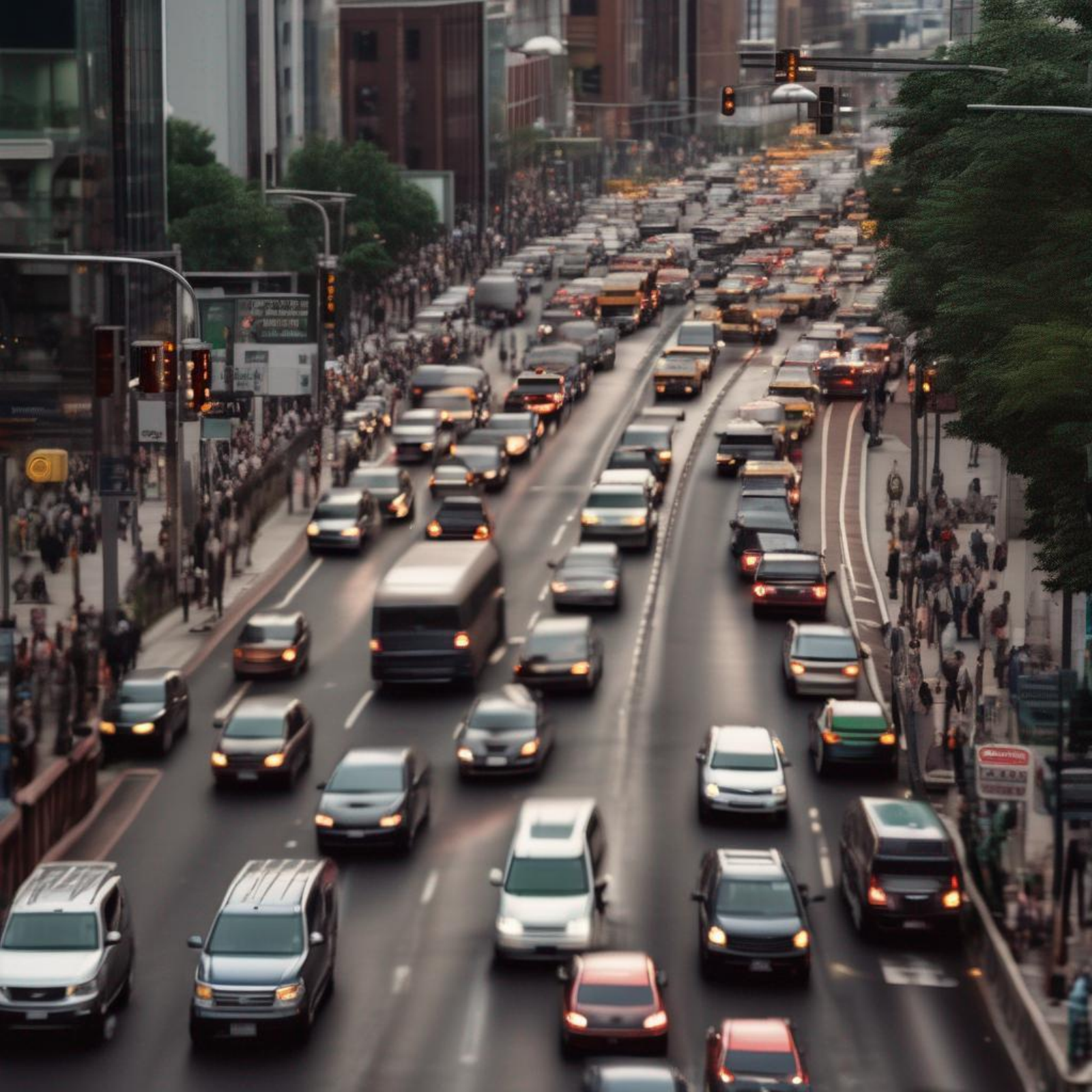}\\\vspace{1mm}
\includegraphics[width = \textwidth,height=24mm]{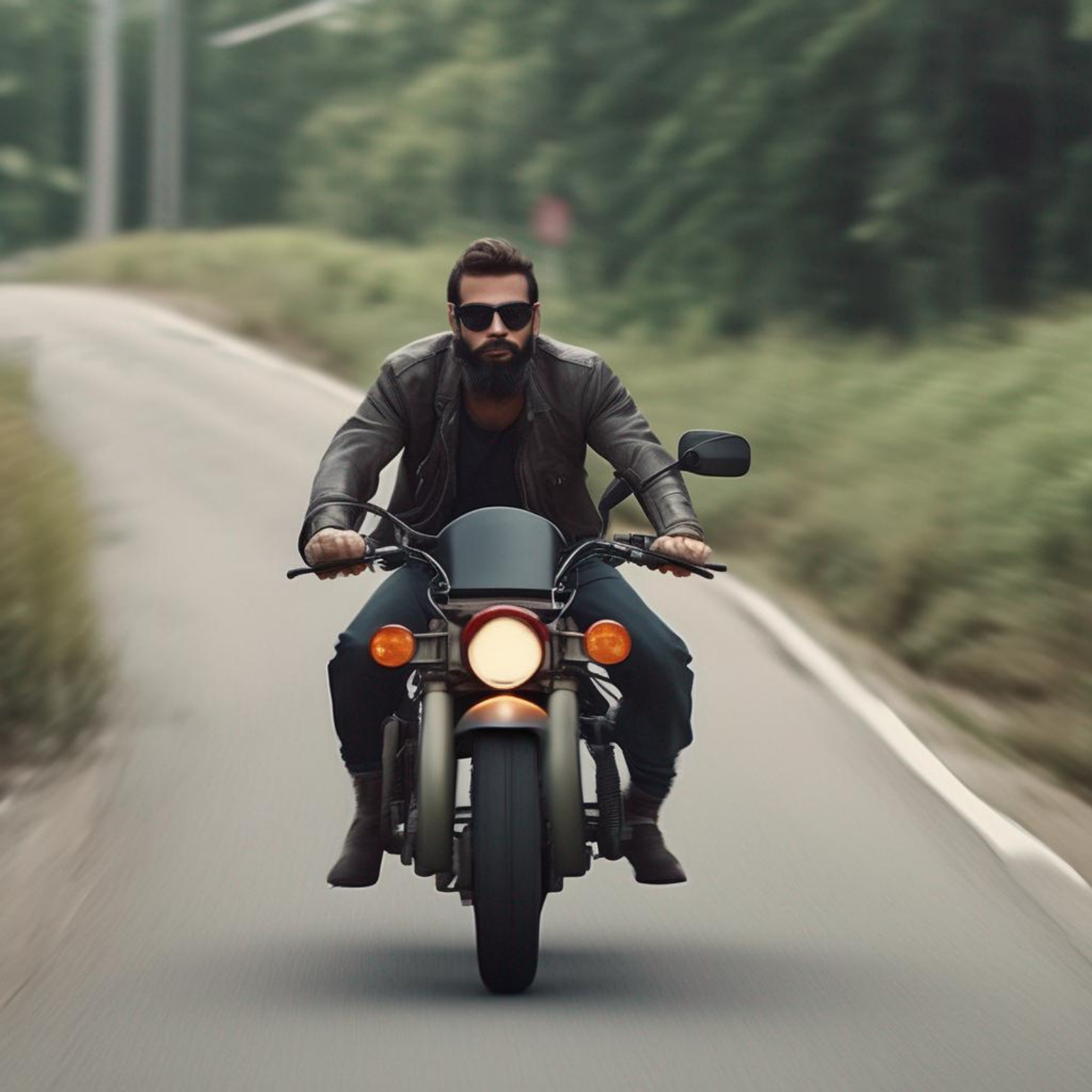}\\\vspace{1mm}
\includegraphics[width = \textwidth,height=24mm]{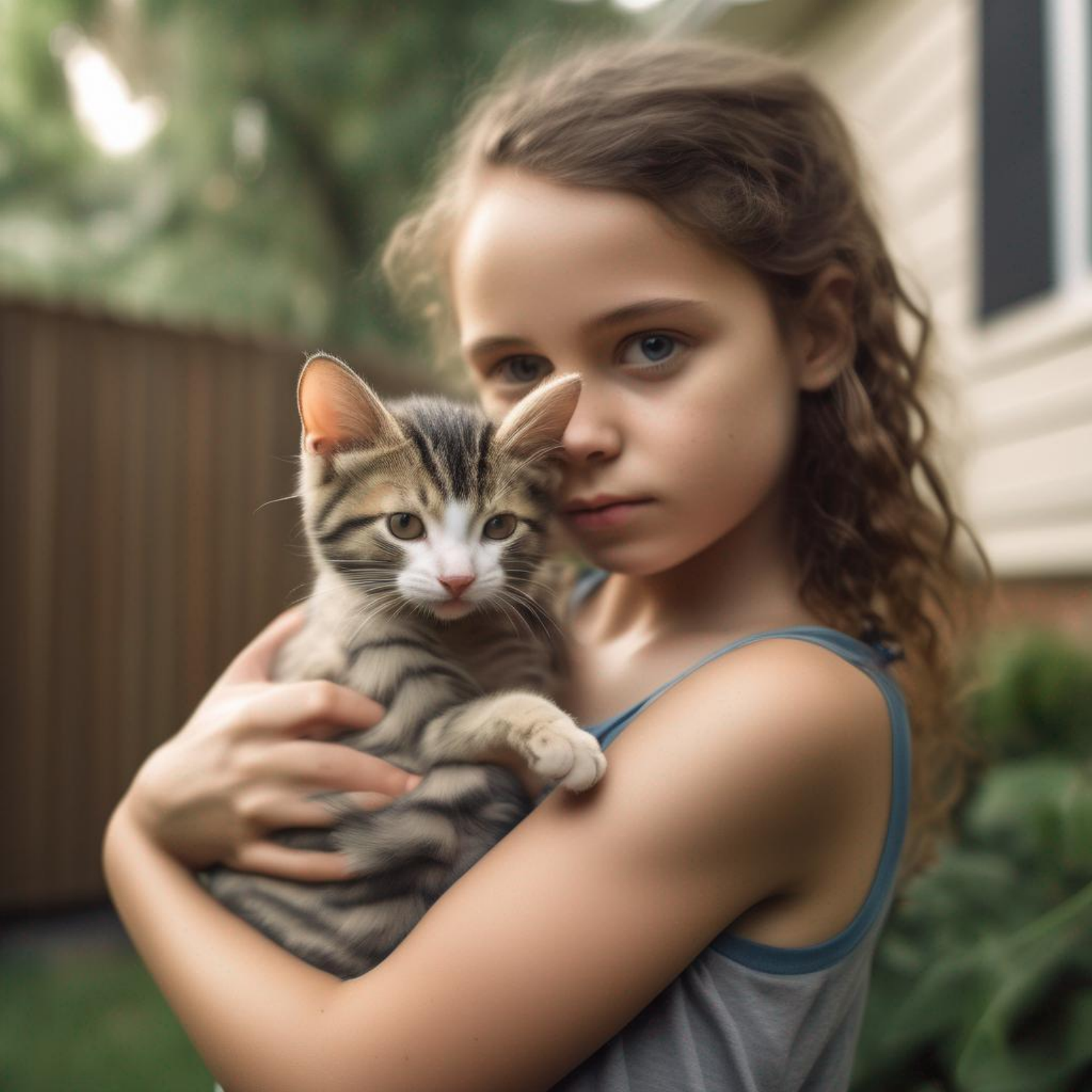}
\parbox[b][6mm][t]{20mm}{\caption*{\centering{COCO captions + SDXL}}}
\end{subfigure}
\end{minipage}
\begin{minipage}[t]{0.19\linewidth}
\centering
\begin{subfigure}[b]{\textwidth}
\centering
\includegraphics[width = \textwidth,height=24mm]{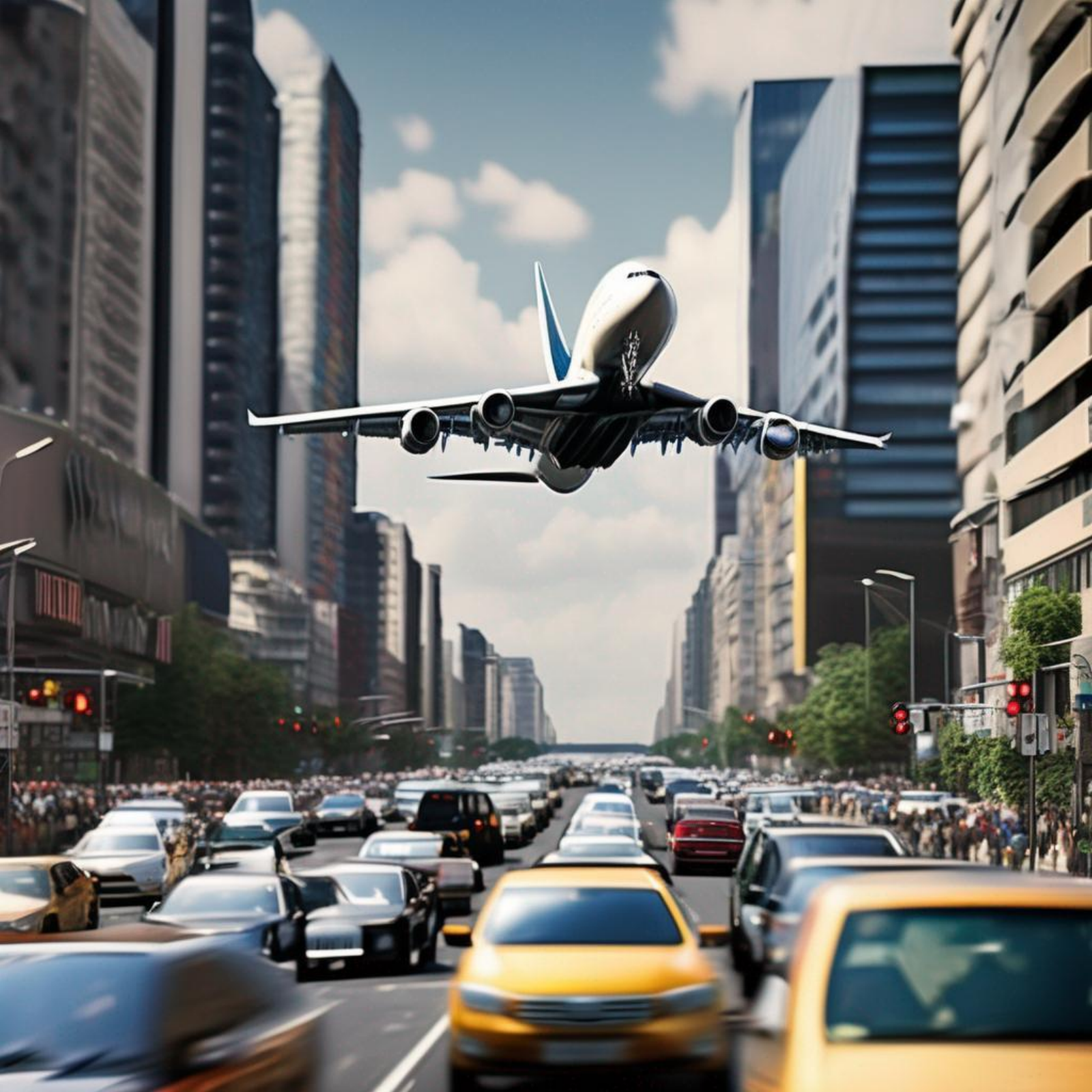}\\\vspace{1mm}
\includegraphics[width = \textwidth,height=24mm]{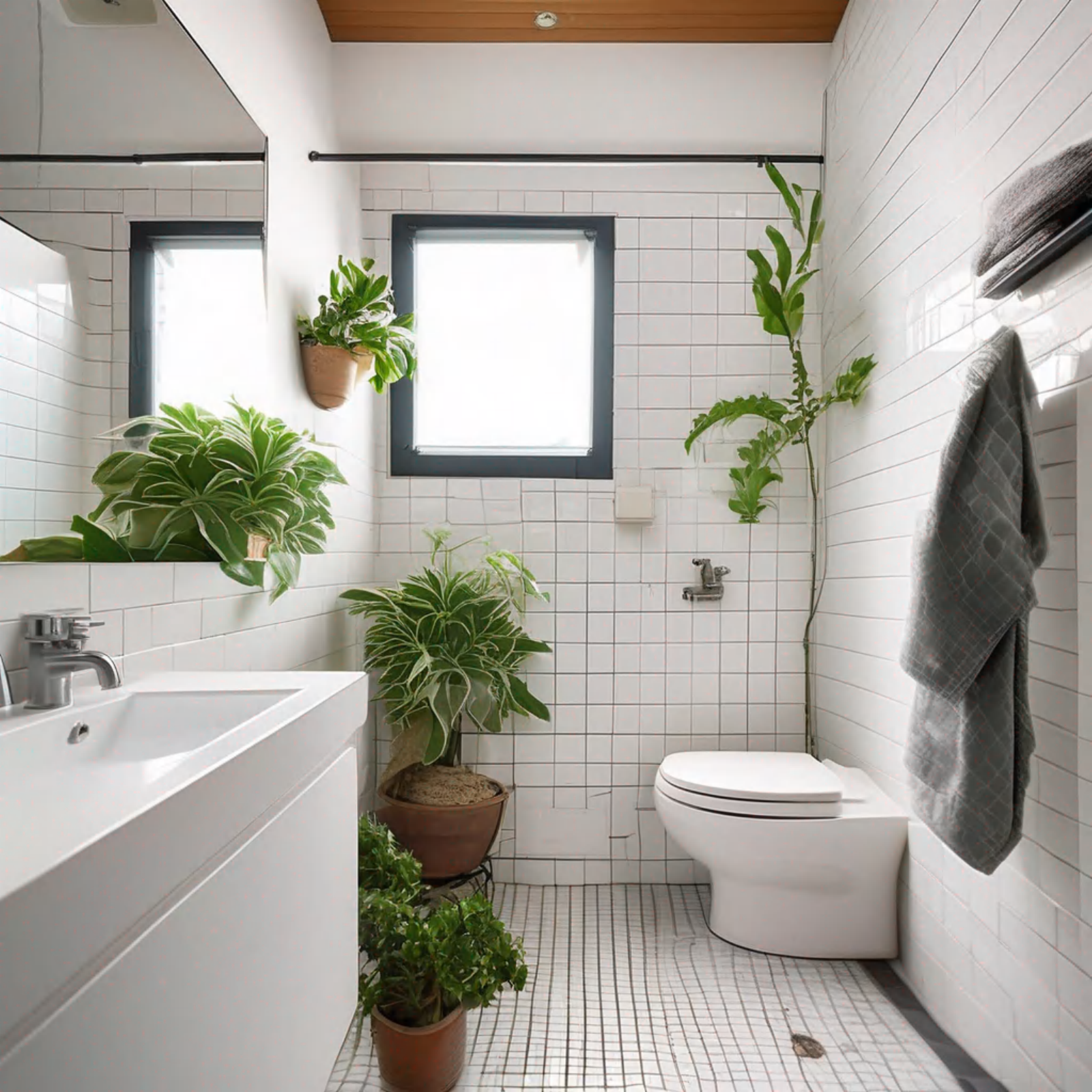}\\\vspace{1mm}
\includegraphics[width = \textwidth,height=24mm]{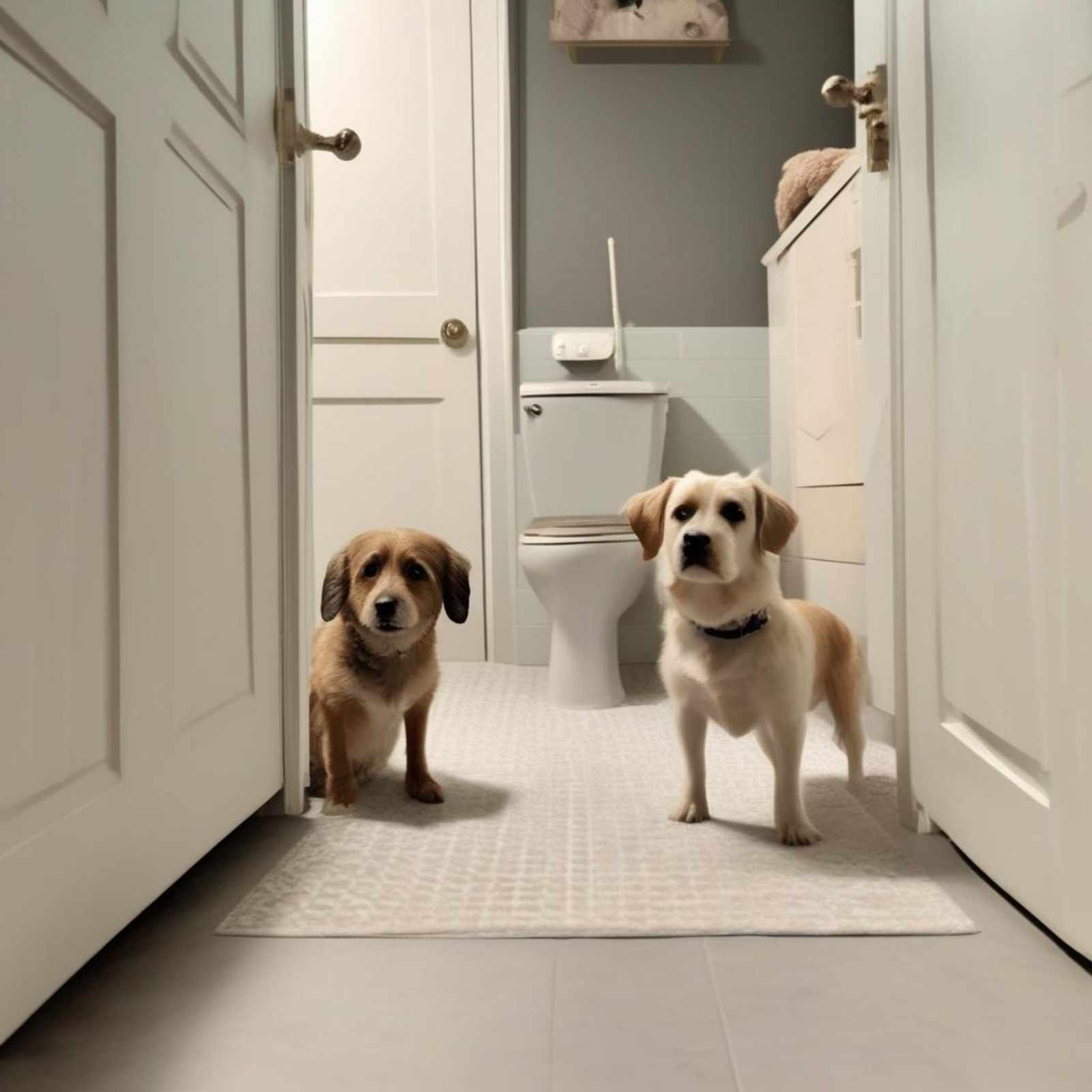}\\\vspace{1mm}
\includegraphics[width = \textwidth,height=24mm]{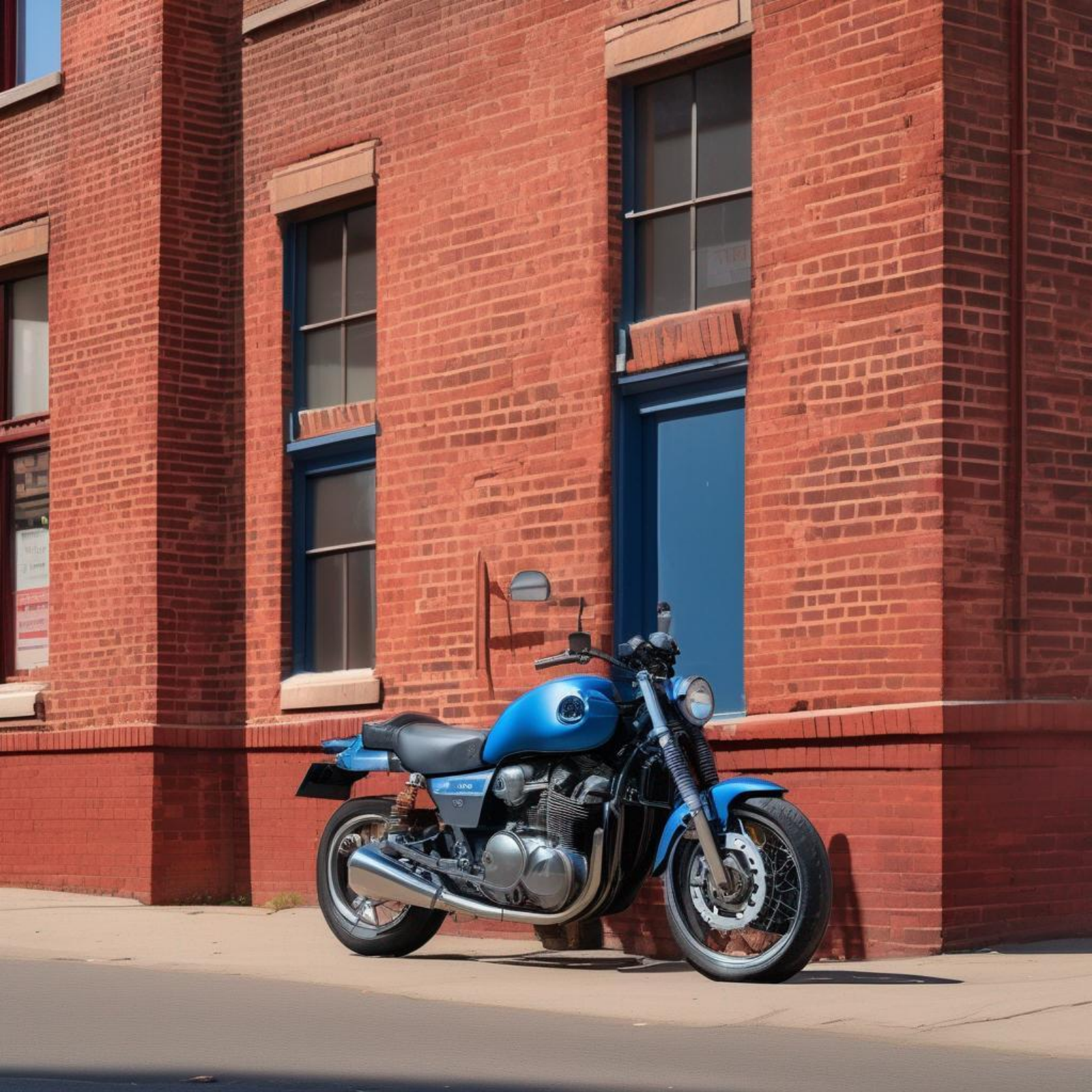}\\\vspace{1mm}
\includegraphics[width = \textwidth,height=24mm]{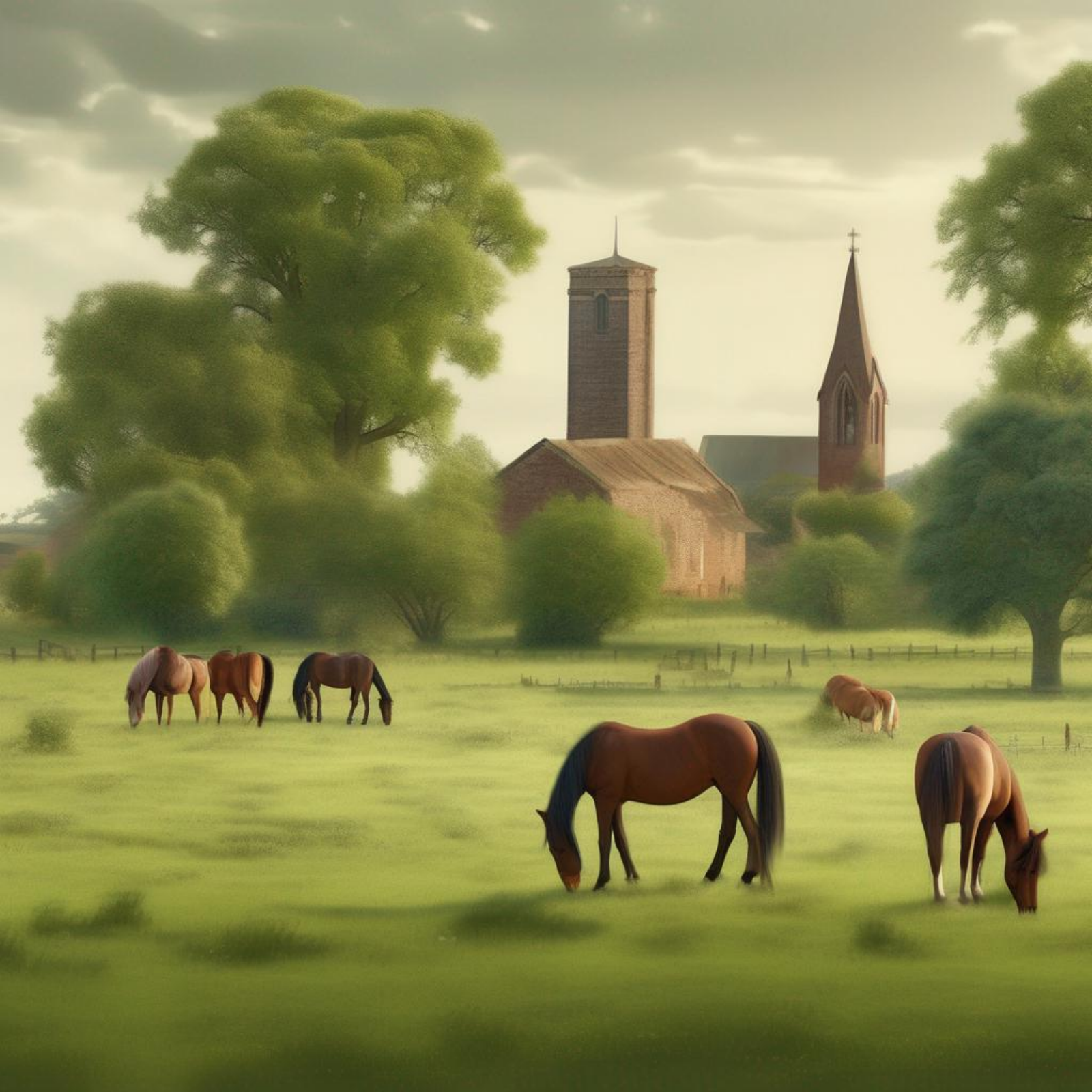}\\\vspace{1mm}
\includegraphics[width = \textwidth,height=24mm]{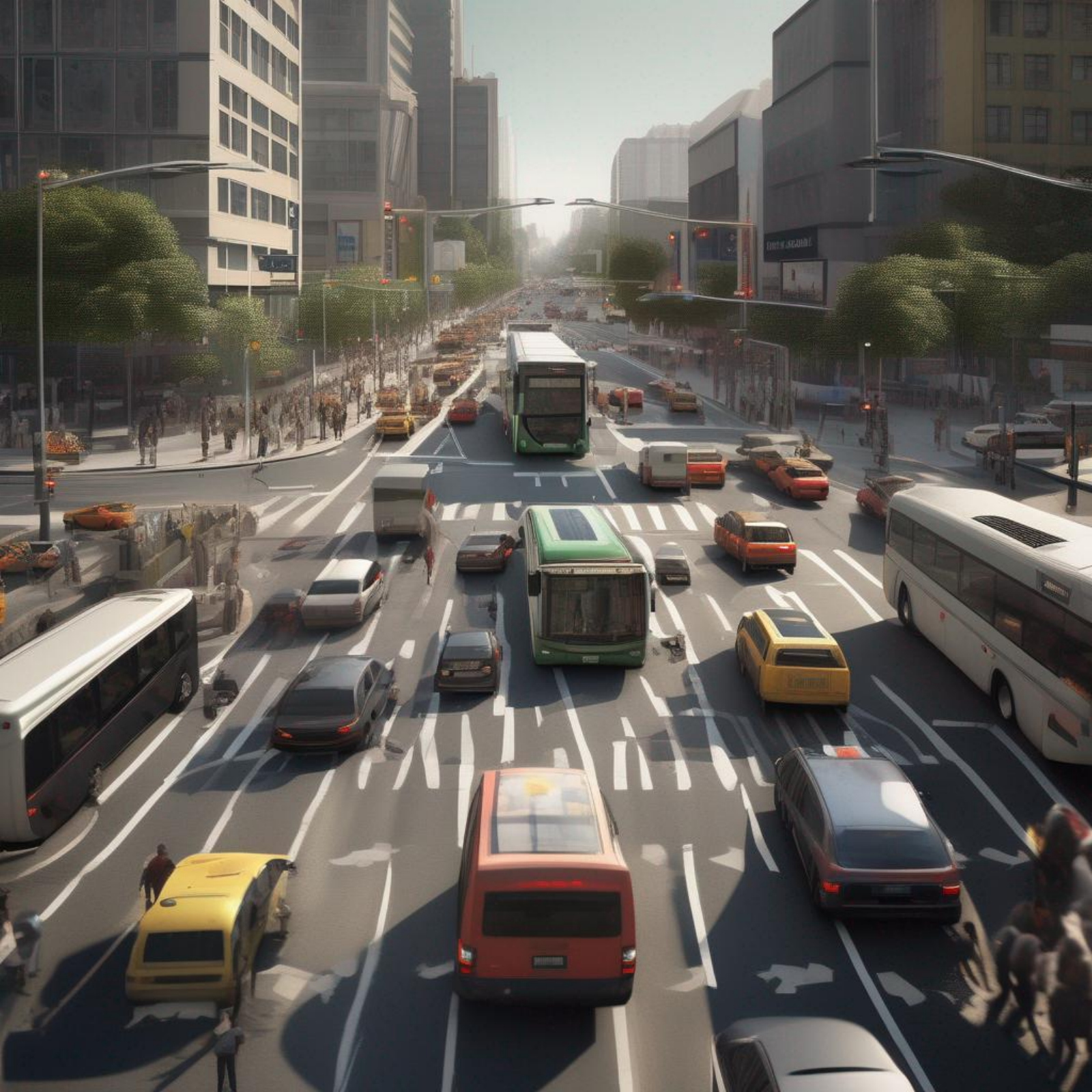}\\\vspace{1mm}
\includegraphics[width = \textwidth,height=24mm]{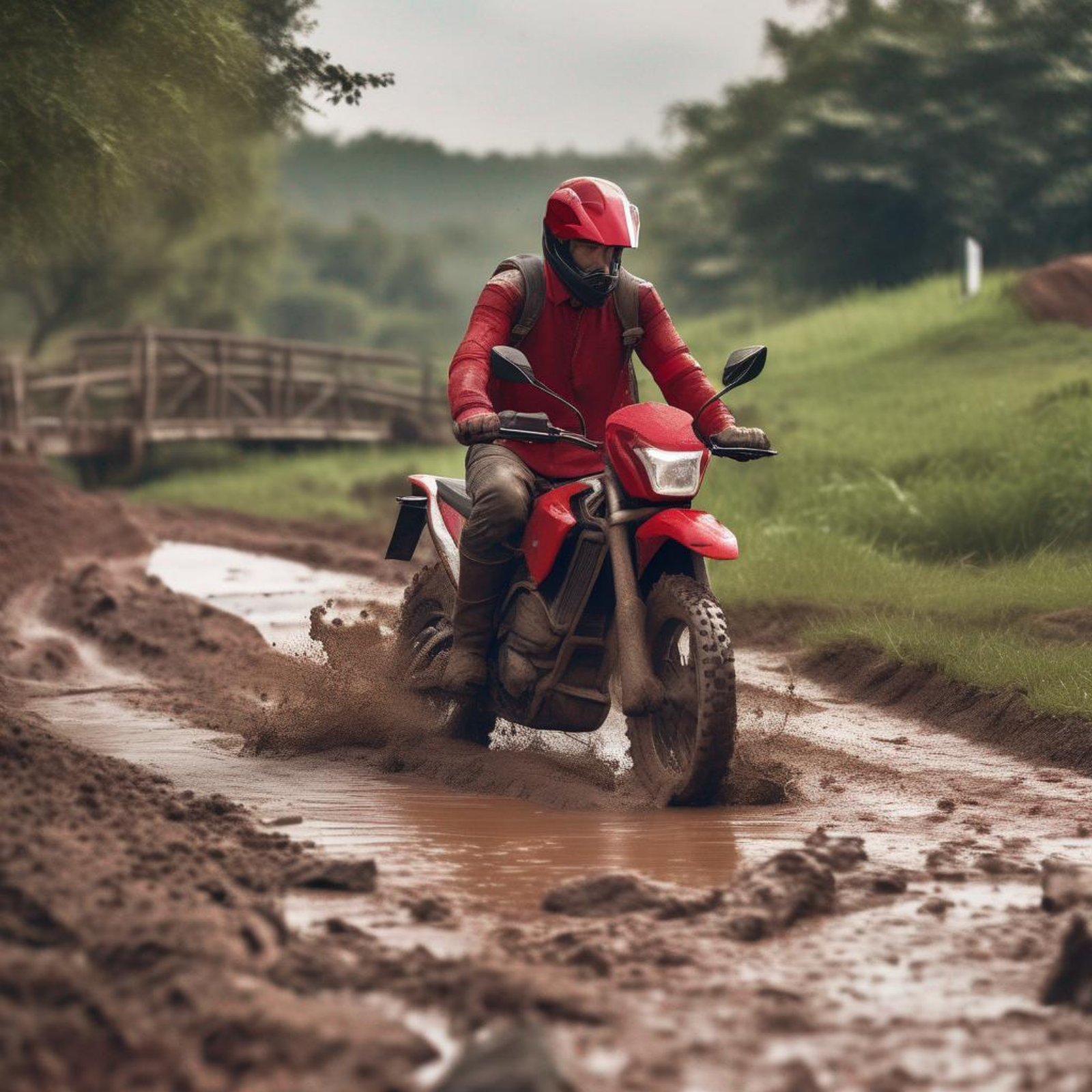}\\\vspace{1mm}
\includegraphics[width = \textwidth,height=24mm]{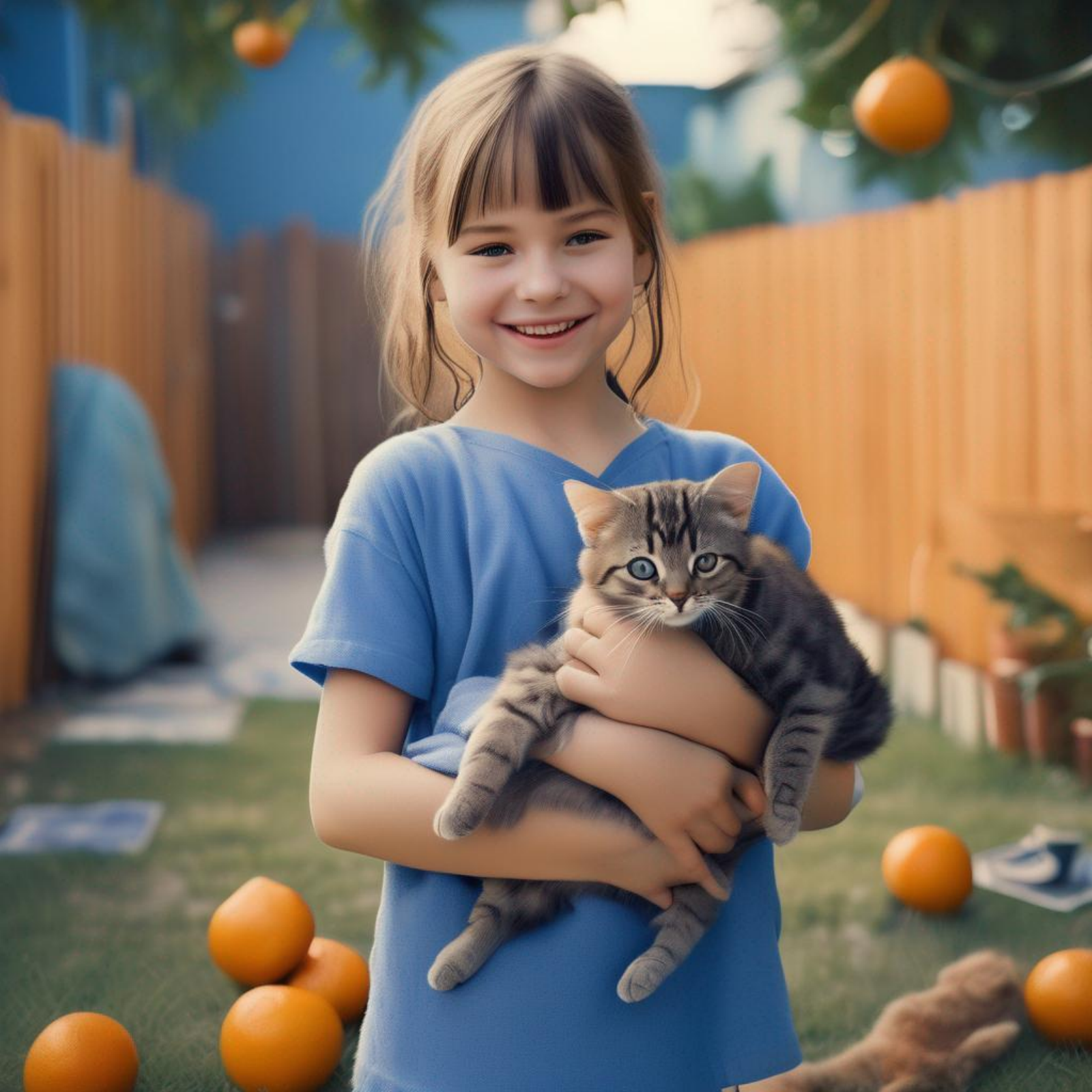}
\parbox[b][6mm][t]{20mm}{\caption*{\centering{LLaVA captions + SDXL}}}
\end{subfigure}
\end{minipage}
\caption{\small{Examples of synthetic images using different captions and generative models. The original images are sampled from COCO \texttt{train} set.}}
\label{fig:vis_synthetic_images}
\end{minipage}
\vspace{-11pt}
\end{figure*}

\textbf{\subsection{Results with synthetic data generated by T2I}}
Last, we investigate the effectiveness of pre-training with synthetic data, which is generated using recent large-scale text-to-image generation models. Specifically, we leverage two representative text-to-image models, ControlNet~\citep{zhang2023adding} and SDXL~\citep{podell2023sdxl}, to generate images. These models take original captions from the COCO dataset or captions generated by LLaVA~\citep{liu2023llava} as prompts for image synthesis.
ControlNet uses predicted depth maps from DPT~\citep{ranftl2021vision} as conditional input to generate images that match both the depth maps and captions. On the other hand, SDXL generates images solely based on the provided captions without any additional conditions. We create a synthetic dataset comprising 2.3 Million generated images. Figure~\ref{fig:vis_synthetic_images} displays some examples.
    
Upon analyzing the images produced by ControlNet, we find that they closely resemble the layout of the original images due to the conditioning on depth maps. This characteristic allows us to use COCO ground-truth data to supervise the pretraining process when using synthetic images generated by ControlNet. Additionally, we also explore the Simple Self-training approach on the synthetic data by pre-training with pseudo-box and pseudo-class predictions that are generated by trained COCO detectors. The process involves pre-training the $\mathcal{H}$-Deformable-DETR model with synthetic images for 3 epochs, followed by fine-tuning on COCO or PASCAL VOC benchmarks for 12 epochs. The results of this evaluation are presented in Table~\ref{tab:result_of_synthetic_images}.
Interestingly, pre-training with the synthetic dataset generated based on COCO demonstrates comparable improvements to pre-training with Objects$365$ real data using the Simple Self-training scheme. This outcome indicates that text-to-image synthesis is an effective method for scaling up the original dataset for pre-training. Furthermore, the results on the PASCAL VOC benchmark showcase the generalization ability of pre-training with synthetic data generated based on COCO. 

Table~\ref{tab:result_of_synthetic_images_o365} shows the results of pre-training with the synthetic data generated based on Objects$365$ by first captioning Objects$365$ image with LLaVA and then synthesizing new images from the caption. They are not as good as pre-training with COCO-based synthetic data on both downstream benchmarks.

%%%%%%%%%%%%%%%%%%%%%%%%%%%%%%%%
\vspace{3mm}
\section{Conclusion}
%%%%%%%%%%%%%%%%%%%%%%%%%%%%%%%%
We investigate the effectiveness of DETReg, a representative self-supervised pre-training approach for DETR, across three distinct DETR architectures. Our findings, unfortunately, do not reveal any performance enhancements of DETReg in recent architectures, thereby challenging the validity of previous conclusions. In response, we reevaluate crucial design aspects, including pre-training targets for localization and classification. As a result of this analysis, we introduce several impactful enhancements and a Simple Self-training scheme that significantly boosts performance across strong DETR architectures.
Additionally, we leverage the powerful text-to-image generative models to construct synthetic datasets for pre-training purposes. Remarkably, our approach yields improvements on par with the achievements of pre-training with Objects365. Moving forward, we plan to extend DETR pre-training to encompass a broader spectrum of vision tasks, such as instance segmentation and pose estimation.
We hope our work can stimulate the research community to reassess the actual capacity of existing self-supervised pre-training methods when employed in the context of strong DETR models and advance the progress on this challenging task. 

%%%%%%%%% REFERENCES
{	
\clearpage
\clearpage
\section*{Data availability statement}
The author confirmed that the data supporting the findings of this study are available within the article. Raw data that support the findings of this study and the generated synthetic dataset are available from the corresponding author, upon reasonable request. 

\bibliographystyle{spbasic}\small
\bibliography{egbib}
}

\end{document}